\documentclass[10pt,onecolumn,letterpaper]{article}
\usepackage{booktabs}

\usepackage{makeidx}
\usepackage{graphicx, subfigure}  
\usepackage{amsmath}
\usepackage{setspace}
\usepackage{cite}

\usepackage{authblk}

\providecommand{\keywords}[1]{\textbf{\textit{Keywords---}} #1}

\begin{document}

\title{A Comparative Study of Techniques of Distant Reconstruction of Displacement Fields by using DISTRESS Simulator}

\author[1,2]{Ghulam Mubashar Hassan}
\author[2]{Arcady V Dyskin}
\author[1]{Cara K MacNish}

\affil[1]{School of Computer Science \& Software Engineering}
\affil[2]{School of Civil \& Resource Engineering \\
University of Western Australia}

\maketitle

\begin{abstract}
Reconstruction and monitoring of displacement and strain fields is an important problem in engineering. We analyze the remote and non-obtrusive methods of strain measurement based on photogrammetry and Digital Image Correlation (DIC). The method is based on covering the photographed surface with a pattern of speckles and comparing the images taken before and after the deformation. In this study, a comprehensive literature review and comparative analysis of photogrammetric solutions is presented. The analysis is based on a specially developed Digital Image Synthesizer To Reconstruct Strain in Solids (DISTRESS) Simulator to generate synthetic images of displacement and stress fields in order to investigate the intrinsic accuracy of the existing variants of DIC. We investigated the Basic DIC and a commercial software VIC 2D, both based on displacement field reconstruction with post processing strain determination based on numerical differentiation. We also investigated what we call the Extended DIC where the strain field is determined independently of the displacement field. While the Basic DIC and VIC 2D are faster, the Extended DIC delivers the best accuracy of strain reconstruction. The speckle pattern is found to be playing a critical role in achieving high accuracy for DIC. Increase in subset size for DIC does not significantly improves the accuracy, while the smallest subset size depends on the speckle pattern and speckle size. Increase in the overall image size provides more details but does not play significant role in improving the accuracy, while significantly increasing the computation cost.
\end{abstract}

\keywords{Remote deformation monitoring, Photogrammetry, Digital Image Correlation, Comparative analysis, DIC, Displacement field reconstruction, Strain field reconstruction}

\section{Introduction}
\label{sec:intro}
Deformation monitoring plays a critical role in engineering and sciences. It is used to reconstruct strain fields and monitor structural health, especially approaching instability and failure \cite{Tyson11}. This study is focused on monitoring small deformation in solids under load.

A proper reconstruction of the strain field requires measuring strain at many points in the solid. Strain gauges commonly used to measure the strain are electro-mechanical components which measure change in electrical resistance due to load and accurately determine the displacement increment over the strain gauge length at the points where they are attached and in the direction of their orientation. Subsequently, measuring different components at multiple points requires the use of multiple strain gauges.

Using multiple strain gauges is labor intensive and expensive. Furthermore, installing strain gauges on the structure and monitoring deformation may interfere with the normal use of structure. For instance, installing and monitoring strain gauges at excavation walls in mines obstructs the excavation process. The strain gauges may also be damaged when structural instability or failure are involved.

Photogrammetry provides a remote and non-obtrusive alternative to monitor the strains of different points in the structure. Since, it only involves taking images of the surface, it reduces obstruction of the normal use of the structure.

The major photogrammetric techniques that can be used to monitor the deformation from digital images are based on comparing two images, \emph{reference} and \emph{deformed}, of the surface with artificial or natural pattern which are taken before and after deformation. The photogrammetric techniques select a pixel and its neighboring area, or subset, in the reference image. The subset is then correlated with the subsets in the deformed image and new position of the pixel in the deformed image is approximated. The process is repeated for all pixels in the reference image and displacement fields, and in some cases displacement gradient fields, are reconstructed. The correlation is based on using either the Least Squares Method or Cross-correlation \cite{Gruen1985}. Currently, they are predominantly used in computer vision to construct three dimensional views of an object from two dimensional images \cite{Barazzetti2011iWitness,Werner2002}.

Many techniques and their variants based on cross-correlation of images are proposed over last three decades for deformation monitoring under different titles depending on their use and the field of
application. Digital Image Correlation (DIC) or Digital Image Tracking is proposed for deformation monitoring in solids \cite{Pan09review, Sutton08review}. Similarly, the Optical Flow Block Matching Method is proposed to monitor deformation and test strength of soil structures \cite{Rodriquez2011,Rodriguez2012}. Particle Image Velocimetry is proposed for use in monitoring flow of fluids or movement in soil \cite{Roux2009}. All these techniques are similar and are based on the statistical concept of cross-correlation of the images, captured during the process of deformation. Cross-correlation is also extensively used in Digital Image Registration in the field of machine vision \cite{Brown1992survey, Shah2010}.

Deformation monitoring involves measuring very small displacements with high accuracy. It is required for accurate reconstruction of strain fields which involves the determination of displacement gradients, typically of the order of tenth and hundredth of percent of a pixel of an image. The required high accuracy of displacement measurement is obtained by using images with high resolution and techniques with subpixel level accuracy. High resolution images significantly increase the processing time while subpixel level accuracy requires good estimation technique.

Since the inception of cross-correlation in images, many of its variants are proposed to find high level accuracy at subpixel level. Unfortunately, no comprehensive study could be found which compares the performance and accuracy of these techniques. This paper aims to close this gap.

This paper focuses on Digital Image Correlation (DIC) which monitors deformation in solids. The DIC technique is used to monitor deformation at different accuracy levels in different situations. At large scales it can be used to analyse satellite images and monitor tectonic plate movement. In this case a pixel of the image may represent kilometers \cite{TectonicPlatesRef2012}. At the opposite end of the scale spectrum is thermal expansion monitoring in circuit components where a pixel of the image represents micrometers \cite{Rodriguez2012}. Given that the actual displacement determination is conducted in the image, the accuracy of DIC is commonly defined in term of pixels \cite{Tong05,Chu85,Hoult2013}.

DIC has the advantage of having simple setup and experimental preparation. It has some limitations, however, which need to be addressed. The major limitation of DIC is that the object under load
needs to develop a continuous displacement field. Discontinuities typically occur with fractures coming out to the surface on which the displacements are measured. Another limitation is the need of a random intensity pattern covering the surface and visible on the image. This is necessary to distinguish different points of the image and be able to trace their movement \cite{Sutton00}. Since photogrammetry involves taking images from a distance its performance may deteriorates in noisy (for example dusty or smoky) environment \cite{Ha09,Gilo11}.

Ideally the verification of accuracy of photogrammetric methods and comparison of different variants could be conducted by targeted experiments and comparing the photogrammetric results with the results of traditional (e.g. strain gauge) measurements of displacement and strain fields. There are however restrictions in the number of tests which can feasibly be undertaken as well as the possibility to control all influencing factors.

Therefore, we propose to conduct the initial analysis of accuracy delivered by different types of the DIC technique using computer experiments based on generation of synthetic images. (We note that synthetic images are used to evaluate the accuracy of the DIC technique in computer vision where the focus is entirely different than that of the deformation monitoring.)

We developed a computer simulator to produce synthetic images which simulate real displacement and strain fields. The images obtained by the simulator are used to find the intrinsic accuracy of the DIC algorithms. Three most commonly used variants of DIC algorithms are selected and their performance is analyzed in this paper.

This paper is organized as follows: the technique of photogrammetry is discussed in Section \ref{sec:Photog}, and is a fundamental for understanding the importance and necessity of the simulator. The proposed simulator is introduced in Section \ref{sec:sim}. The computer experiments for the evaluation and comparative study of three different variants of DIC algorithms are discussed in Section \ref{sec:exp}. The paper is concluded with the comparison of the techniques and recommendations.

\section{Photogrammetry}
\label{sec:Photog}
The process of determining geometrical properties of objects from images is known as Photogrammetry. Its use dates back to the mid 20th century with the advent of modern photography. A simple instance of photogrammetry is measuring the real world distance between two points from an image. For determining geometrical properties of an object from image(s), a relationship needs to be developed between the real world coordinate system and the image coordinate system. In simple cases, this relationship can be a simple scaling factor which provides appropriate accuracy level to measure the size of the object from images. In complex cases, this relationship can be a complex set of equations which accounts for different distortion factors associated with the camera position, optics used and atmospheric perturbations; these would often require additional calibration. Irrespective of the complexity level of the relationship, photogrammetry mainly depends upon four factors: resolution of the image or quality of the camera, size of the specimen or object, number of images used, and geometric layout of the images with respect to the object and each other \cite{White2003}.

Photogrammetry is used in different disciplines for analysis. For deformation monitoring, it was introduced in 1963 using X-ray images \cite{Roscoe1963}. Then, in the early 1970s, a stereo-photogrammetric technique was proposed which was later used to measure planar displacement of sand grains \cite{Andrawes1973,Butterfield1970}.

With the invention of the laser in the 1960s, Laser Doppler Velocimetry (LDV) was used to study two dimensional fluid flow \cite{Foreman1965}. An enhanced technique of Laser Speckle Velocimetry (LSV) was introduced in the 1970s to study fluid flows \cite{Barker1977}. In the 1980s, a photogrammetric based technique of Particle Image Velocimetry (PIV) was introduced to study the flow of fluids and soil by using images which was later improved by using digital images instead of analog images \cite{Adrian1986, White2003, Mohammad2001, White2005}.Around the same time, Digital Image Correlation (DIC) was introduced for deformation measurements in deformable solids \cite{Peters82}.

Both PIV and DIC techniques are contactless photogrammetric techniques which correlate a deformed image with a reference image. They help to monitor deformation and reconstruct displacement and strain fields of the loaded material \cite{Mccormick2012,Hall2010}. However, improvement in DIC was slow as compared to PIV because the reconstruction of the strain field in deformable solids requires very high accuracy.

\subsection{Setup for Photogrammetry}
\label{sec:setup}

Typical digital photogrammetric setup involves four elements: the specimen, light source(s), camera and computer. The light source needs to be stable, uniform and bright enough to enable recording intensity value for each pixel in an image which represents the features of an object under test. The specimen is required to be placed perpendicularly to the axis of focal length of the camera. The relative positions of the specimen and the camera must remain constant during the deformation process.

Normally, high resolution and good quality CCD or CMOS digital SLR cameras are used. High resolution cameras of 24-bits for true colors and 48-bits for deep color were expected to provide
better results \cite{Luo93}. However, experimental results show that the increase in the resolution levels has only minor effect on the accuracy due to uncontrollable experimental conditions such as camera vibrations, non-uniformity of light, image distortion, grey level variability in images, variable photometry or sensor response and motion out of plane of the image of the specimen \cite{Sutton08review,White2003}.

\begin{figure}[tb]
\centering
\includegraphics[bb=105 235 490 620, clip, width=3in]{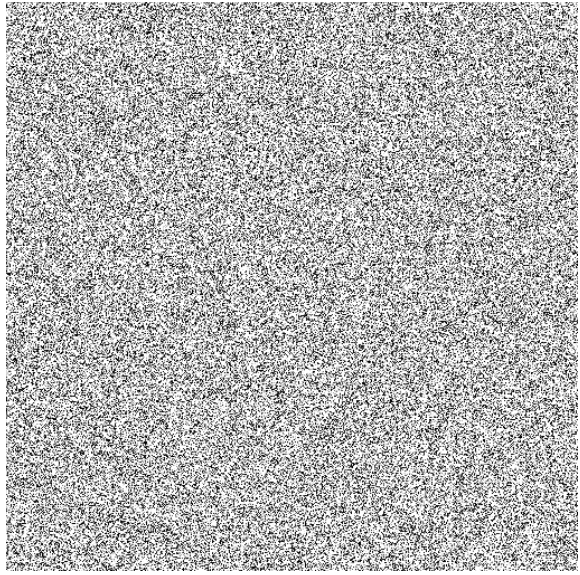}
\caption{An example of computer generated speckle pattern}
\label{fig:Speckle}
\end{figure}

In principle, in order to determine the displacement field, the after-deformation position of each pixel of the reference image has to be found and its displacement calculated. In order to uniquely identify the pixels of the image, a random pattern is required on the specimen surface, for example speckle of dots or spots, either pre-existing or painted on it. A possible method of painting the surface of the object is to cover it with white enamel and then spread unfiltered Xerox toner powder on the wet white paint. There are also several other ways to create a random pattern on the specimen \cite{Sutton08review, Chu85}. An  example of a computer generated speckle pattern is shown in Figure \ref{fig:Speckle}.

\subsection{Digital Image Correlation}
\label{sec:DIC}

The process of Digital Image Correlation involves a reference image representing a state of the object before deformation followed by capturing the image(s) after its deformation. Both reference and deformed images are two dimensional (2D) projections of the surface of the object. In some cases, multiple deformed images are analyzed together for detailed understanding of the deformation process \cite{Besnard2012}.

DIC presumes the displacement continuity which means that the neighboring pixels in the reference image remain neighboring in the deformed image. For subpixel accuracy, the intensities of both the reference and deformed images are interpolated before the DIC process \cite{Gilo11}. The interpolation process increases the size of the reference and deformed images and smoothes the intensity variation between the pixels. This helps in improving the accuracy of the determination of the subpixel displacement but increases the computation time. A high order b-spline interpolation technique is recommended to obtain good results \cite{Schreir00, Rohde09}.

To evaluate the correlation of images, both the reference and deformed images are divided into small segments using grids; each segment is known as a \emph{subset}. The size of the subset plays a critical role in the DIC method. For reliable results, the size of the subset should be large enough to have distinctive intensity variation for matching. On the other hand, the size of the subset needs to be small compared to the characteristic length of the strain field variations in order to achieve the necessary accuracy of the strain field reconstruction \cite{Pan09review,Liu10,Pan06}. Thus, the intensity variations in the image need to be high enough to allow having a small subset size.

In the DIC method, each subset of the reference image is matched with the subsets of the deformed image using the correlation criterion $C$, either maximizing Cross-Correlation (CC) or minimizing Sum of Squared Differences (SSD). An equivalent criterion to link both is: $C_{SSD}=2-2C_{CC}$ \cite{Pan07c}. It can be observed that the value of the cross-correlation criterion $C_{CC}$ increases with the increase in similarity and maximum match is represented by the value $1$, while the value of the sum of squared difference criterion $C_{SSD}$ decreases with the increase in similarity and maximum match is represented by the value $0$. Many variants of mathematical representation of $C$ have been proposed since the introduction of the DIC. Zero-Normalized Cross Correlation (ZNCC) or Zero-Normalized Sum of Squared Differences (ZNSSD) are highly recommended due to the property of being insensitive to fluctuation in illumination \cite{Pan09review,Tong05,Pan2009a}. ZNCC and ZNSSD are presented as:
\noindent
\begin{eqnarray}
C_{ZNCC} &= \sum\limits_{i=-M}^{M} \sum\limits_{j=-M}^{M} [ \frac{ \{ f(x_{i},y_{j})-f_{m}\} \{g(x_{i}^{'},y_{j}^{'})-g_{m} \} }{ \Delta f \Delta g} ] \nonumber \\
C_{ZNSSD} &=  \sum\limits_{i=-M}^{M} \sum\limits_{j=-M}^{M} [ \frac{  f(x_{i},y_{j})-f_{m}} {\Delta f}  - \frac{  g(x_{i}^{'},y_{j}^{'})-g_{m}} {\Delta g}  ]^{2}
\label{equ:DIC}
\end{eqnarray}
where $f(x_i,y_i)$ and $g(x_{i}^{'},y_{j}^{'})$ represent the gray intensity of $i$th pixel with coordinates $(x_i,y_i)$ and $(x_{i}^{'},y_{j}^{'})$ in the reference and deformed images respectively, and
\noindent
\begin{eqnarray}
f_m &=& \frac{1}{ (2M+1)^2 } \sum\limits_{i=-M}^{M} \sum\limits_{j=-M}^{M} f(x_i,y_j) \nonumber \\
g_m &=& \frac{1}{ (2M+1)^2 } \sum\limits_{i=-M}^{M} \sum\limits_{j=-M}^{M} g(x_i^{'},y_j^{'}) \\
\Delta f &=& \sqrt{ \sum\limits_{i=-M}^{M} \sum\limits_{j=-M}^{M} [ f(x_i,y_j) - f_m ]^2 } \nonumber \\
\Delta g &=& \sqrt{ \sum\limits_{i=-M}^{M} \sum\limits_{j=-M}^{M} [ g(x_i^{'},y_j^{'}) - g_m ]^2 } \nonumber
\label{equ:DICpara}
\end{eqnarray}

\begin{figure}[tb]
\centering
\includegraphics[bb=30 75 542 800, clip, width=0.8\textwidth,angle=-90]{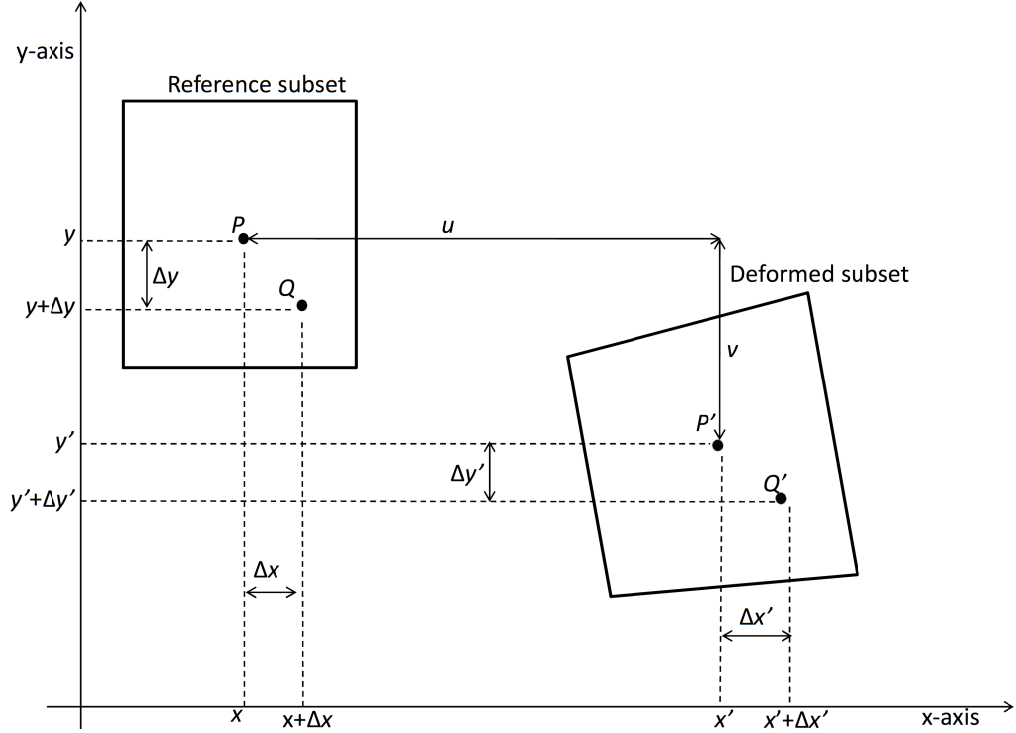}
\caption{Subset before and after deformation}
\label{fig:subset}
\end{figure}

The subset in the reference image is expected to be displaced as well as deformed, as illustrated in Figure \ref{fig:subset}. Here $P$ is the center of the subset and is represented by location $(x,y)$ in the reference image. This point moves to a new position $(x^{'},y^{'})$ in the deformed image and is represented by $P^{'}$ after deformation. Let $Q$ represent another point in the subset and its location in the reference image be $(x+\Delta x,y+\Delta y)$. After deformation, point $Q$ undergoes displacement $(u(Q),v(q))$ and is moved to $(x^{'}+\Delta x^{'},y^{'}+\Delta y^{'})$. We denote it by $Q^{'}$. The gray levels at points $P$, $P^{'}$, $Q$ and $Q^{'}$ are:
\begin{eqnarray} f(P) &=& f(x,y) \nonumber \\
f^{'}(P^{'}) &=& f^{'}(x^{'},y^{'}) = f^{'}[x+u(P),y+v(P)] \nonumber \\
f(Q) &=& f(x+\Delta x,y+\Delta y) \nonumber \\
f^{'}(Q^{'}) &=& f^{'}(x^{'}+\Delta x^{'},y^{'}+\Delta y^{'})
\label{equ:defint}
\end{eqnarray}
The last equation in eq.(\ref{equ:defint}) can be rewritten using displacement.
\begin{equation}
f^{'}(Q^{'}) = f^{'}(x+u(Q)+\Delta x,y+v(Q) + \Delta y)
\label{equ:defint2}
\end{equation}
Since, deformation is considered to be small, if there is a straight line $\overline{PQ}$ in the reference subset then it is approximated to remain straight in the deformed subset. By using Taylor
series expansion about point $P$ and keeping only linear terms, displacement in horizontal and vertical directions are obtained as:
\begin{eqnarray}
u(Q)&=& u(P) + \frac{\partial u(P)}{\partial x}\Delta x + \frac{\partial u(P)}{\partial y}\Delta y + ... \nonumber \\
v(Q)&=& v(P) + \frac{\partial v(P)}{\partial x}\Delta x + \frac{\partial v(P)}{\partial y}\Delta y + ...
\label{equ:defpt}
\end{eqnarray}
The higher order terms may be included if the deformation is expected to be more complicated and to improve the results as suggested in \cite{Lu00}. Substituting eq.(\ref{equ:defpt}) into eq.(\ref{equ:defint2}), we get
\begin{eqnarray}
f^{*}(Q^{'}) &=& f^{*}[ x + u(P) + \frac{\partial u(P)}{\partial x}\Delta x + \frac{\partial u(P)}{\partial y}\Delta y + \Delta x,\nonumber \\
& & y + v(P) + \frac{\partial v(P)}{\partial x}\Delta x + \frac{\partial v(P)}{\partial y}\Delta y + \Delta y ]
\label{equ:var6}
\end{eqnarray}

After substituting eq.(\ref{equ:var6}) into (\ref{equ:DIC}), it is seen that minimizing $C_{ZNSSD}$ or maximizing $C_{ZNCC}$ involves varying six variables: $u(P)$, $v(P)$, $\frac{\partial u(P)}{\partial x}$, $\frac{\partial u(P)}{\partial y}$, $\frac{\partial v(P)}{\partial x}$ and $\frac{\partial v(P)}{\partial y}$. If higher order terms are considered in eq.(\ref{equ:defpt}), then corresponding additional variables will also need to be varied \cite{Ma2012}. In the simplest case, only two variables, $u(P)$ and $v(P)$ are involved.

At the time of introduction of DIC for deformation monitoring by Peter et al \cite{Peters82}, the estimation of these six variables was obtained via the coarse-fine search algorithm. This method was applied and thoroughly analyzed in 1983 \cite{Peters83} but was found to be computationally expensive and have low accuracy. The errors in displacement were found to be around $10$ percent. The
coarse-fine search algorithm based DIC technique was improved in 2006 to reduce computation time by Zhang et al \cite{Zhang06} but accuracy level could not be significantly improved.

In 1989, a DIC algorithm which used the Newton-Raphson method to estimate the six unknown variables for maximizing cross-correlation between the reference and the deformed subset was proposed
\cite{Bruck89}. This method provided high accuracy but was computationally slow. The computation time of this method was improved in 1998 and 2011 by Vendroux \cite{Vendroux98} and Pan
\cite{Pan11fastDIC} respectively, without compromising accuracy. Similarly, other iterative numerical optimization techniques such as a Levenberg-Marquart Algorithm \cite{Schreier02} and a Quasi
Newton-Raphson Method \cite{Wang02} were proposed to overcome limitations of computational complexity posed by the Newton-Raphson Method but could not significantly improve it without compromising on
the accuracy.

In 1993, another approach was proposed by Chen which is based on the Peak-Finding Algorithm to estimate two variables  $u(P)$ and $v(P)$ instead of six \cite{Chen93}. In this algorithm, the subset of the reference image is cross-correlated with the subset of the deformed image at different pixels around the estimated deformed position. This process is known as displacement searching scheme and
provides the displacement estimation with accuracy of $\pm 1$ pixel. Afterwards, the local discrete correlation matrix near a pixel having the maximum cross-correlation coefficient is considered. For subpixel accuracy, the biparabolic least square fitting algorithm is used to fit the local discrete correlation matrix and the highest point is found by interpolation. This method proved to be computationally faster than others, however at the cost of accuracy.

In 2001, Artificial Neural Networks were used to find the maxima of the cross-correlation between subsets of the deformed and reference images \cite{Pitter01}. Later in 2003 and 2004 \cite{Shaopeng2003,Pilch2004}, Genetic Algorithms were also proposed to solve the same problem. While exhibiting high success in finding global maxima of cross-correlation, the Genetic Algorithms typically require long computation time.

As discussed above, many techniques have been used in DIC to find the maximum of cross-correlation. These techniques can be categorized as the Basic DIC and Extended DIC. The Basic DIC determines displacement field only, that is, it maximizes the cross-correlation by varying only two displacement components, $u(P)$ and $v(P)$. In other words the Basic DIC presumes that the displacement is uniform within a subset, which is a crude approximation.

The Extended DIC (the name is ours) presumes that the displacement field linearly changes within the subset, that is the displacement gradients (or strain and rotation) are constant within the subset. This involves estimating six variables- two displacement components, $u(P)$ and $v(P)$, and four displacement gradients, $\frac{\partial u(P)}{\partial x}$, $\frac{\partial u(P)}{\partial y}$, $\frac{\partial v(P)}{\partial x}$ and $\frac{\partial v(P)}{\partial y}$, which increases the accuracy at the expense of increasing the processing time.

While the reviewed literature discusses accuracy, the comparative analysis of the accuracy of displacement/strain reconstruction offered by different methods is still lacking. Due to the difficulty in setting up experiments, undertaking the required number of reproducible experiments, and controlling the experimental conditions, it is difficult to compare and analyze the accuracy of different variants of the DIC technique and their dependence on different parameters.

In order to progress the comparative analysis we developed a simulator which simulates the load testing experiment. It applies specified displacements to the speckle pattern and generates the photographic images. The selected algorithms used the images in order to determine which of them offers better accuracy of displacement/strain reconstruction.

The two most commonly used variants of DIC selected in this research from the above-mentioned categories of the Basic DIC and the Extended DIC are Peak-Finding Algorithm as proposed by Chen et al \cite{Chen93} and improved DIC using Newton-Raphson method as proposed by Vendroux and Knauss \cite{Vendroux98} respectively. Both variants of DIC are implemented in Matlab ver. R2013a for this study. Furthermore, the performance of commercially available software Vic-2d developed by Correlated Solutions Inc is also compared and analyzed.

In the Basic DIC, the Peak-Finding Algorithm is used to find the highest cross-correlation points in the deformed image. The cross-correlation coefficient is maximized by correlating the subset of
the deformed image with the subset of the reference image by varying the position of the center of the subset. It is presumed that the subset does not rotate in the process. The maximum cross-correlation coefficient provides pixel level accuracy. After obtaining the pixel level accuracy, the subpixel level accuracy is obtained by considering the area of the $3 \times 3$ matrix around the pixel where maximum cross-correlation coefficient is obtained. The second order polynomial is fitted to the selected $3 \times 3$ matrix and the peak point of cross correlation is obtained.

In the case of the Extended DIC, the process is finding the maximum point of correlation using Newton-Raphson iteration method. The termination criterion is based on the change in sum of all the six unknown variables and change in cross correlation coefficient. The values of $u$ and $v$ providing the highest cross-correlation coefficient value are the estimated new position of the point. The convergence of the Newton-Raphson method depends upon the initial guess \cite{Pan09review}.

\subsection{Limitations of DIC}
\label{sec:limitDIC}

The DIC technique has limitations. The main limitation comes from the assumption that the displacement field is continuous as observed from eqs.(\ref{equ:defint2}) and (\ref{equ:defpt}). Discontinuity can lead to large errors in determining the displacement. Usually, deformable solids such as rocks, wood and sometimes concrete possess pre-existing fractures or develop fractures before the ultimate failure. The displacement field around the fractures is discontinuous: the two neighboring points sitting on the fracture before deformation starts moving in different directions after the deformation. This makes the DIC recovering of displacements difficult and imprecise \cite{Poissant2010}.

From eq.(\ref{equ:DIC}), it can be observed that the cross-correlation requires subsets from the reference and deformed images. Each subset is square having dimensions $2M \times 2M$ and the point of focus is the center of the subset. Thus, the point of correlation needs to be $M$ pixels away from the sides of the image. Thus, the deformation on the borders of the image cannot be analyzed.

Another limitation of the DIC can be observed from eq.(\ref{equ:DIC}) which shows its dependence on a gray intensity variation or pattern on the specimen. The pattern is required to have good contrast and variation in a subset to uniquely identify each pixel in the image \cite{Lecompte2006,Hua2011}. For this reason, an artificial pattern is usually applied to the object at the stage of measurement instead of using the natural pattern of the object to improve the performance of DIC.

The performance and accuracy of DIC deteriorates significantly in the presence of noise in the images \cite{Ha09}. Thus, it is challenging to conduct measurements in dusty environment.

\subsection{Camera Errors}

Further studies indicated another source of errors; the errors due to self-heating of CCD and CMOS sensors of digital cameras \cite{Ma12}. It was reported that the temperature of the CCD and CMOS cameras increases due to heating up of the ICs and electronic circuit boards inside the camera. This self-heating may also affect the mechanical parts in the camera resulting in a slight change in geometry and subsequently the deformation of the image which can lead to the error determination of strain in the range of 74-227$\mu\varepsilon$. It is recommended that for higher accuracy, the cameras should be warmed up for 1 to 1.5 hours before use and the camera heating errors should be measured so that the final results could be corrected. Alternatively, cooled cameras may be used in DIC measurements.

Errors are also introduced due to image distortion which can be caused by one or several of the following sources \cite{White2001}:
\begin{itemize}
  \item The lens focal axis is not perpendicular to the plane of the object resulting in distorted image captured by the camera.
  \item The camera is placed correctly but is not firm at its position causing the camera shift and not isolated from vibrations distorting the image capturing process.
  \item For high speed applications, camera frame rate and camera motion mitigation causes distortion in the images \cite{Reu2008}.
  \item Lens optical distortions are not properly controlled (if for instance a cheap lens is used) \cite{Slama1980}.
  \item An object lying behind a viewing window is causing image distortion through refraction.
  \item The pixels of the CCD are presumed to be square but actually they are not square. This presumption causes errors in linear scaling of pixels to the real object.
\end{itemize}

It is recommended by White el al \cite{White2001} to use an eighteen parameter deformation measurement system to avoid these image distortions which provides precision of $1/15000^{th}$ of the field of view in an image.

\section{The DISTRESS Simulator}
\label{sec:sim}
Measuring the accuracy of DIC reliably is a critical problem. Usually, it is found in the literature that the accuracy of DIC is tested by laboratory experiments. The strain gauges are affixed
on the surface of the specimen before the specimen is loaded. After loading the specimen, the measurements are obtained by the strain gauges. For DIC, images are also captured before and after loading the specimen. DIC is used to find the displacement and then the results are matched with the measurements obtained by the strain gauges \cite{Sutton83, Bruck89, Luo93, KJetter90, Pan11fastDIC,Vendroux98, Haddadi08}.

However, strain gauges cannot be used to monitor the strain field at all points on the specimen surface. If used together with the photogrammetry, the strain gauges obstruct the view exactly at the points where the comparison is intended. Furthermore, the strain gauges having a finite size do not measure the strain field as such, only its average value over the strain gauge length is recorded. It is also difficult to accurately replicate the lab experiments due to the sensitivity to many parameters and it is not possible to reconstruct accurate and precise displacement and strain fields.

In a few studies, synthetic images are used to measure the accuracy of DIC \cite{Rodriguez2012, Chu85, Rodriquez2011}. The reference synthetic image is either taken from the lab experiments or from
earlier research data. Afterwards, a known fixed displacement is introduced in the entire reference image and DIC is used to recover that displacement. Also, the introduced displacements are in terms of pixels rather than in subpixels. Recently, a study \cite{Tomicevc2013} is proposed to circumvent this problem by regularizing DIC using a priori information available about the material of the specimen.

Thus, a reliable and fast process is required to measure the subpixel level accuracy of the DIC method of reconstruction of realistic non-uniform displacements. It also needs to be independent of the obstructions from the strain gauges and other experimental errors.

To fulfil this gap of measuring subpixel level accuracy, a Digital Image Synthesizer to Reconstruct Strain in Solids (DISTRESS) Simulator is developed which simulates the displacement field developed under a specified load. It consists of two main parts where the first part simulates the load-induced displacement, while the second part simulates the image capturing process.

In the load-induced displacement part of the simulator, the specimen surface is digitally covered with random speckle pattern. The size of speckles and average space among speckles are input parameters to the simulator. The entire surface of the specimen is divided into small grids depending on the average spacing among speckles. Then, one speckle is placed in each grid and the position of the center of the speckle is randomly selected within the boundaries of the grid. This simulates a random distribution of the speckles on the surface of the specimen. Thus, the final speckle pattern is unique at each point of the specimen's surface.

The generated speckles are allowed to be overlapping as often happen in the real world scenario. The speckles are considered to be hard particles which do not change their shape during loading process. Barranger et al \cite{Barranger12} mentioned in their analysis that difference in the performance of DIC between the rigid and the soft particles used as the speckles on the surface of the specimen is negligible and does not affect the deformation measurements .

The speckle size, average spacing among speckles and dimensions of the specimen are introduced as fractions of the length of the specimen, that is the length of the specimen is assumed to be $L=1$. The equations for displacement $u(x,y)$ and $v(x,y)$ are provided to the load-induced displacement part of the simulator.

In the image capturing part of the simulator, the dimensions and resolution of the images are specified. Images of the specimen surface before and after the deformation are simulated. This part of simulator is designed to maintain the high quality and precision of images irrespective of their dimensions. The intensity of each pixel of the image captured by the simulator is represented by $256$ gray levels. The intensity level of the pixels partially covered by a speckle is taken as proportional to the area of the pixel covered by the speckle. If a pixel is fully covered by a speckle, it will be assigned the highest intensity level. If the pixel is at the edge of the speckle, then speckle is expected to cut the pixel. Then the intensity is reduced by the factor and the percentage of the area of the pixel covered by the speckle is calculated.

\begin{figure}[!tb]
\centering
\includegraphics[bb=1 10 250 500, clip, width=3in]{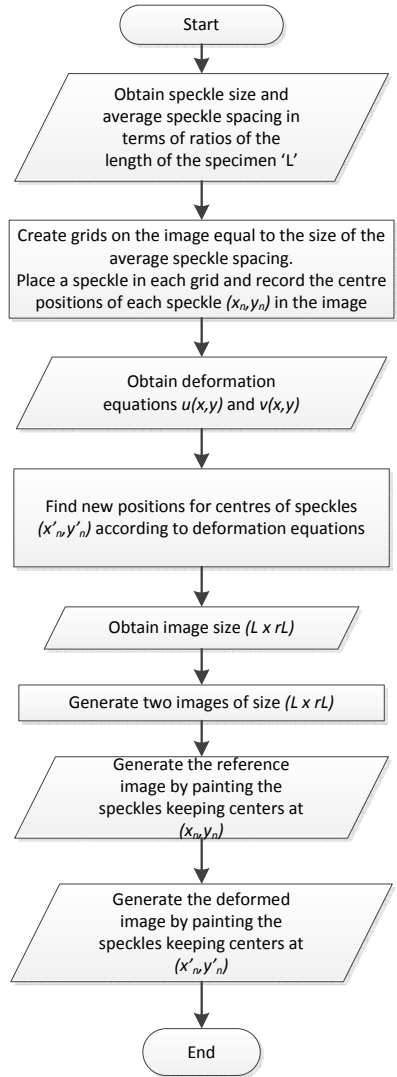}
\caption{Flowchart of the DISTRESS Simulator}
\label{fig:flowchart}
\end{figure}

Figure \ref{fig:flowchart} shows simple flowchart to illustrate the functionality of the proposed simulator. The simulator successfully simulates the real world scenario and images obtained are accurately depicting the loading experiments. The images provided by the simulator are without the experimental and equipment errors which are usually introduced during the practical lab experiments.

\section{Results and Discussion}
\label{sec:exp}
Using the simulator explained in the previous section, the performance and accuracy of the Basic DIC, Extended DIC and the commercial DIC software VIC-2d are compared. The analysis of all three followed the same standard processes mentioned in Section \ref{sec:Photog}

For the Extended DIC, the iteration process of correlation maximization using the Newton Raphson method is continued till the changes in the sum of all the displacements and displacement gradients, and change in the cross correlation coefficient become smaller than $ 0.5 \times 10^{-8} $ and $ 1 \times 10^{-8} $ respectively. The maximum number of iterations is limited to $40$ to avoid indefinite loops. The first initial guess for the displacements and their gradients for the first point of the image is obtained by the pixel level cross correlation of subset from the reference image to all possible subsets of deformed image. For the subsequent initial guesses for the cross correlation optimization for the other points of the reference image, the final result of the nearest pixel, for which displacement result is already obtained, is used.

During the numerical experiments, it is observed that the maximum number of iterations is never reached. Few worst case experiments are designed to achieve the maximum number of iterations in which a subset of image not related to reference image is cross correlated with reference image. In this cases the iteration process reaches its maximum number of iterations without meeting other termination criteria.

\begin{figure}[tb]
\centering
\includegraphics[bb=-60 260 662 690,clip, width=\textwidth]{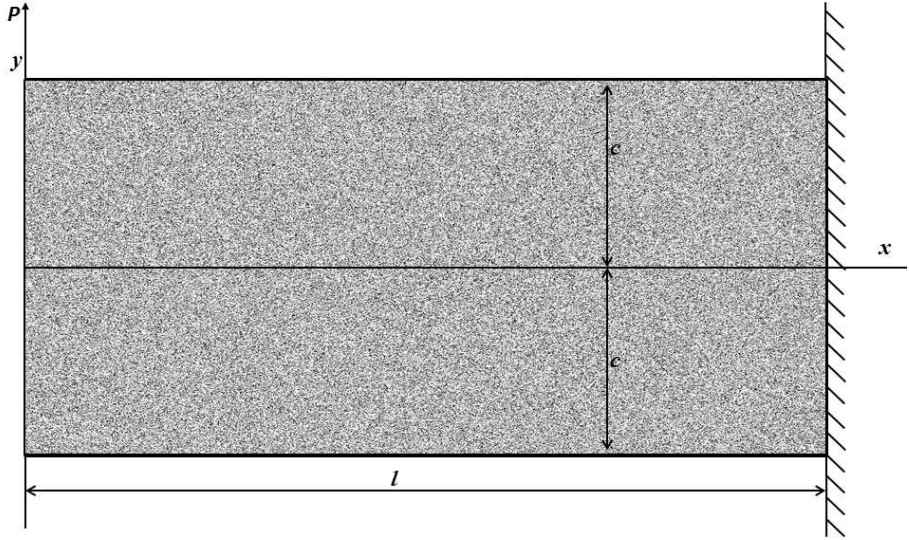}
\caption{Bending of a cantilever loaded at the end}
\label{fig:UVdisp}
\end{figure}

Figure \ref{fig:UVdisp} presents the displacement field produced by bending of a cantilever loaded at an end, as a benchmark. The corresponding equations for the displacements read (e.g., \cite{BookTimoshenko})
\begin{eqnarray}
u &=& \frac{P y}{6 I} [ \frac{1}{E} \{ 3(L^{2} - x^{2}) - \nu y^{2} \} + \frac{1}{G} ( y^{2}-3 c^{2}) ] \nonumber \\
v &=& \frac{P}{6 E I} [ x (3 \nu y^{2} + x^{2}) + L^{2}( 2 L - 3 x) ]
\label{equ:UVdisplacement}
\end{eqnarray}
where  $P$ is the applied force, $E$ is the Young's modulus, $I$ is the moment of inertia of cross section of the cantilever, $\nu$ is the Poisson's ratio, $G$ is the shear modulus, $2c$ is the width of the cantilever and $L$ is the length. For the numerical experiments, material of Aluminium is considered and we used the following values, $P=4.8 N$, $E=69 GPa$, $I=175 mm^4$, $\nu=0.334$, $G=26 GPa$, $c=25 mm$ and $L=110 mm$. The maximum displacements are $u_{max}=0.28 L$ and $v_{max}=0.5 L$.

\begin{figure}[!tb]
\begin{center}
\includegraphics[bb=-190 55 795 785, clip, width=4in, angle=0]{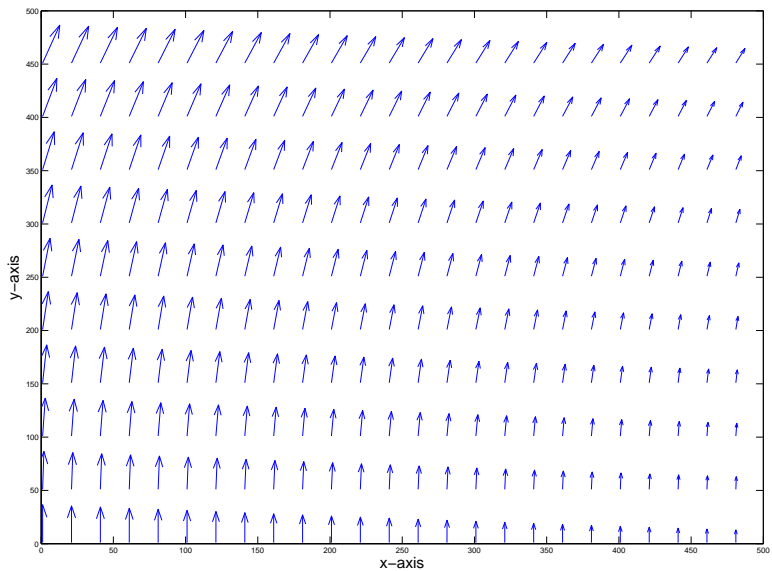}
\end{center}
\caption{Displacement vectors for image size of $500 \times 500$ pixels}
\label{fig:disp}
\end{figure}

The displacements for each point of the reference image are generated by DISTRESS Simulator. The result of displacement field is shown in Figure \ref{fig:disp} for an image size of $500 \times 500$ pixels.

In the initial experiments, the speckles size and average speckle spacing were varied over the range $0.005 L$ to $0.02 L$. The analysis of the initial experiments showed that the minimum subset size needs to be greater than the size of a speckle and average speckle spacing. With the random placement of speckles in the grid of an image having size equal to the average speckle spacing, the maximum and minimum distance between center of speckles can be twice the average speckle spacing and one pixel respectively. Furthermore, the image size also effects the minimum subset size due to increase in the number of pixels covered by a speckle with the increase in the image size. Keeping in mind these limitations, a relationship between the speckle size and average speckle spacing is proposed to determine the minimum subset size required for DIC.
\begin{eqnarray}
ss_{min} = \max(i_{d}) [2 r_a - r_d] & for & r_a \sim r_d
\label{equ:ss}
\end{eqnarray}
where $ss_{min}$ is the minimum recommended size of the subset to have sufficient intensity variations for successful DIC, $i_{d}$ is the image size, $r_a$ is the ratio of the average speckle spacing (distance between the speckle centres)  and the image size and $r_d$ is the ratio of the speckle size and the image size.

Eq. (\ref{equ:ss}) suggests that increase in $r_a$ or decrease in $r_d$ causes speckles to be far apart from each other, while decrease in $r_a$ or increase in $r_d$ causes the speckles to have higher overlap with each other. We keep $r_a$ and $r_d$ close to each other. It also suggests that DIC may fail at some points on the surface of the specimen because either subset covers the speckles only or space between the speckles. This causes no intensity variations in the subset. It has also been found that the increase of the subset size or decrease in $r_a$ or $r_d$ will increase the intensity variations in the subset. In this study, both $r_a$ and $r_d$ are taken as $0.01$ which restricts the $ss_{min}$ to $21 \times 21$ pixels for the image resolution of $ 2000 \times 2000$ pixels.

Three sets of images having dimensions of $ 500 \times 500$, $ 1000 \times 1000$ and $ 2000 \times 2000$ pixels are obtained from DISTRESS Simulator. For each set of images,
different subset sizes ranging from $21 \times 21$ to $101 \times 101$ with an increment of $10 \times 10$ pixels are used. Each subset is a square and is represented in the plots by a size of its side. For instance, subset size of $21 \times 21$ pixels is represented by $21$.

\begin{figure}[!tb]
\begin{center}

\subfigure{
\label{fig:PF500}
\includegraphics[bb=105 270 470 570, clip, width=3in, angle=0]{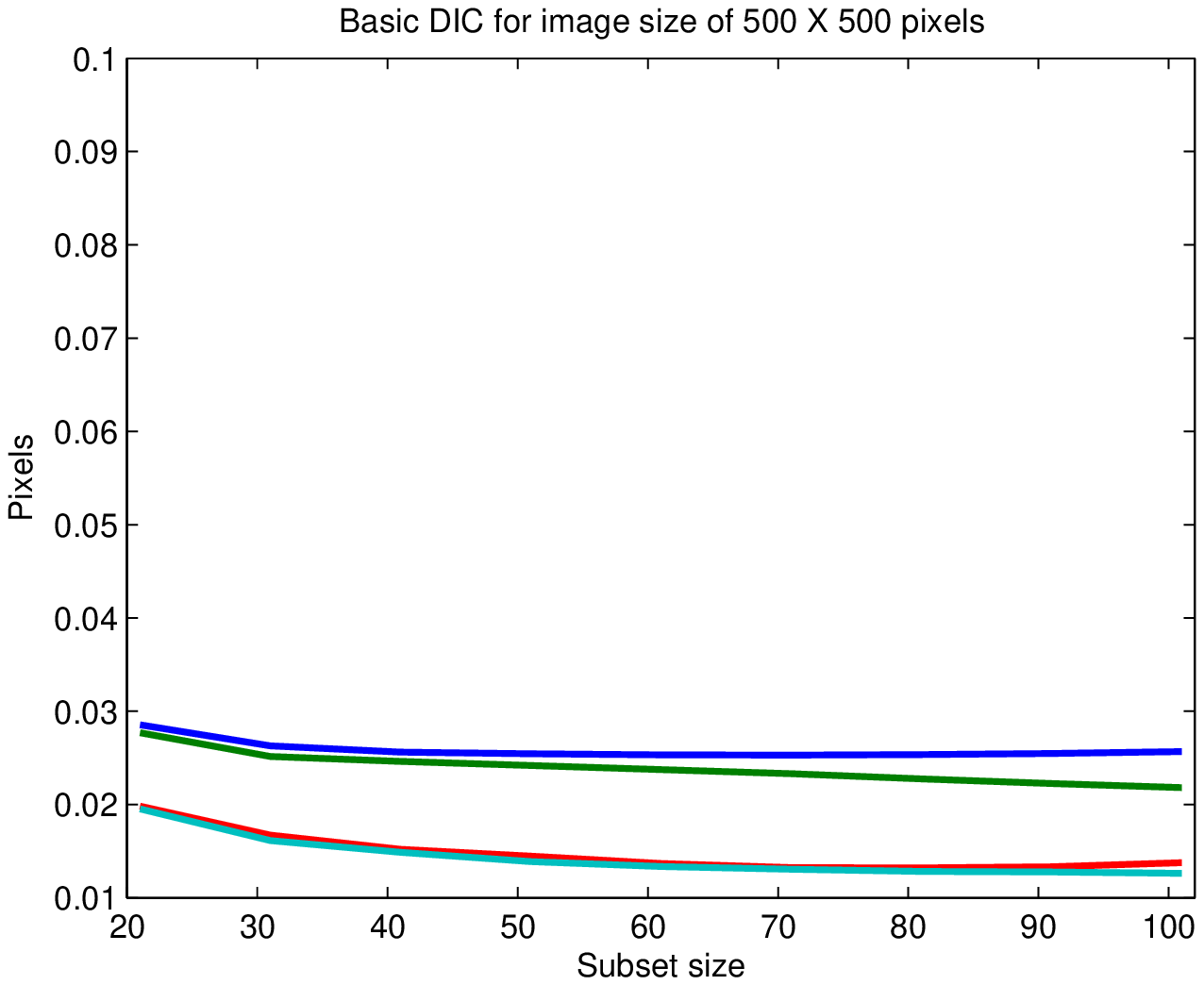}
}\\

\subfigure{
\label{fig:PF1000}
\includegraphics[bb=105 270 470 570, clip, width=3in, angle=0]{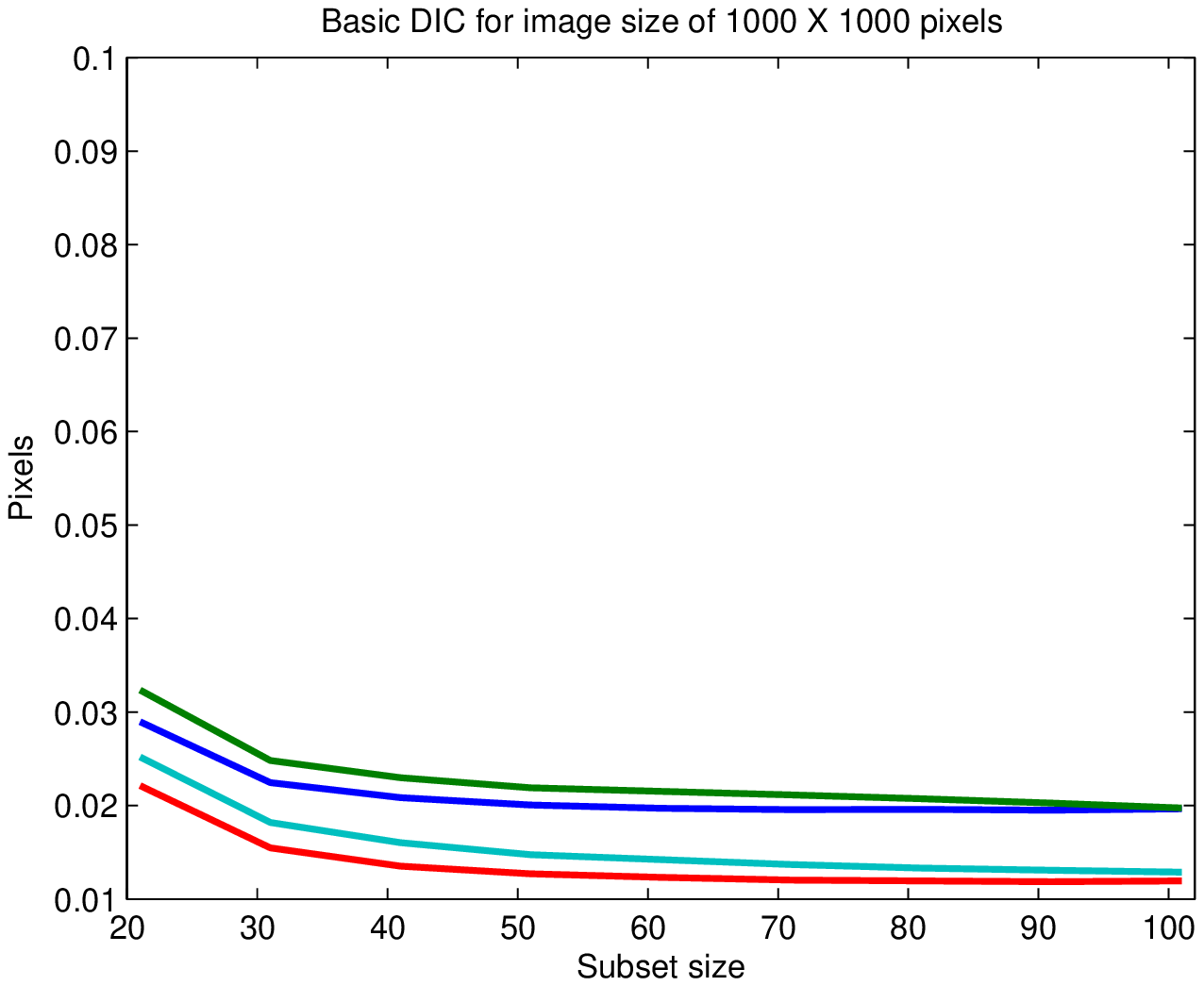}
}\\

\subfigure{
\label{fig:PF2000}
\includegraphics[bb=105 245 470 610, clip, width=3in, angle=0]{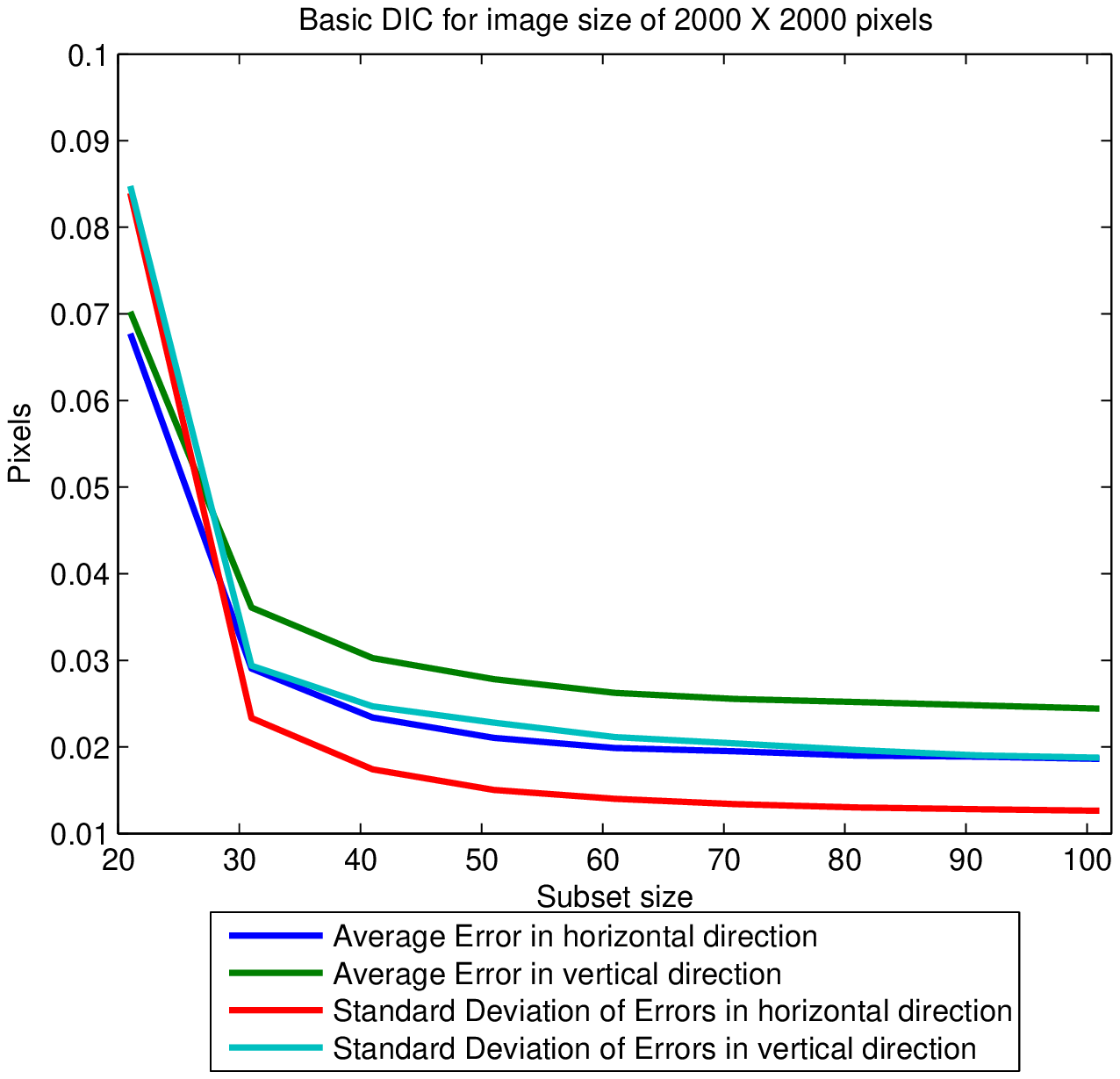}
}
\end{center}
\caption{Average Errors and Standard Deviations for the Basic DIC}
\label{fig:PF}
\end{figure}

\begin{figure}[!tb]
\begin{center}

\subfigure{
\label{fig:NR500}
\includegraphics[bb=100 270 470 570, clip, width=3in, angle=0]{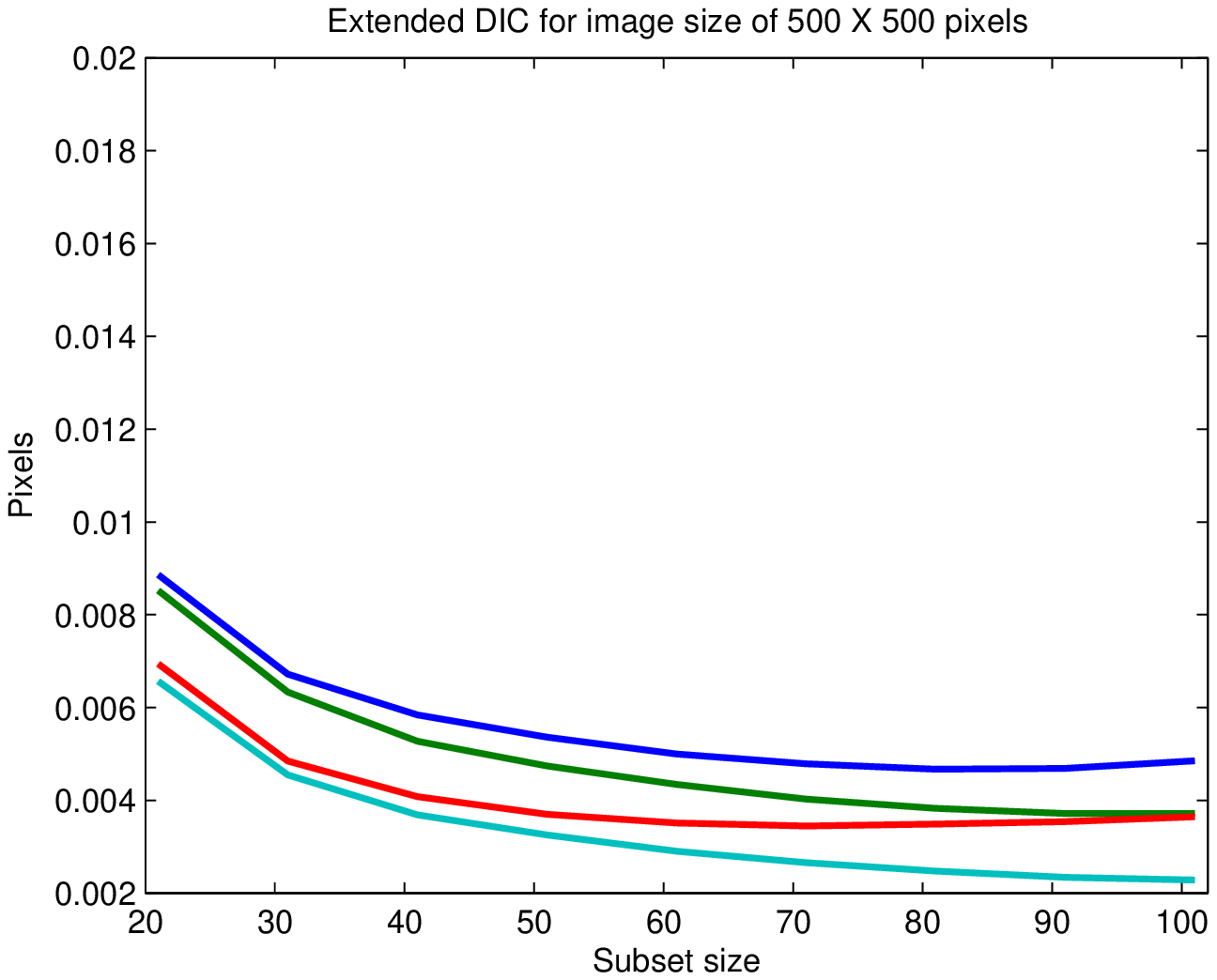}
}\\

\subfigure{
\label{fig:NR1000}
\includegraphics[bb=100 270 475 570, clip, width=3in, angle=0]{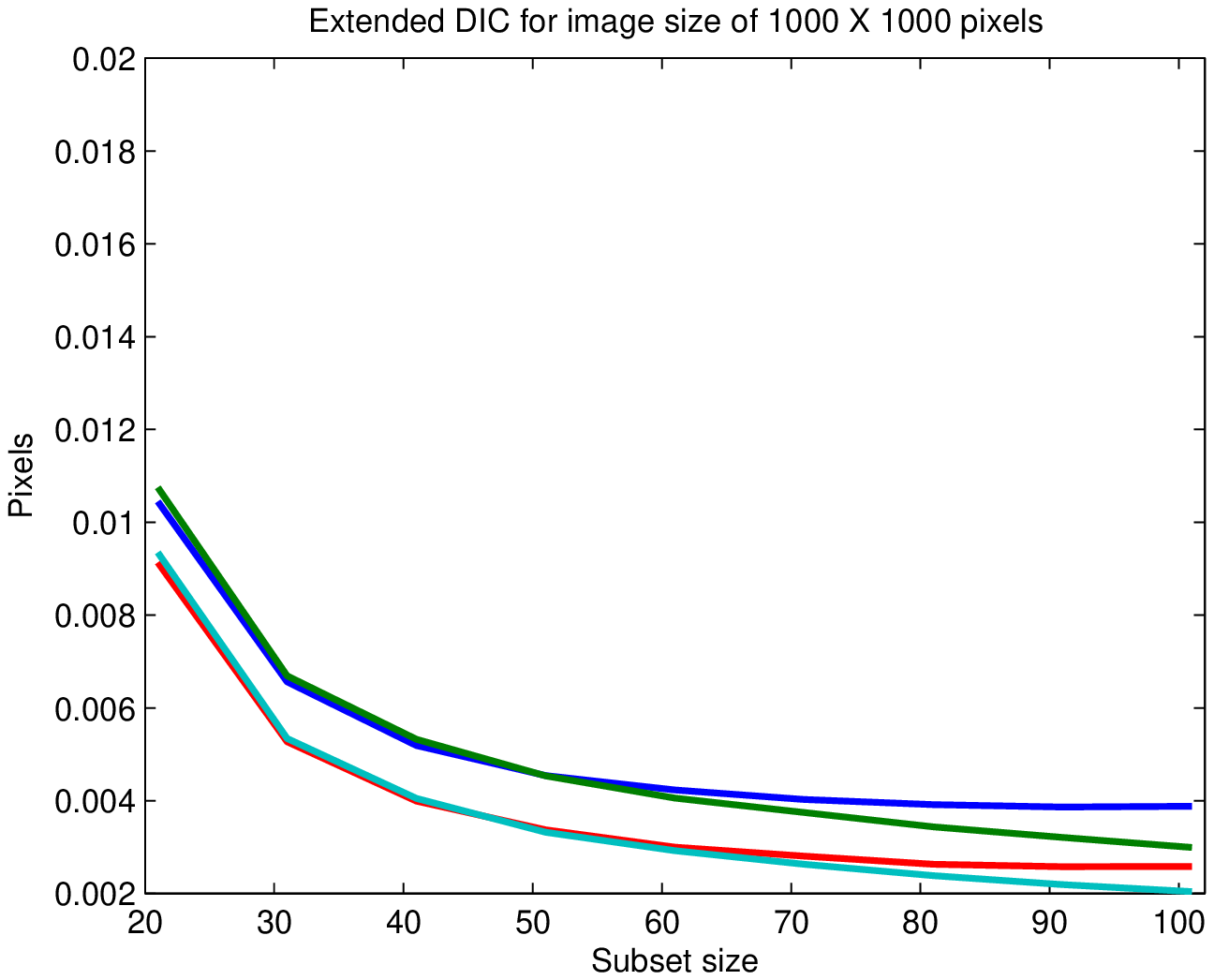}
}\\

\subfigure{
\label{fig:NR2000}
\includegraphics[bb=100 245 475 610, clip, width=3in, angle=0]{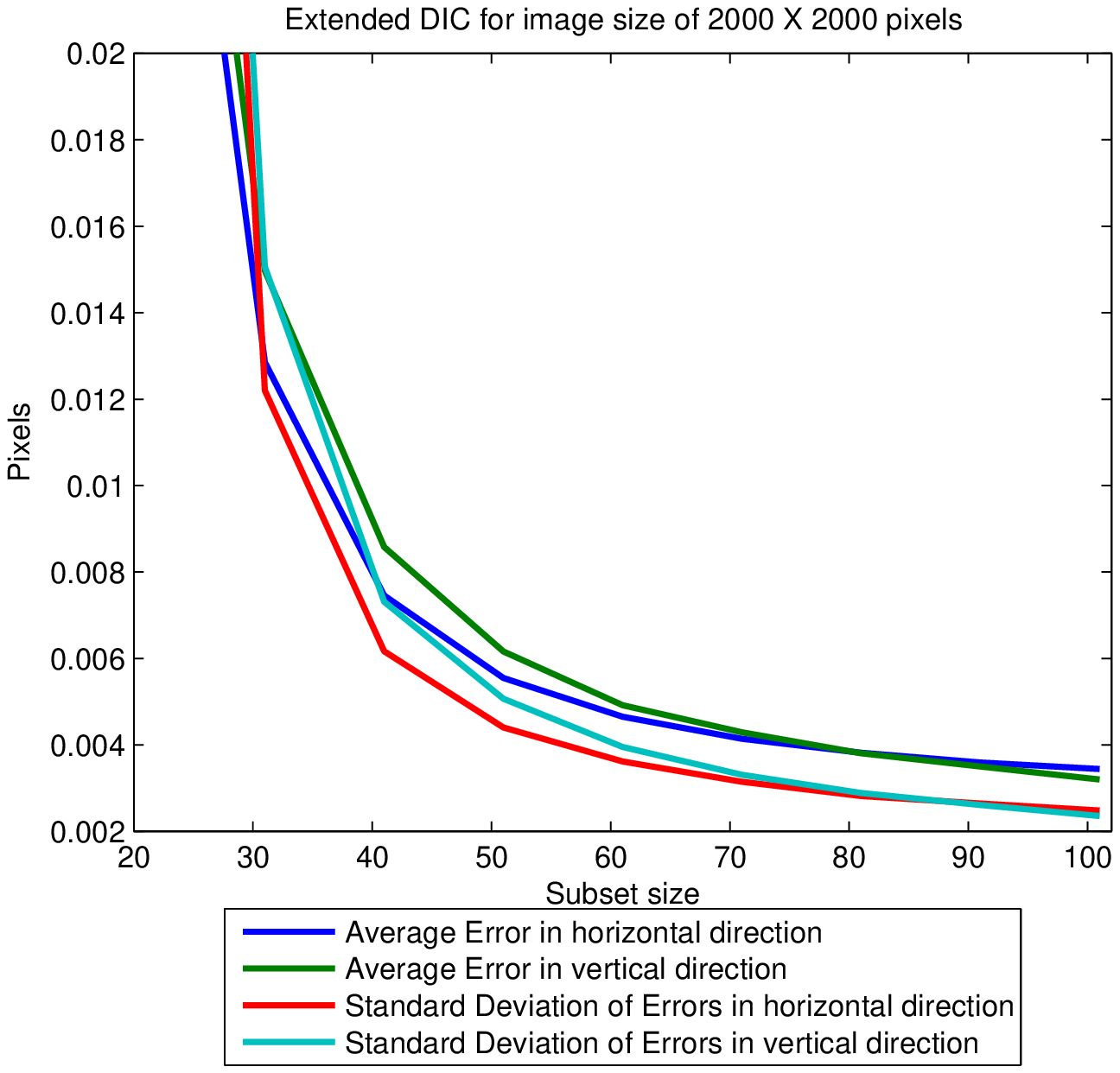}
}
\end{center}
\caption{Average Errors and Standard Deviations for the Extended DIC}
\label{fig:NR}
\end{figure}

\begin{figure}[!tb]
\begin{center}

\subfigure{
\label{fig:VIC500}
\includegraphics[bb=100 270 470 570, clip, width=3in, angle=0]{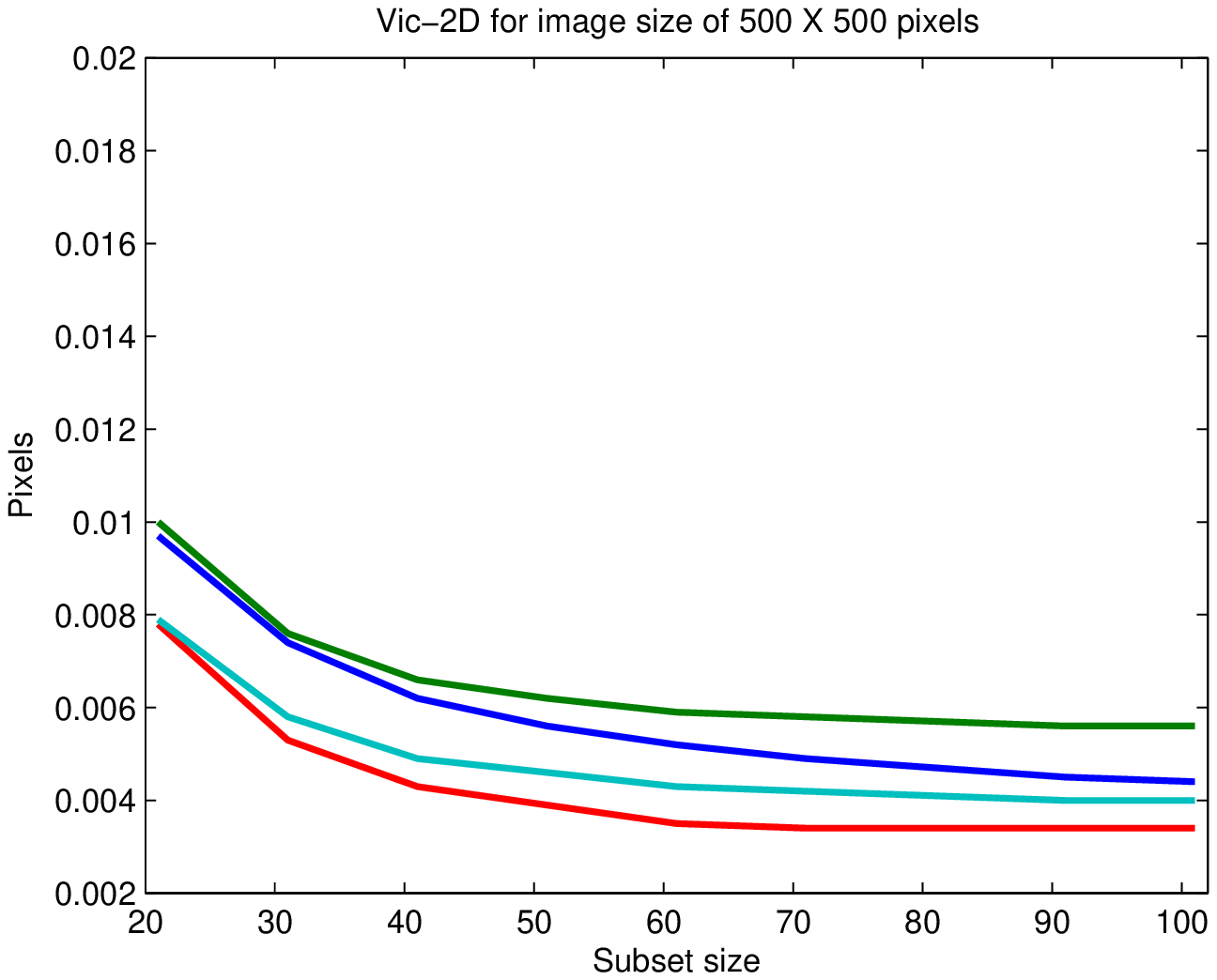}
}\\

\subfigure{
\label{fig:VIC1000}
\includegraphics[bb=100 270 470 570, clip, width=3in, angle=0]{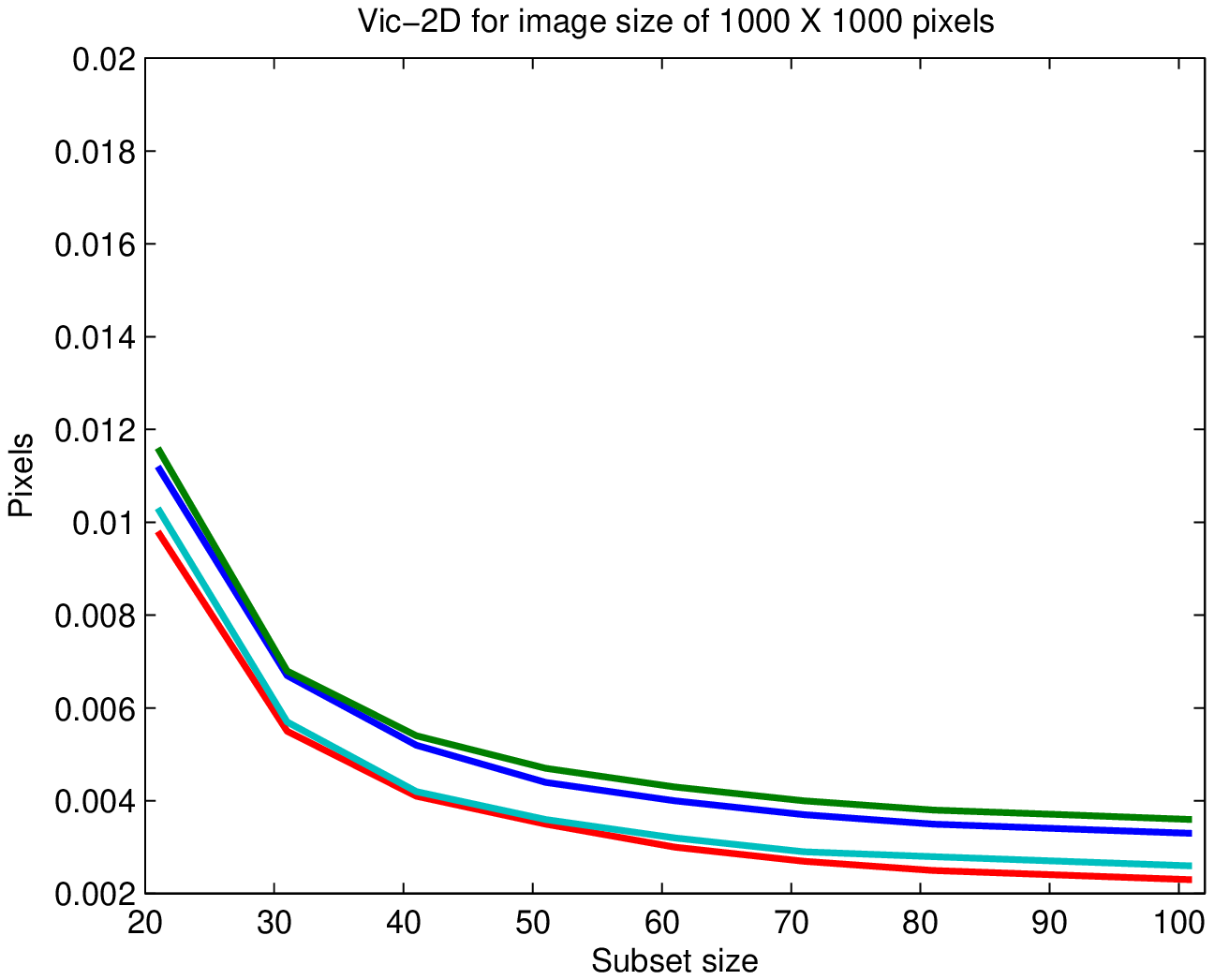}
}\\

\subfigure{
\label{fig:VIC2000}
\includegraphics[bb=100 245 470 610, clip, width=3in, angle=0]{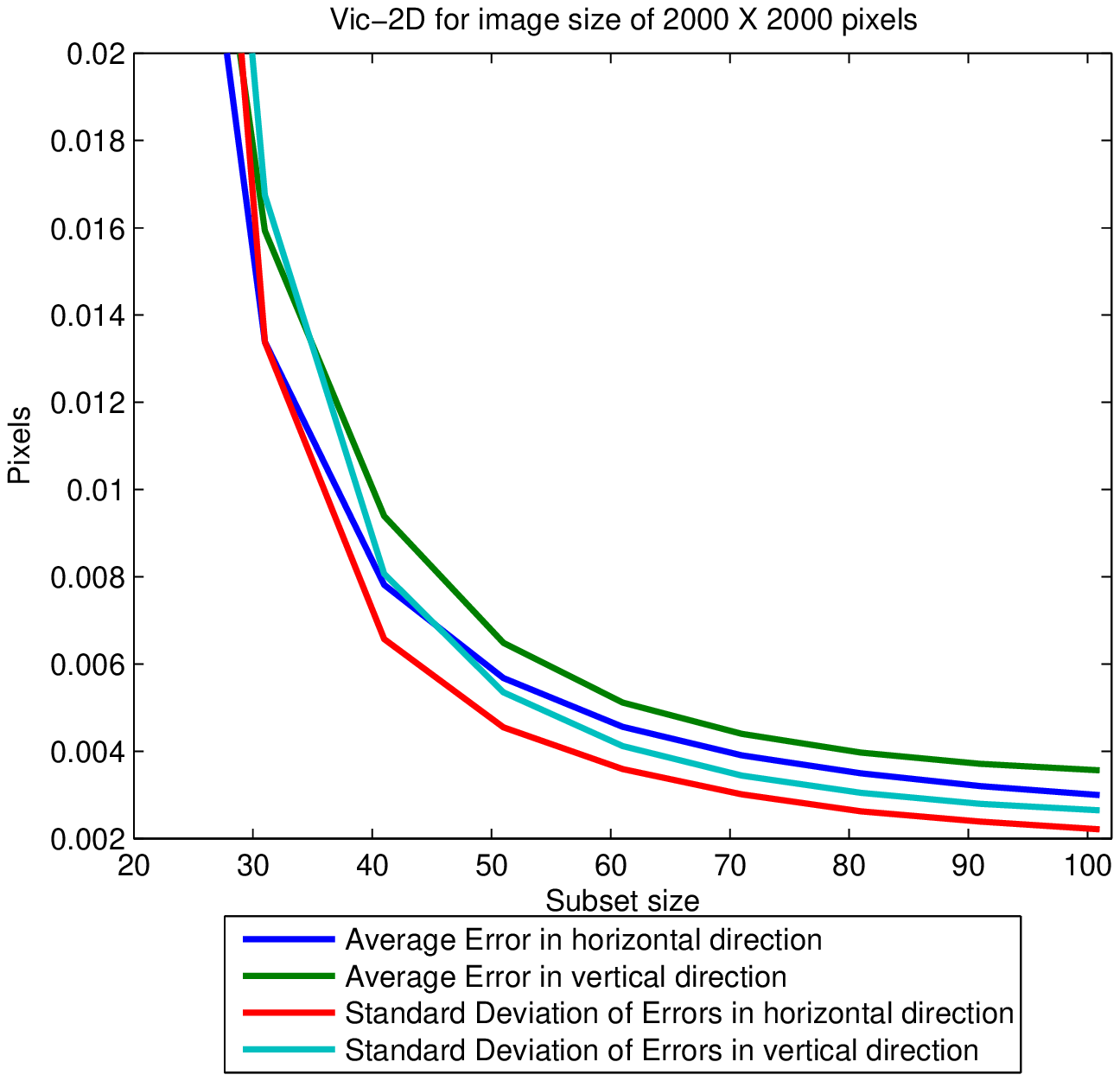}
}
\end{center}
\caption{Average Errors and Standard Deviations for the Vic-2D software}
\label{fig:VIC}
\end{figure}

The results obtained by the Basic DIC are presented in Figure \ref{fig:PF}. Similarly, the results obtained by the Extended DIC are plotted in Figure \ref{fig:NR} and results of Vic-2d are presented in Figure \ref{fig:VIC}. Figures \ref{fig:PF}, \ref{fig:NR} and \ref{fig:VIC} present the average errors and their standard deviations in displacements in horizontal $u$ and vertical $v$ directions for image sizes of $500 \times 500$, $1000 \times 1000$ and $2000 \times 2000$ pixels.

The average error represents the absolute difference in pixels between the displacements recovered by the particular variant of the DIC and the original displacements. The standard deviation of both horizontal $u$ and vertical $v$ errors represent the consistency of accuracy of the technique.

\begin{figure}[!tb]
\begin{center}

\subfigure{
\label{fig:PF_E2E500}
\includegraphics[bb=100 270 470 570, clip, width=3in, angle=0]{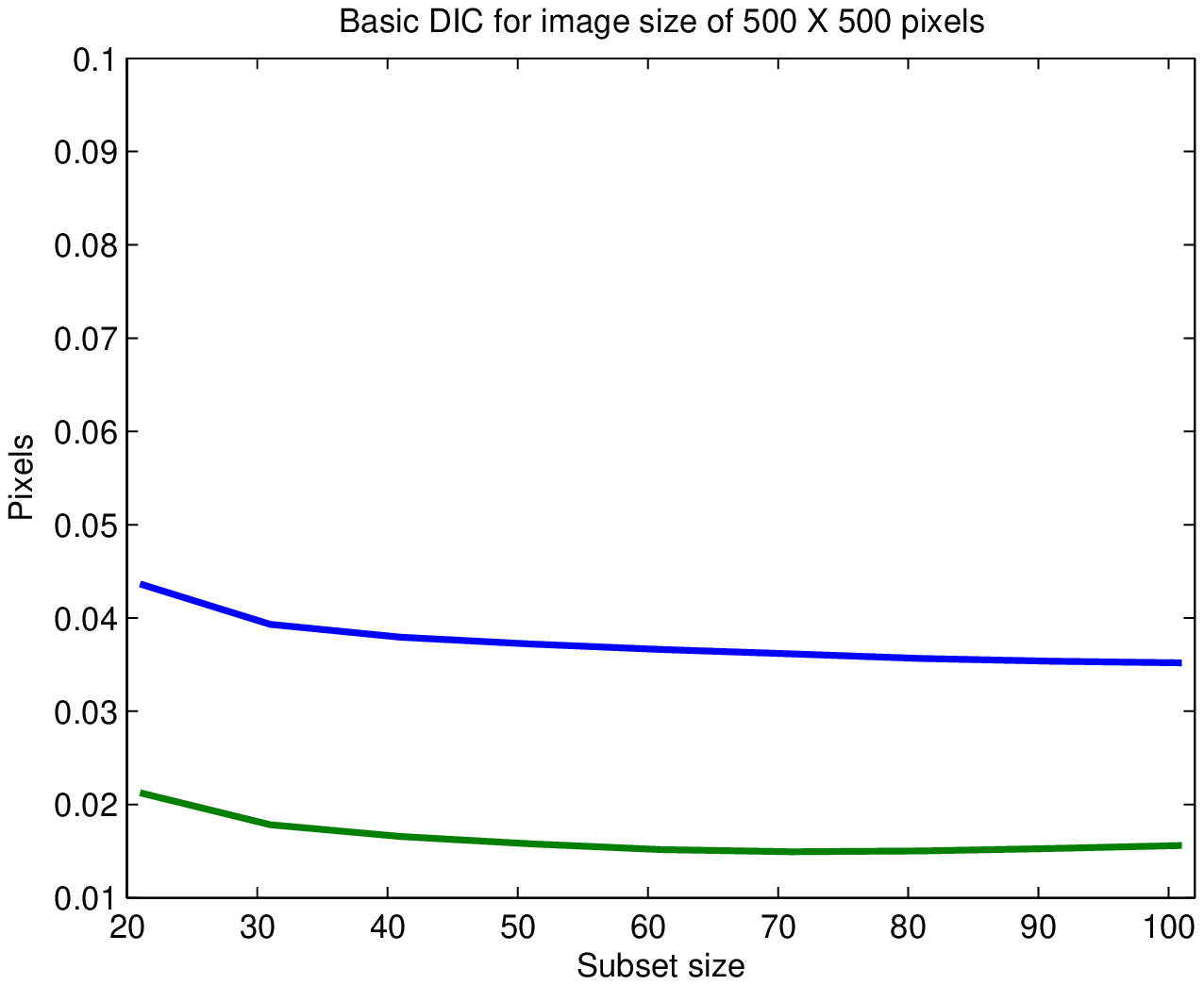}
}\\

\subfigure{
\label{fig:PF_E2E1000}
\includegraphics[bb=100 270 470 570, clip, width=3in, angle=0]{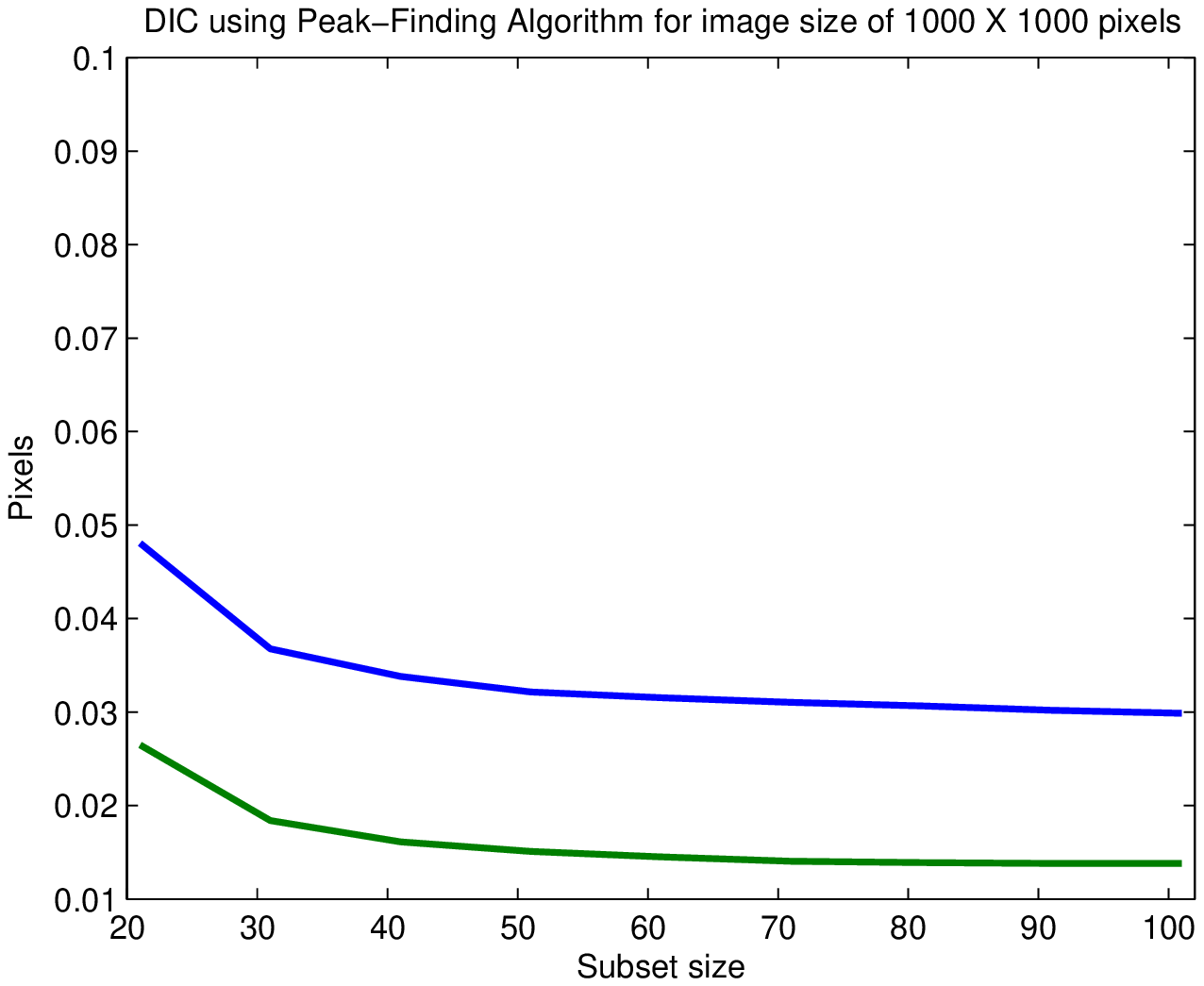}
}\\

\subfigure{
\label{fig:PF_E2E2000}
\includegraphics[bb=100 260 470 595, clip, width=3in, angle=0]{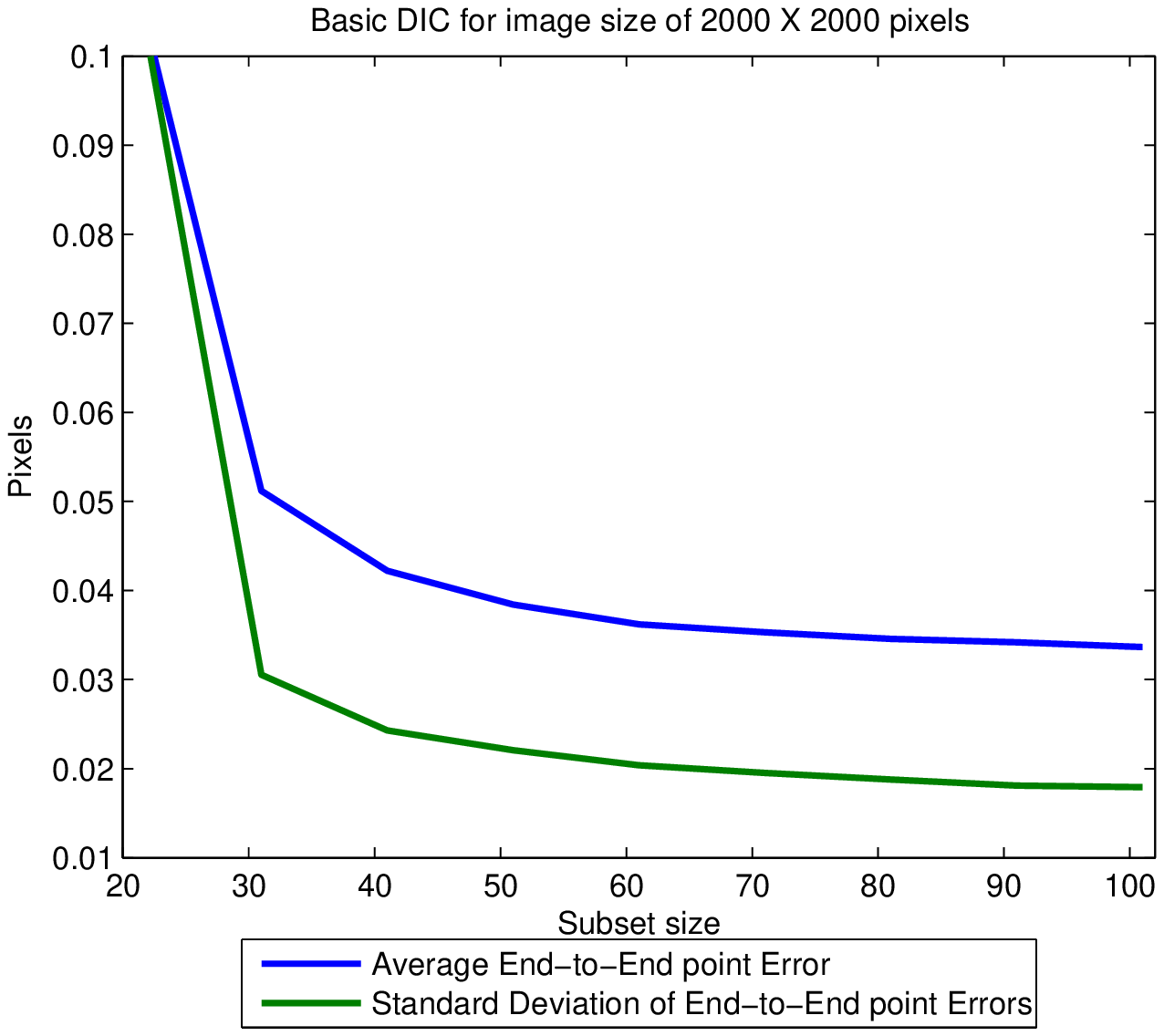}
}
\end{center}
\caption{Average End-to-End point errors and Standard Deviations for the Basic DIC}
\label{fig:PF_E2E}
\end{figure}

\begin{figure}[!tb]
\begin{center}

\subfigure{
\label{fig:NR_E2E500}
\includegraphics[bb=100 270 470 570, clip, width=3in, angle=0]{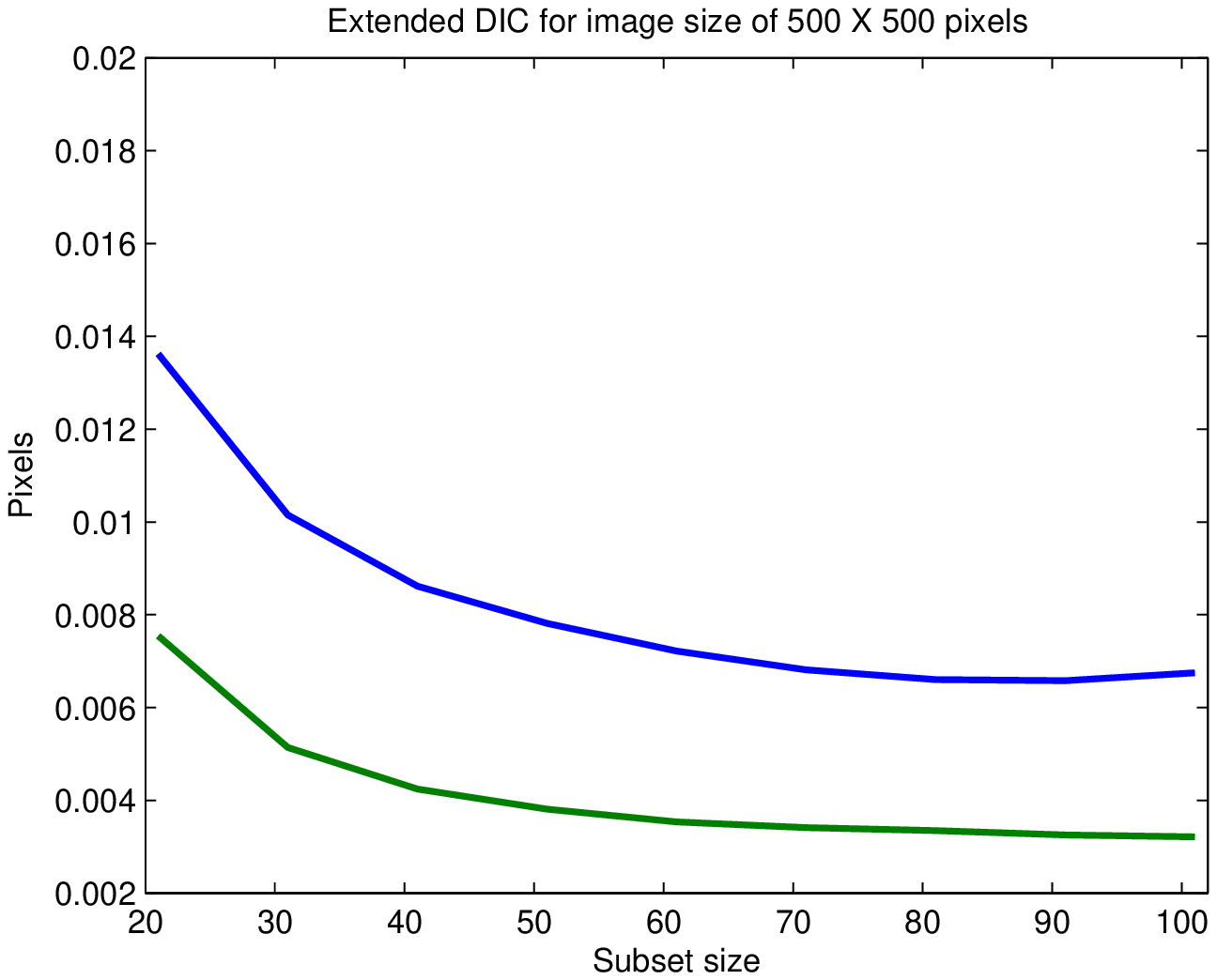}
}\\

\subfigure{
\label{fig:NR_E2E1000}
\includegraphics[bb=100 270 475 570, clip, width=3in, angle=0]{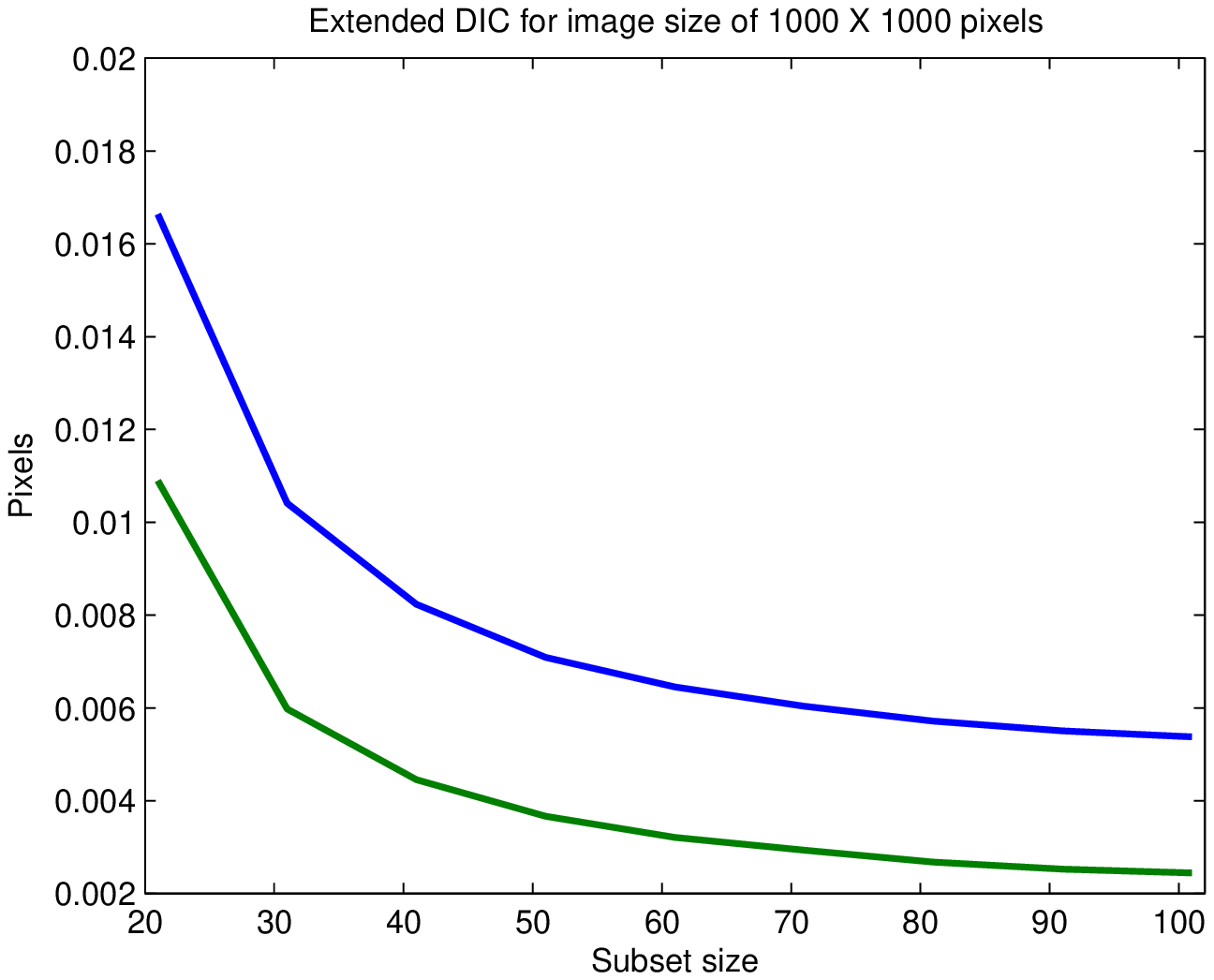}
}\\

\subfigure{
\label{fig:NR_E2E2000}
\includegraphics[bb=100 260 475 595, clip, width=3in, angle=0]{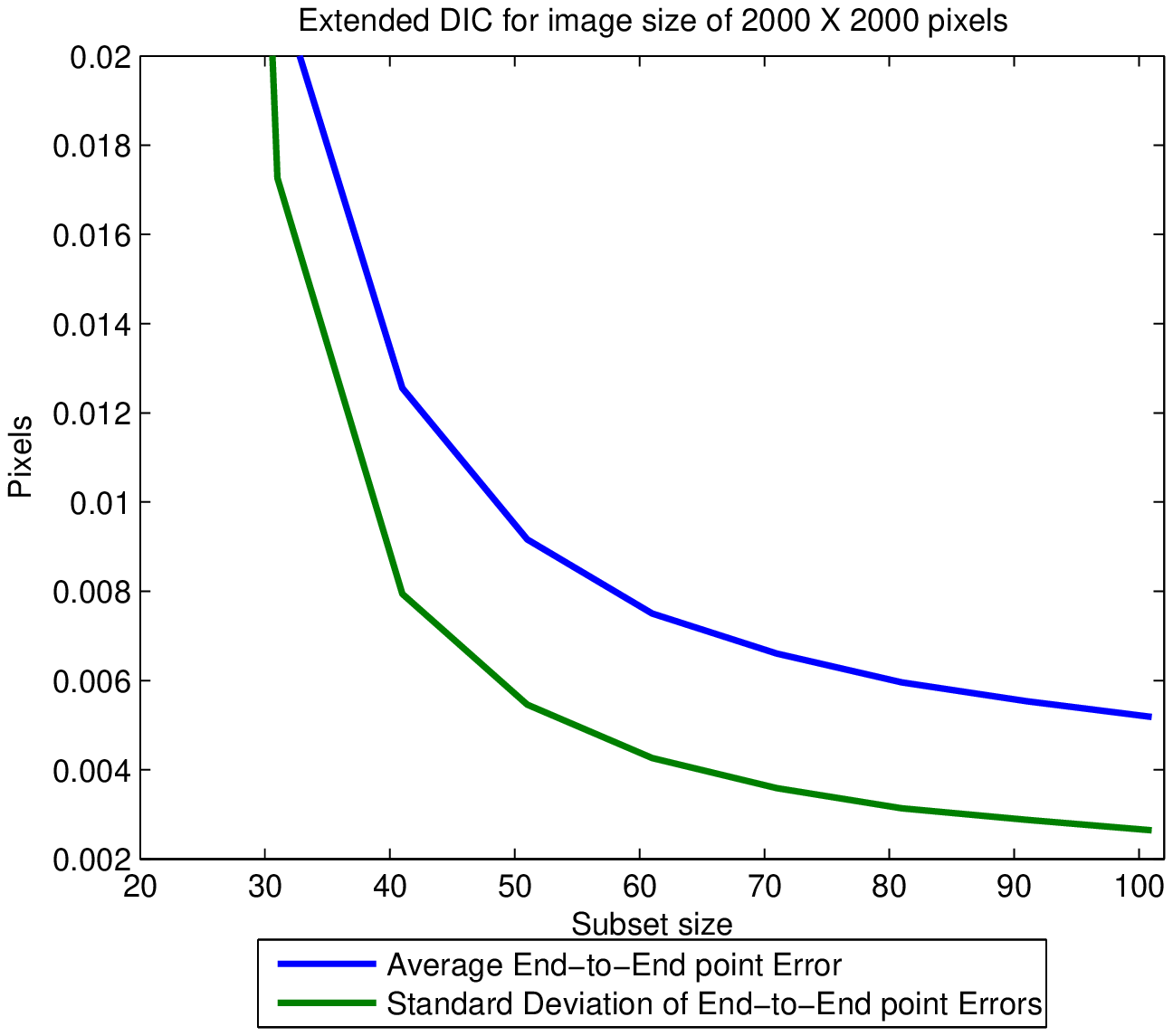}
}
\end{center}
\caption{Average End-to-End point errors and Standard Deviations for the Extended DIC}
\label{fig:NR_E2E}
\end{figure}

\begin{figure}[!tb]
\begin{center}

\subfigure{
\label{fig:VIC_E2E500}
\includegraphics[bb=100 270 470 570, clip, width=3in, angle=0]{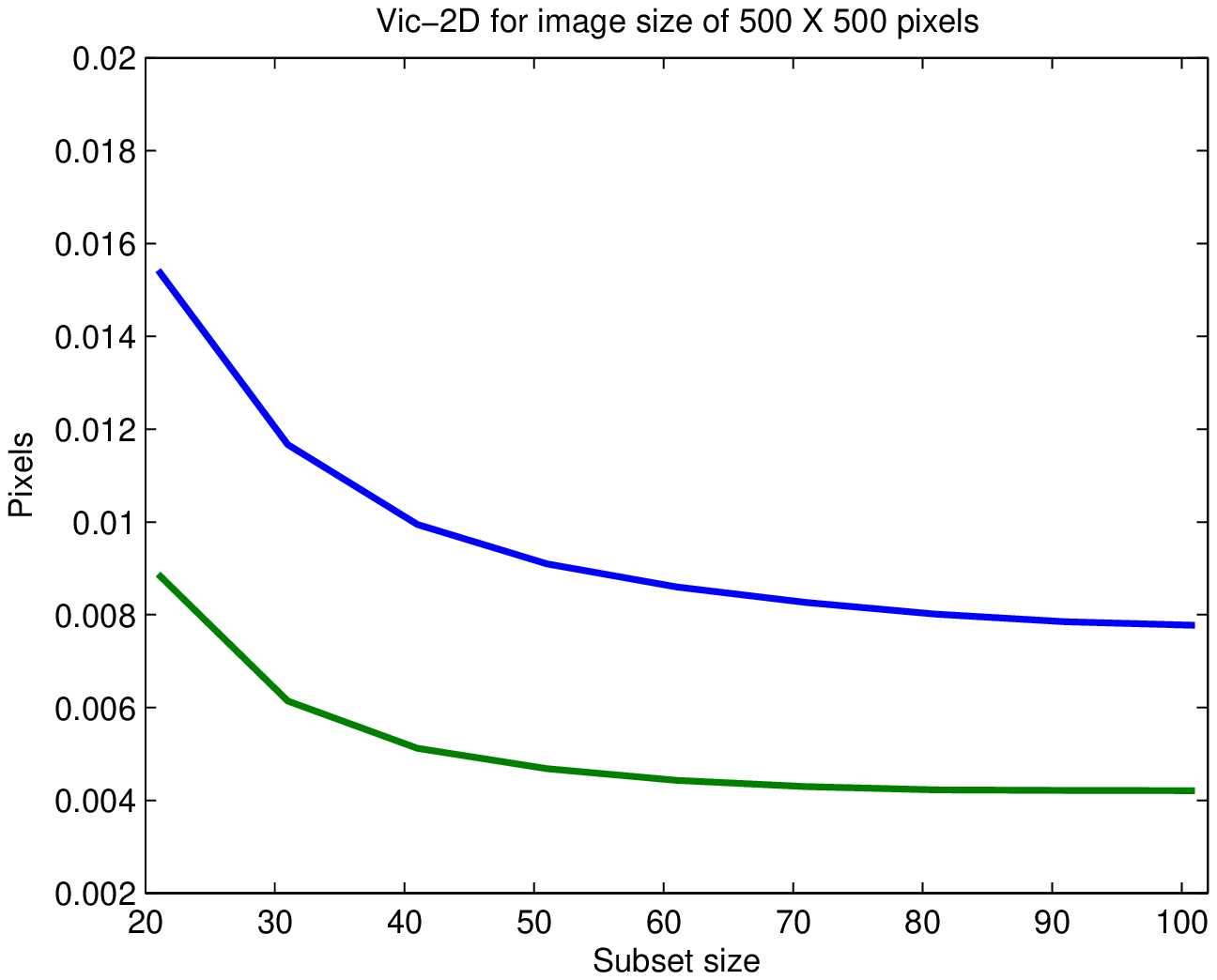}
}\\

\subfigure{
\label{fig:VIC_E2E1000}
\includegraphics[bb=100 270 470 570, clip, width=3in, angle=0]{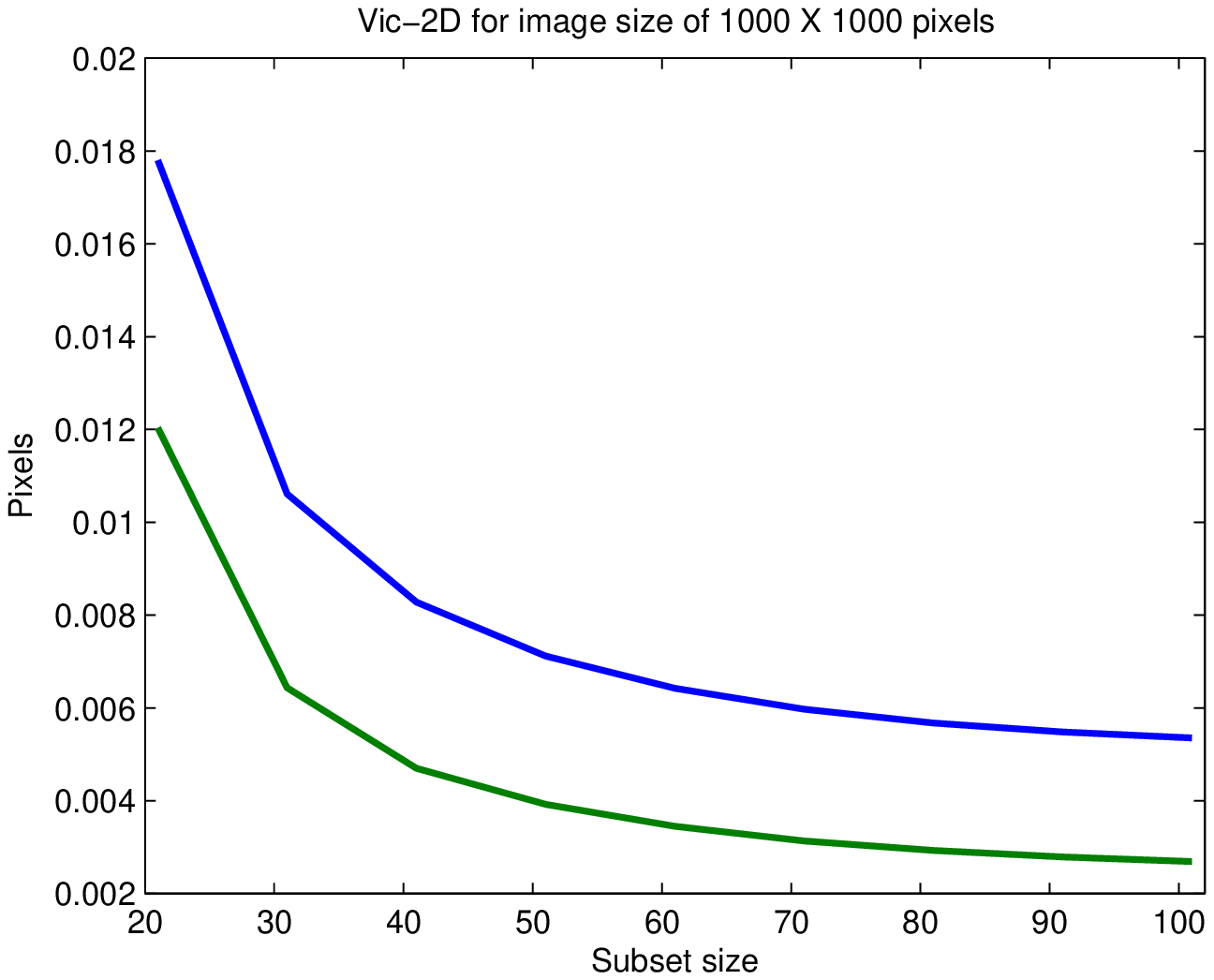}
}\\

\subfigure{
\label{fig:VIC_E2E2000}
\includegraphics[bb=100 260 470 595, clip, width=3in, angle=0]{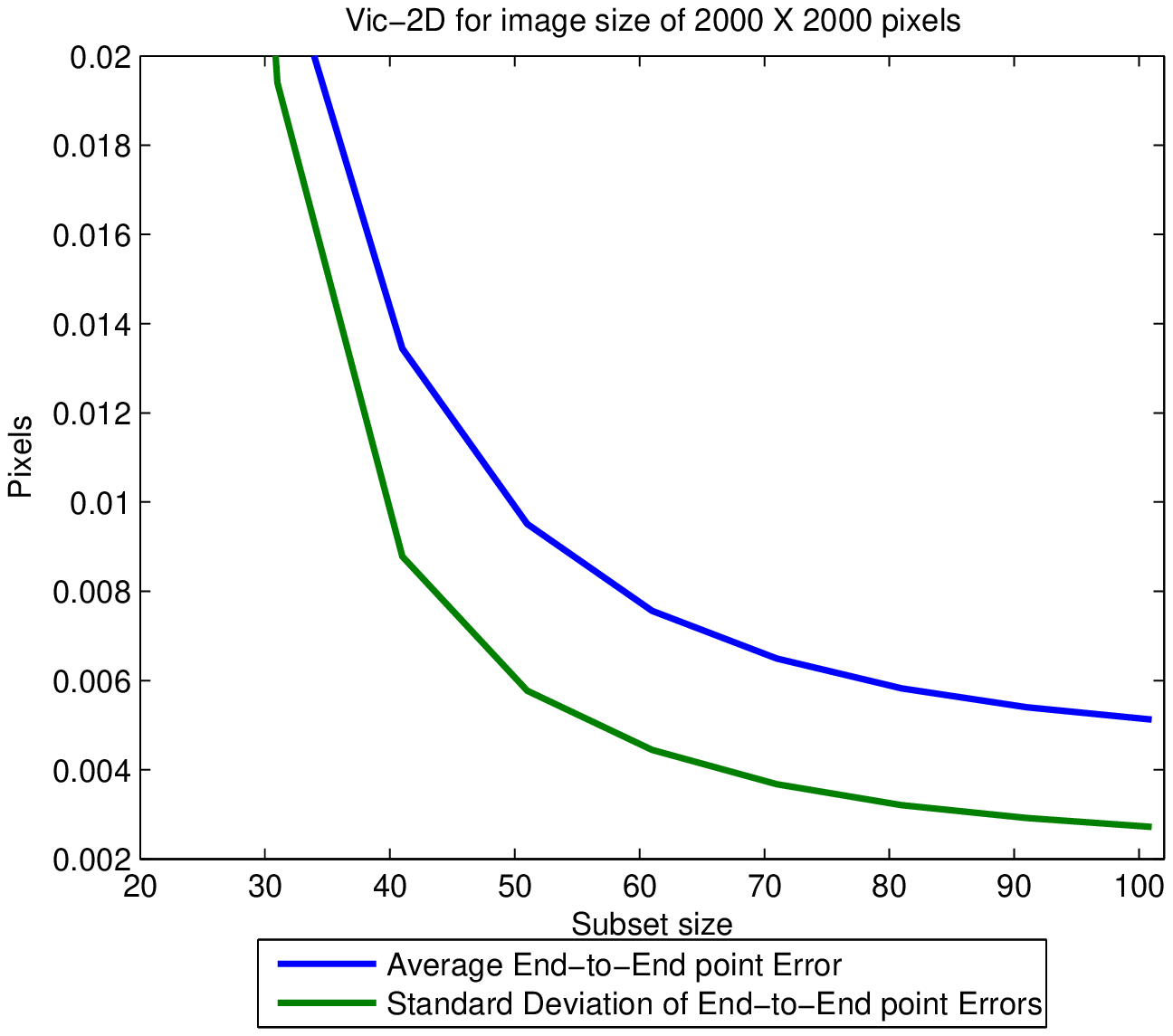}
}
\end{center}
\caption{Average End-to-End point errors and Standard Deviations for the Vic-2D software}
\label{fig:VIC_E2E}
\end{figure}

The averages of end-to-end point errors for the three DIC variants are presented in Figure \ref{fig:PF_E2E}, \ref{fig:NR_E2E} and \ref{fig:VIC_E2E}. The end-to-end error is the mean square of errors
of both horizontal $u$ and vertical $v$ directions:
\begin{eqnarray}
E_{e} = \sqrt{{E_{u}}^2 + {E_{v}}^2}
\label{equ:e2e}
\end{eqnarray}
where $E_{e}$, $E_{u}$ and $E_{v}$ are the errors in end-to-end points, horizontal direction and vertical direction, respectively.

It can be observed from Figures \ref{fig:PF}, \ref{fig:NR} and \ref{fig:VIC} that the performance of the analysed variants of DIC improve with the increase in the size of the subset, and, average errors and standard deviations of errors decrease with the increase in the size of the subset. It is also observed that increase of the subset size initially improves the performance but then the improvement diminishes.

Figure \ref{fig:PF} shows that the Basic DIC provides the worse accuracy of one-tenth of a pixel. The accuracy of the Extended DIC at the lowest subset size of $21 \times 21$ pixels for image size of $500 \times 500$ pixels is better than for higher image sizes of $1000 \times 1000$ and $2000 \times 2000$ pixels. However, the best accuracy is found at large subset size of $101 \times 101$ pixels for the image size of $2000 \times 2000$ pixels.

The rate of improvement in accuracy of the Basic DIC increases with the increase of the image size, while for each image size this rate of improvement is higher at lower subset sizes and decreases sharply with the increase in the subset size. For each image size, the accuracy level saturates at $0.03$ pixels. For the case of image size of $500 \times 500$ pixels, the accuracy level of $0.03$ is achieved for all subset sizes and thus very small improvement is observed with the increase of subset size. However, for the cases of image sizes of $1000 \times 1000$ and $2000 \times 2000$ pixels, the saturation in the improvement in accuracy is achieved by subset sizes of $41 \times 41$ and $51 \times 51$ pixels, respectively. The low standard deviations suggest the reliability of the results.

It is found that the accuracy of the Basic DIC differs in measuring horizontal component $u$ and vertical component $v$. The difference is small but exists in all image sizes. The difference found to be decreasing for the image size of $1000 \times 1000$ pixels while it is higher for the case of the image size of $2000 \times 2000$ pixels. The simulated displacements in eq. (\ref{equ:UVdisplacement}) are also not similar in both directions, that is, deformation in vertical direction $v$ is higher than the horizontal direction $u$ which may causes difference in the performance of measuring deformation in horizontal and vertical directions.

To exclude the factor of the difference in the accuracy measuring in horizontal and vertical directions, the average end-to-end point errors are calculated according to eq. (\ref{equ:e2e}). It measures the combined effect of both errors and helps comparing the overall performance of the DIC variants easily.

Figure \ref{fig:PF_E2E} depicts the average end-to-end point errors for the Basic DIC. It is seen that for all image sizes, the average end-to-end point errors decrease with increase in the subset size. For the image size of $500 \times 500$ pixels, the average end-to-end error remains between $0.03$ and $0.04$ pixels. Similarly, for the image size of $2000 \times 2000$, the average end-to-end points error decreases sharply with the increase in the size of the subset. However, with higher sizes of subsets, it remains between values of $0.03$ and $0.04$ pixels. For the image size of $1000
\times 1000$ pixels, the average End-to-End points error decreases to the value lower than $0.03$ pixels at large subset sizes. This standard deviation for End-to-End point error is very low which
proves the consistent performance of the Basic DIC.

For the Extended DIC, Figure \ref{fig:NR} shows higher accuracy to at least one-hundredth of a pixel which is 10 times better than the Basic DIC. Similarly to the case of Basic DIC, it is
observed that the better accuracy is already achieved at low image size of $500 \times 500$ pixel for small subset sizes while the higher accuracy is achieved for the case of high image size of $2000
\times 2000$ pixels and high subset sizes. The difference in accuracy of the horizontal $u$ and vertical $v$ components of displacements is negligible at the high image sizes of $1000 \times
1000$ and $2000 \times 2000$ pixels. The average end-to-end point error presented in Figure \ref{fig:NR_E2E} shows that the accuracy improves with the increase of the subset size while the rate of
the improvement keeps decreasing with the increase of the subset size.

It can also be observed from Figure \ref{fig:NR_E2E} that the shapes of the curve of the average end-to-end point errors for all image sizes are similar. The lowest average end-to-end point error achieved for the image size of $500 \times 500$ pixels is approximately $0.006$ pixels, while the same error decreases to below $0.006$ pixels for the image sizes of $1000 \times 1000$ and $2000
\times 2000$ pixels. The standard deviation for all errors remain very low which indicates the consistency in performance of the Extended DIC.

For the commercial software of Vic-2d, the documented accuracy is expected to be one-hundredth of a pixel \cite{Sutton08review}. It is observed from Figure \ref{fig:VIC} that the accuracy level does achieve one-hundredth of a pixel. The performance of Vic-2d improves with the increase in subset size for all image sizes. The rate of improvement is higher for low image size of $500 \times 500$ pixels as compared to the high image sizes of $1000 \times 1000$ and $2000 \times 2000$ pixels. The difference between the performance in measuring the horizontal $u$ and the vertical $v$ displacements is the lowest in the image size of $1000 \times 1000$ pixels while for the lower image size of $500 \times 500$ pixels, the difference is found to be the highest.

The average end-to-end point errors for Vic-2d are plotted in Figure \ref{fig:VIC_E2E}. It shows the same pattern as observed for the case of the Extended DIC but with comparatively higher errors. For the image size of $500 \times 500$ pixels, the average end-to-end point error decreases to approximately $0.008$ pixels while for the image sizes of $1000 \times 1000$ and $2000 \times 2000$ pixels, the average end-to-end error decreases to slightly below $0.006$ pixels. The low standard deviation indicates reliability of the results.

\begin{figure}[!tb]
\begin{center}

\subfigure{
\label{fig:PF_time}
\includegraphics[bb=100 270 470 570, clip, width=3in, angle=0]{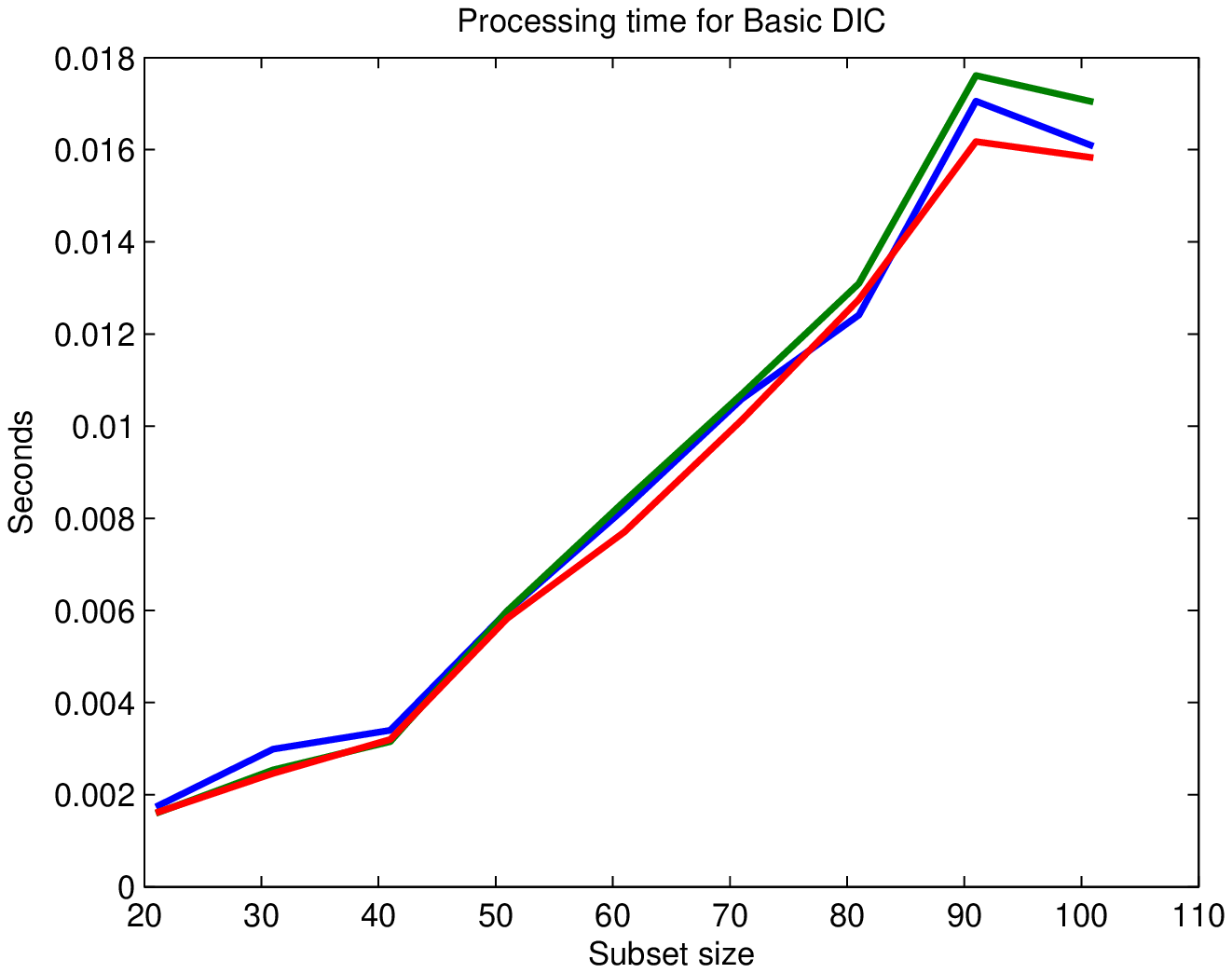}
}\\

\subfigure{
\label{fig:NR_time}
\includegraphics[bb=100 270 470 570, clip, width=3in, angle=0]{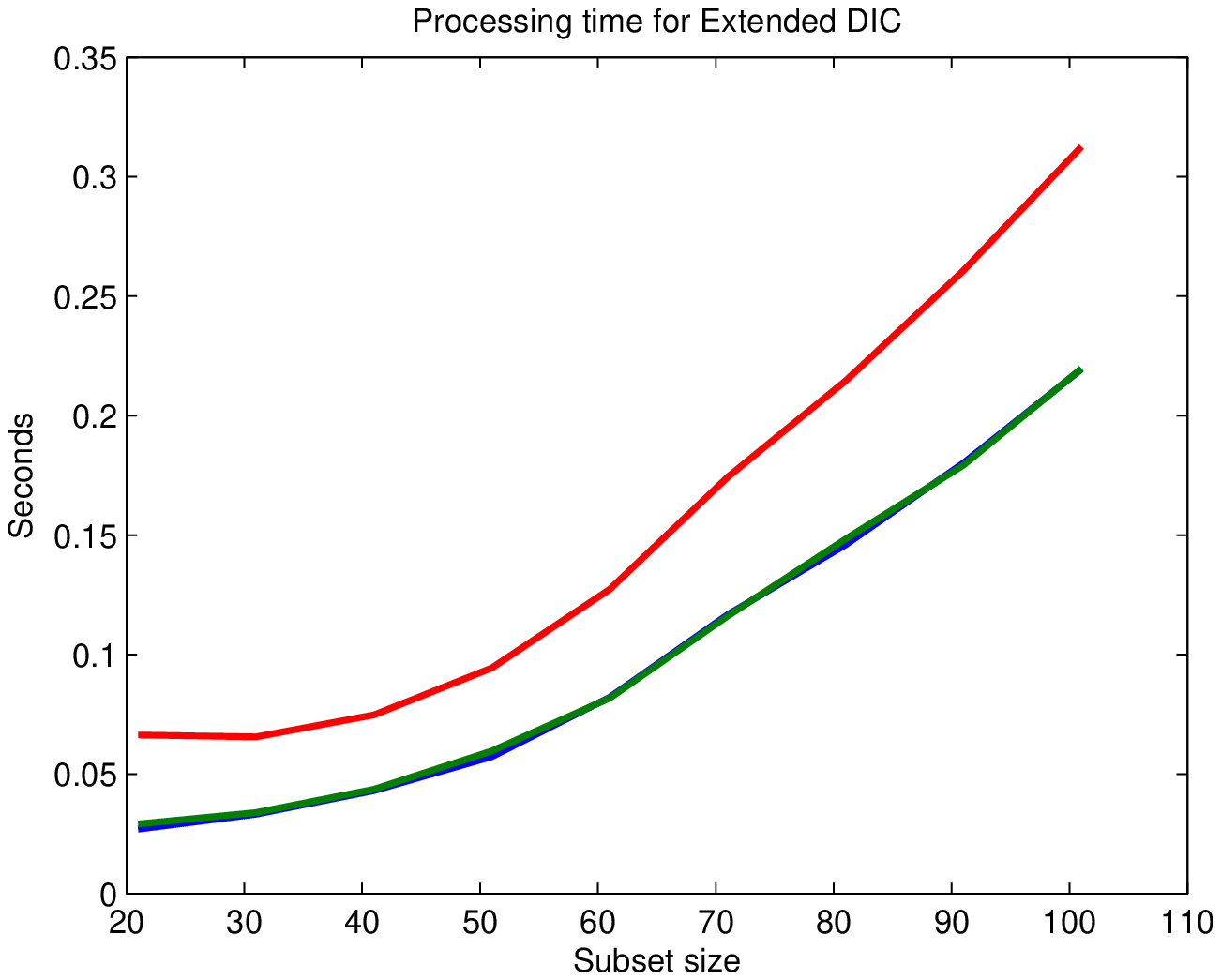}
}\\

\subfigure{
\label{fig:VIC_time}
\includegraphics[bb=100 250 470 600, clip, width=3in, angle=0]{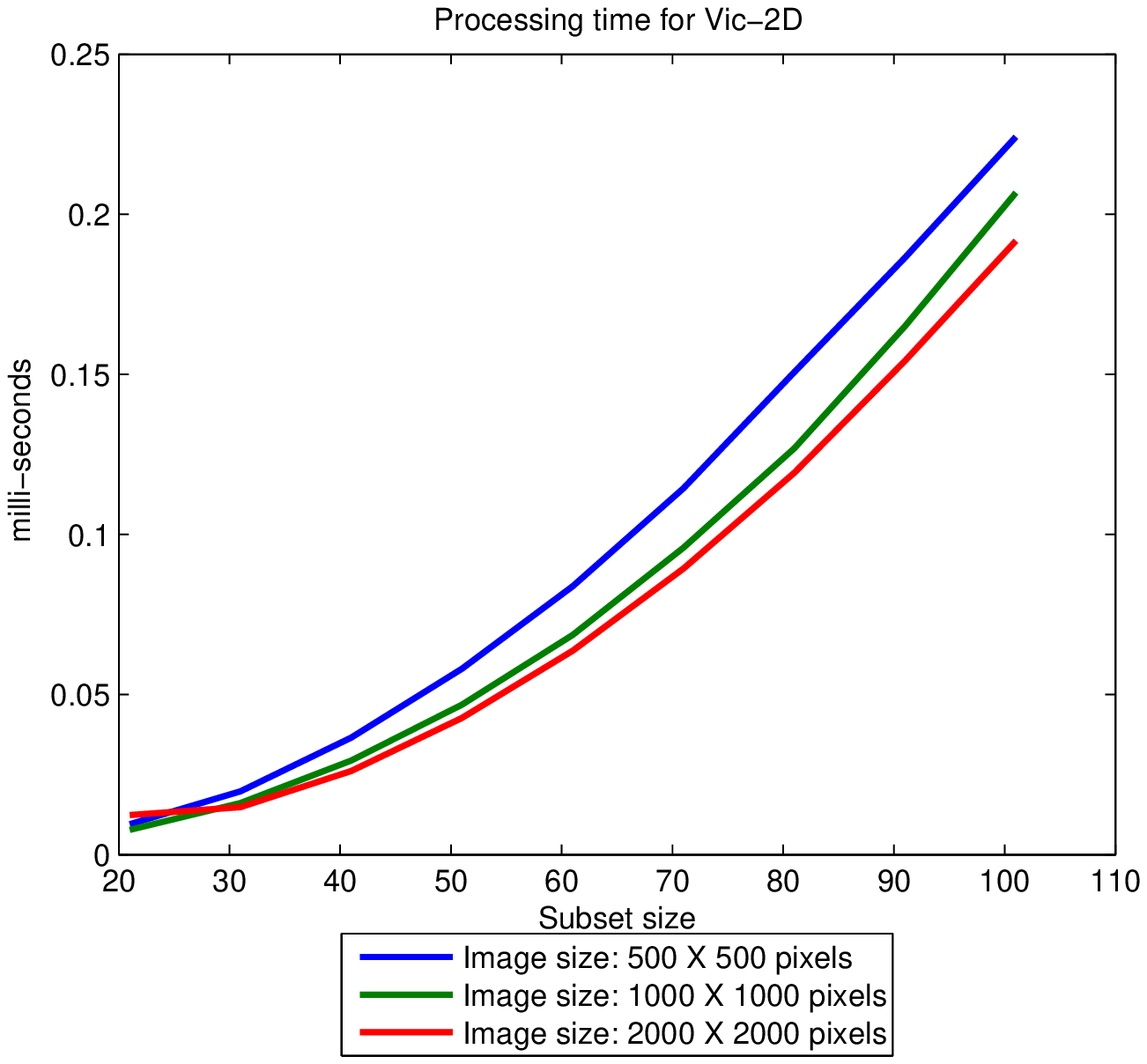}
}
\end{center}
\caption{Average computational time of processing one subset}
\label{fig:time}
\end{figure}

The computation times for the DIC techniques are presented in Figure \ref{fig:time}. The 16-bit Windows 7 Professional operating system running on  Intel$\textregistered$ Core-i7-2600 CPU @ 3.40GHz
with 16 GB RAM is used for processing data of all experiments. As mentioned earlier, the Basic DIC and the Extended DIC are implemented in Matlab and the time consumed is measured using Matlab tools. However, Vic-2d is a commercial software and the time presented is the time reported by the software itself.

\begin{figure}[!tb]
\begin{center}
\includegraphics[bb=110 270 500 570, clip, width=3in, angle=0]{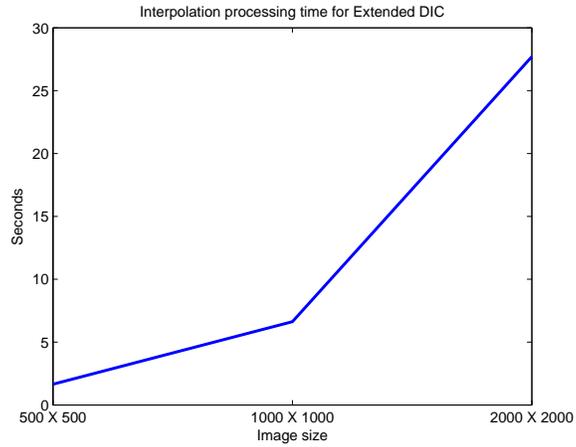}
\end{center}
\caption{Computational time interpolation of an image}
\label{fig:NR_interp_time}
\end{figure}

\begin{figure}[tb]
\begin{center}
\includegraphics[bb=110 255 470 595, clip, width=3in, angle=0]{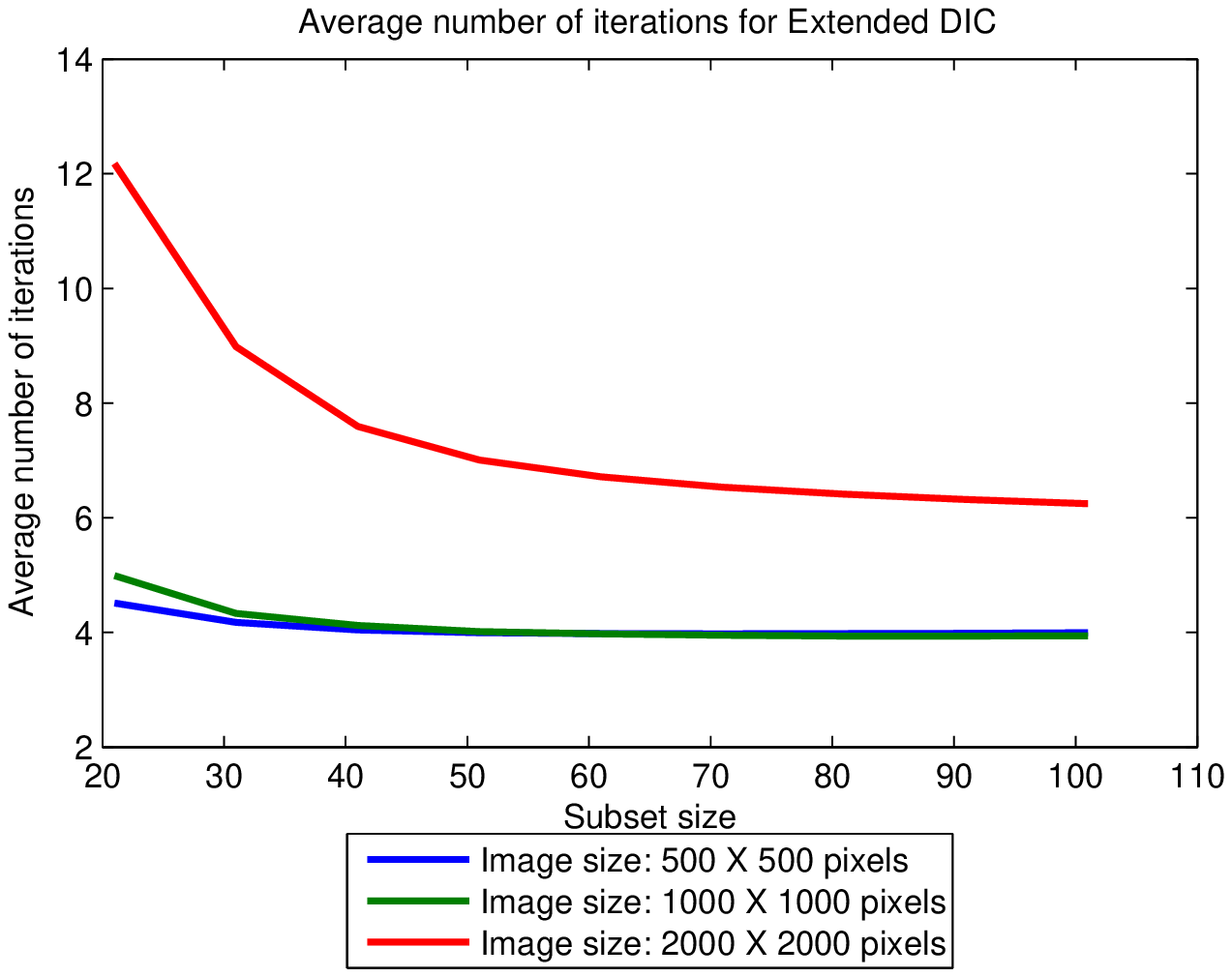}
\end{center}
\caption{Average number of iterations by Newtion-Raphson method of maximizing the cross correlation point}
\label{fig:NR_iteration}
\end{figure}

The computation time taken by the Extended DIC can be divided into two parts. The first part is related to the bicubic spline interpolation of the deformed image, which is executed once for each set of images and is presented in Figure \ref{fig:NR_interp_time}. The second part of computation time is presented in Figure \ref{fig:NR_time} which describes the average time taken by the optimization technique (the Newton-Raphson method) to find the highest cross-correlation coefficient point for the center point of the subset. The average numbers of iterations taken by the optimization are presented in Figure \ref{fig:NR_iteration}.

The average computation time required to find the highest cross-correlation of the center point of the subset by the Basic DIC is presented in Figure \ref{fig:PF_time}.

The Vic-2d software is not an open source software and details of its algorithm are not known. To calculate the computation time consumed by Vic-2d, the entire possible area of reference image is selected for the cross correlation. The time consumed by cross correlating entire images is divided by the number of the possible cross-correlation points. The time consumed and the number of possible correlation points are obtained from Vic-2d. The average computation time calculated for the different subset sizes are presented in Figure \ref{fig:VIC_time}.

From the analysis of Figure \ref{fig:NR_interp_time}, it can be observed that the time consumed by the bicubic spline interpolation in the Extended DIC increases with the increase in the image size and is independent of the subset size. As presented in Figure \ref{fig:NR_time}, the average time for the optimization increases with the increase of the subset size while it also increases partially with the increase in the image size. However, after analyzing Figure \ref{fig:NR_iteration} and \ref{fig:NR_time}, it is found that the time consumed by the optimization mainly depends on the number of iterations taken by the Newton-Raphson method and is independent of the image size. It is observed in Figure \ref{fig:NR_iteration} that the average number of iterations for the image sizes of $500 \times 500$ and $1000 \times 1000$ pixel is approximately $4$ while for image size of $2000 \times 2000$ pixels, the number of iterations increases to the range of $6$ to $12$.

The average time consumed by the Basic DIC, Figure \ref{fig:PF_time}, is increasing with the increase in the subset size but does not significantly increases with the increase of the image
size. The computation time remain the same for the subset sizes of $91 \times 91$ and $101 \times 101$ pixels.

The time consumed by Vic-2d, Figure \ref{fig:VIC_time}, is increasing with the increase of the subset size. It is also seen that the image size does not affect the computational performance. Strangely, the increase in size of the image reduces the time taken by the Vic-2d to process a subset. The DIC process is conducted on entire image and it is not possible to find the exact reason for this strange result. Also, the computation time taken by Vic-2d cannot be compared to the computation time of other DIC variants because the implementation platform, algorithm and software engineering technique of Vic-2d are unknown. According to the comparative analysis presented by Andrews in \cite{Andrews2012}, it can be safely stated that the computation time for the Basic DIC and the Extended DIC implemented in Matlab will be improved significantly if a low level development language and good parallel processing algorithms are used.

The analysis of Figure \ref{fig:time} shows that the computation time increases parabolically with the increase in the subset size for the Extended DIC and Vic-2d, while, the computation time increases linearly with the increase in the subset size for the Basic DIC.

\subsection*{Strain Reconstruction}

The performance of DIC is commonly measured by the accuracy of the displacement reconstruction and, subsequently measured in pixels. However, the analysis of the material behavior requires reconstruction of strain fields. Strain fields can be obtained by numerically differentiating displacement fields. Furthermore, strain is also automatically obtained in the Extended DIC, where six unknown parameters are estimated which include strain parameters. Only few studies mention the evaluation of DIC in terms of strain but are focused on application of DIC to study the properties of the material. These studies do not analyze the accuracy and precision of DIC \cite{Hoult2013,Amy2011,Sutton2008,Barranger12,Barone2001}. Pan \cite{Pan09review} recommended that reliable strain fields can be obtained if noise in the reconstructed displacement is smoothed by a filter and then strain is obtained by numerically differentiating displacement.

In this study, the strain field in a bend cantilever reads \cite{BookTimoshenko}
\begin{eqnarray}
\epsilon_{x} &=& \frac{\partial u}{\partial x} = -\frac{3 P x y}{2 c^{3} E} \nonumber \\
\epsilon_{y} &=& \frac{\partial v}{\partial y} = \frac{3 \nu P x y}{2 c^{3} E} \nonumber \\
\gamma_{xy}  &=& \frac{\partial u}{\partial y} + \frac{\partial v}{\partial x} = -\frac{3 P}{4 c G} (1-\frac{y^{2}}{c^{2}})
\label{equ:strain}
\end{eqnarray}

Two sets of images with sizes of $500 \times 500$ and $1000 \times 1000$ pixels are selected for the analysis. Three different methods are used to calculate the strain and results are matched with results of eq. (\ref{equ:strain}). The moving average smoothing filter is used which replaces each value $g(x,y)$ by the average of the values in a square of size $ (2n+1) \times (2n+1)$ centered around $g(x,y)$. It is given by
\begin{eqnarray}
g(x,y) = \sum\limits_{j=-n}^{n} \sum\limits_{k=-n}^{n} g(x+j,y+k)
\label{equ:filter}
\end{eqnarray}
where $(2n+1)$ is the size of the filter.

\begin{figure}[!tb]
\begin{center}

\subfigure{
\label{fig:NR_disp_strain500_1}
\includegraphics[bb=110 270 475 570, clip, width=3in, angle=0]{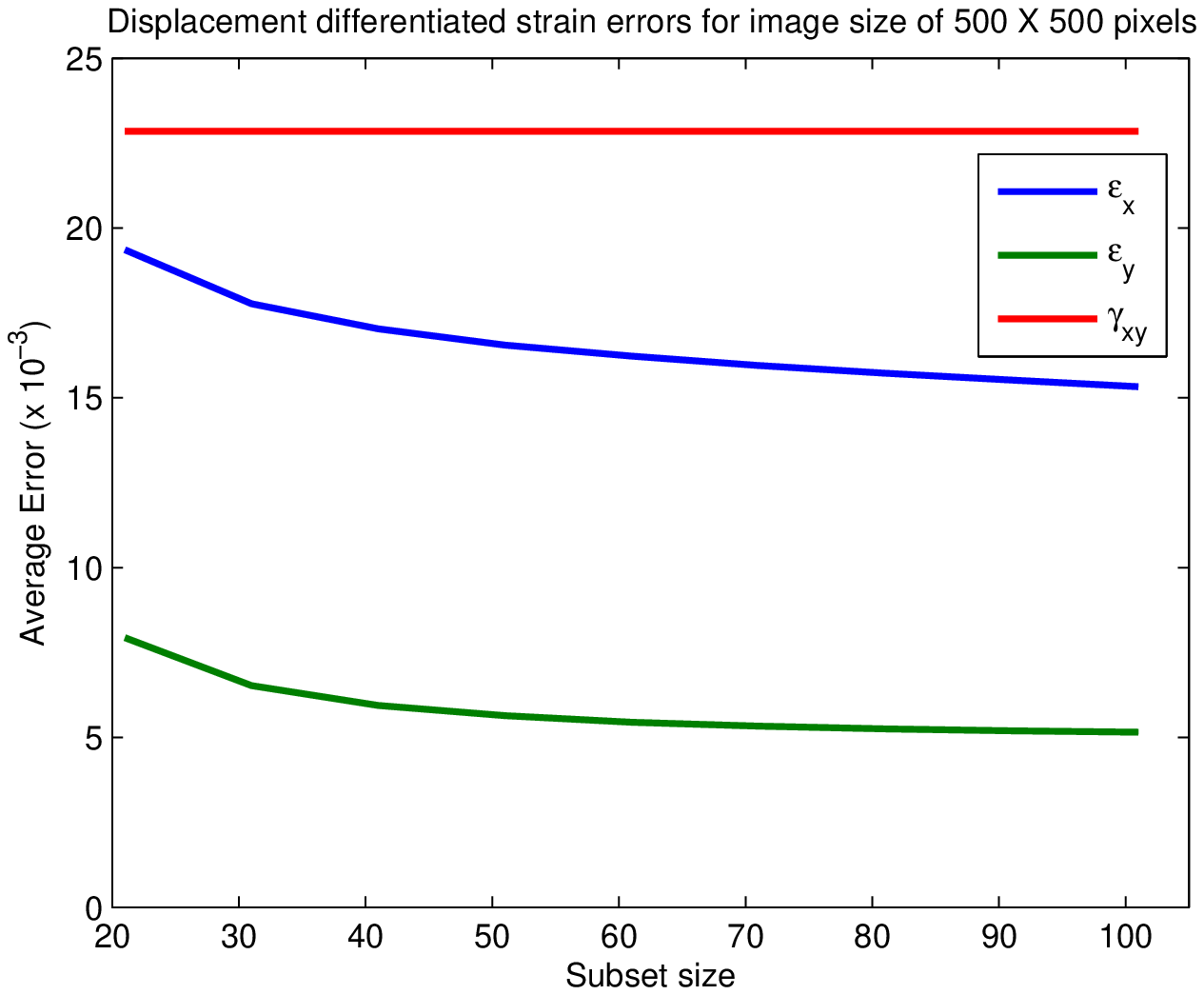}
}\\

\subfigure{
\label{fig:NR_disp_strain1000_1}
\includegraphics[bb=110 270 476 570, clip, width=3in, angle=0]{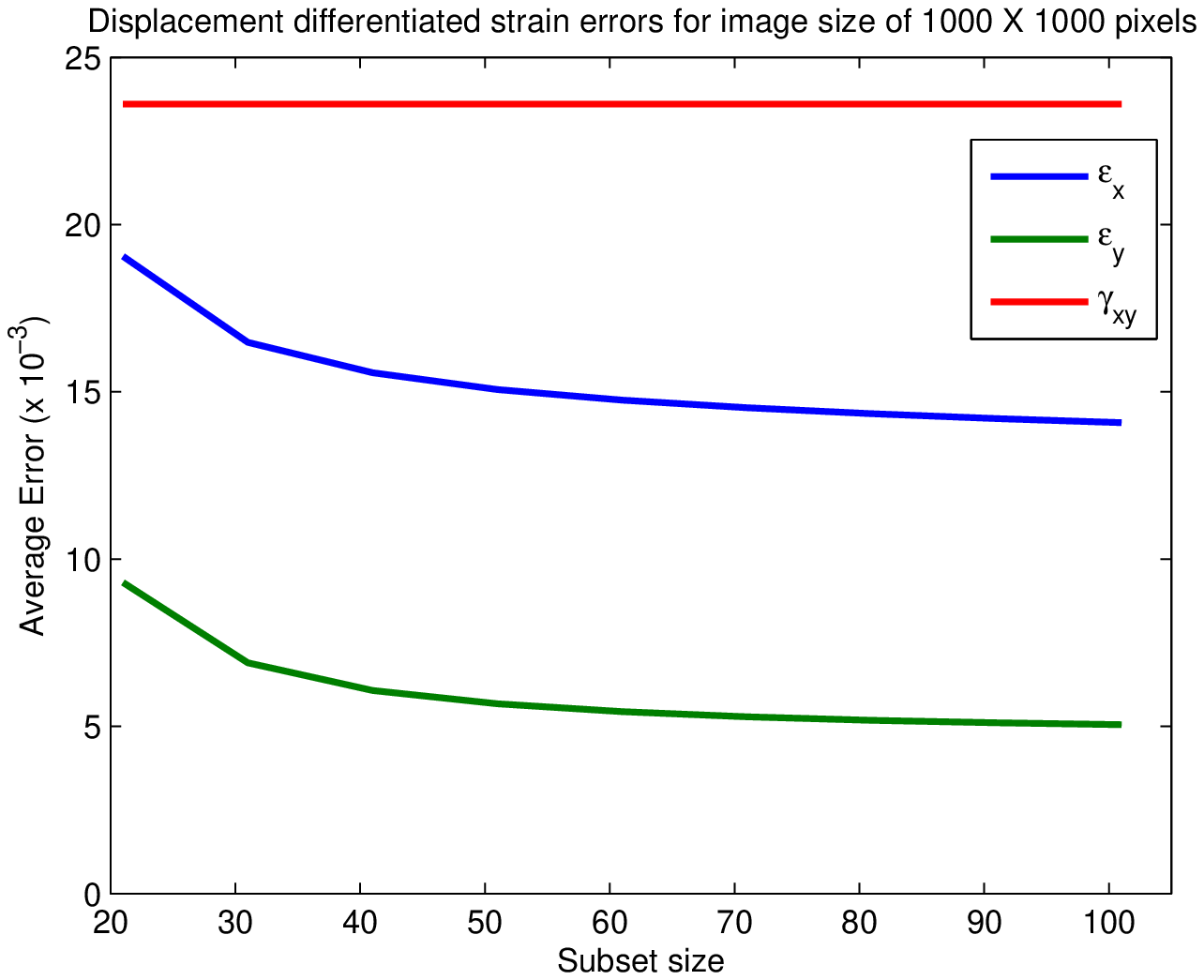}
}\\

\end{center}
\caption{Strains obtained by differentiating the displacement fields obtained by the Extended DIC for image sizes of (a) $500 \times 500$, and (b) $1000 \times 1000$}
\label{fig:NR_disp_strain_1}
\end{figure}

\begin{figure}[!tb]
\begin{center}

\subfigure{
\label{fig:NR_disp_strain500_2}
\includegraphics[bb=110 270 485 570, clip, width=3in, angle=0]{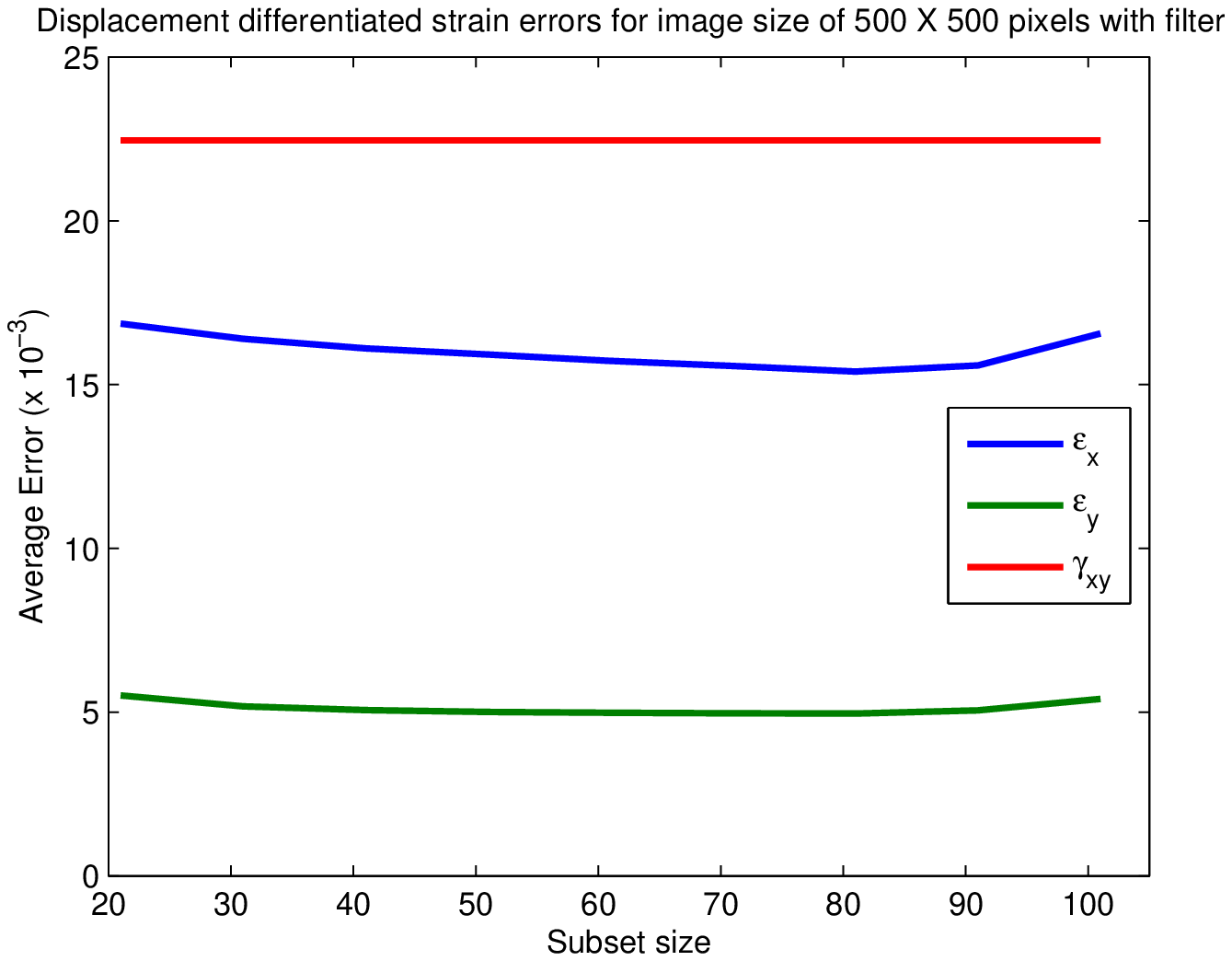}
}\\

\subfigure{
\label{fig:NR_disp_strain1000_2}
\includegraphics[bb=110 270 490 570, clip, width=3in, angle=0]{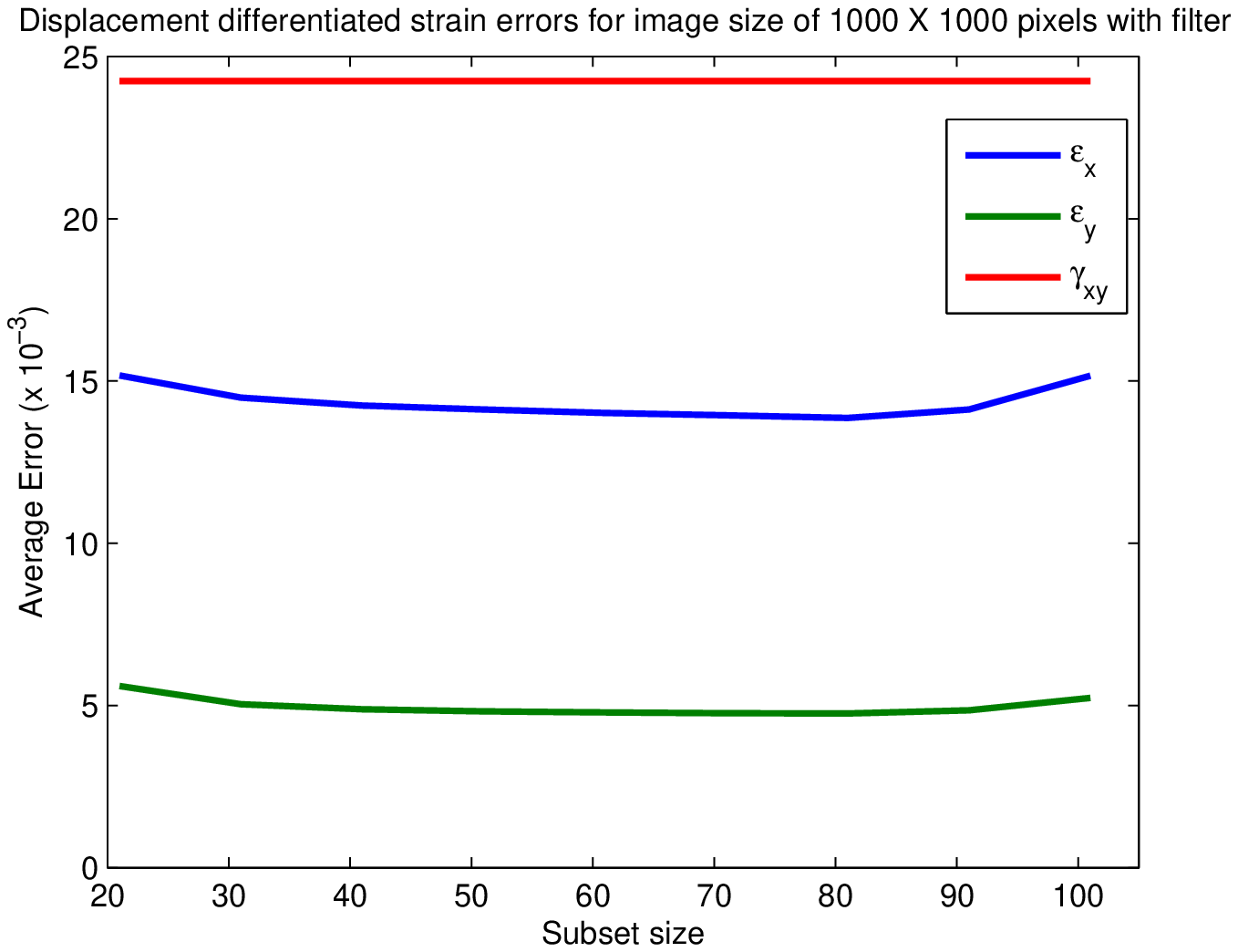}
}\\

\end{center}
\caption{Strains obtained by smoothing the displacement fields obtained by the Extended DIC and then differentiating them for image sizes of (a) $500 \times 500$, and (b) $1000 \times 1000$}
\label{fig:NR_disp_strain_2}
\end{figure}

In the first method, the measured displacements obtained by the Extended DIC are numerically differentiated to calculate strain. The Extended DIC is selected due to its high accuracy of displacement
results. The results are presented in Figure \ref{fig:NR_disp_strain_1}. Then according to Pan's \cite{Pan09review} recommendation, the displacement results obtained by the Extended DIC are smoothed by a filter of size equal to the half of the size of the subset. Then, the smoothed displacements are numerically differentiated to obtain strain fields which are presented in Figure \ref{fig:NR_disp_strain_2}.

\begin{figure}[!tb]
\begin{center}

\subfigure{
\label{fig:NR_strain500_1}
\includegraphics[bb=110 270 470 570, clip, width=3in, angle=0]{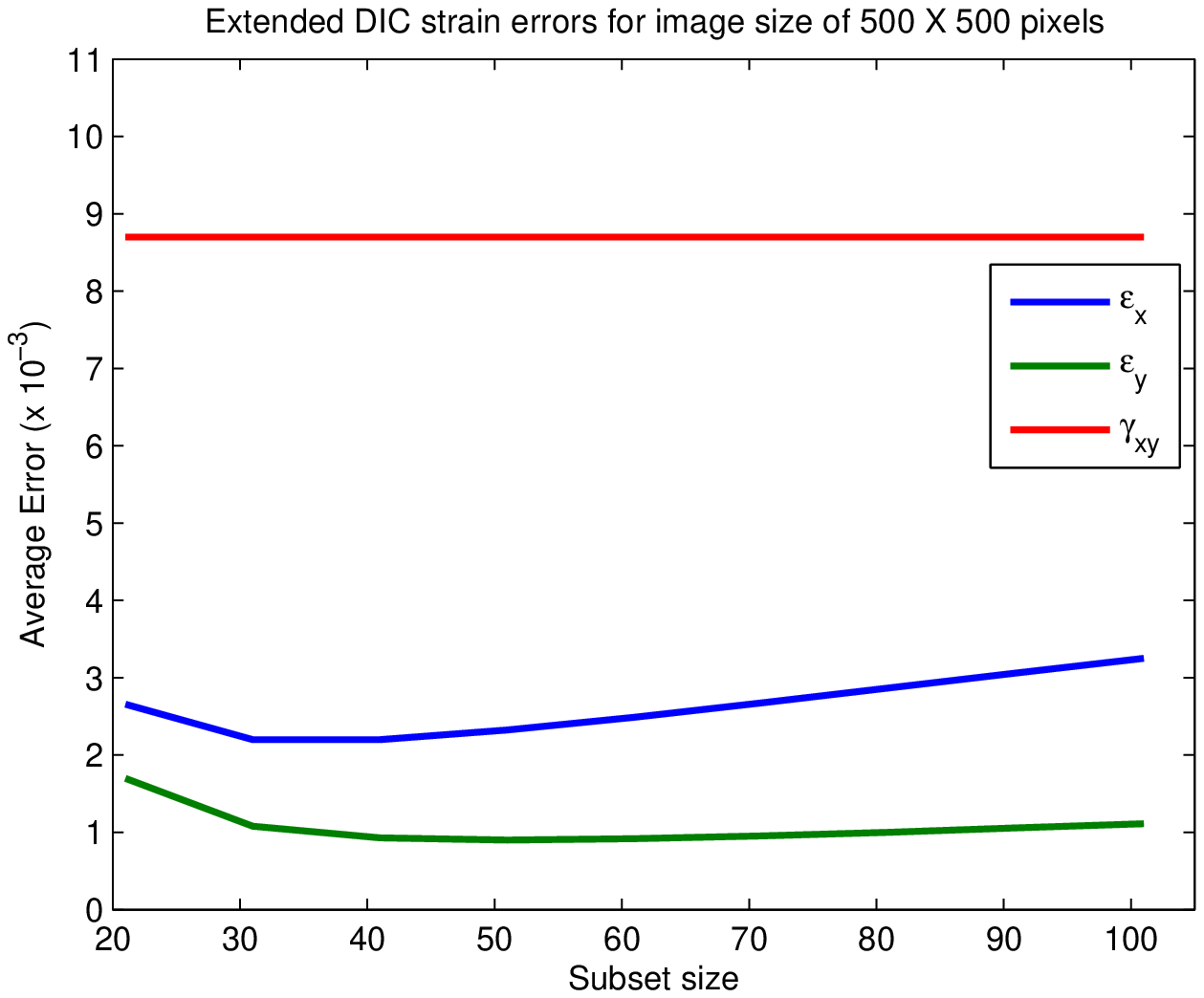}
}\\

\subfigure{
\label{fig:NR_strain1000_1}
\includegraphics[bb=110 270 470 570, clip, width=3in, angle=0]{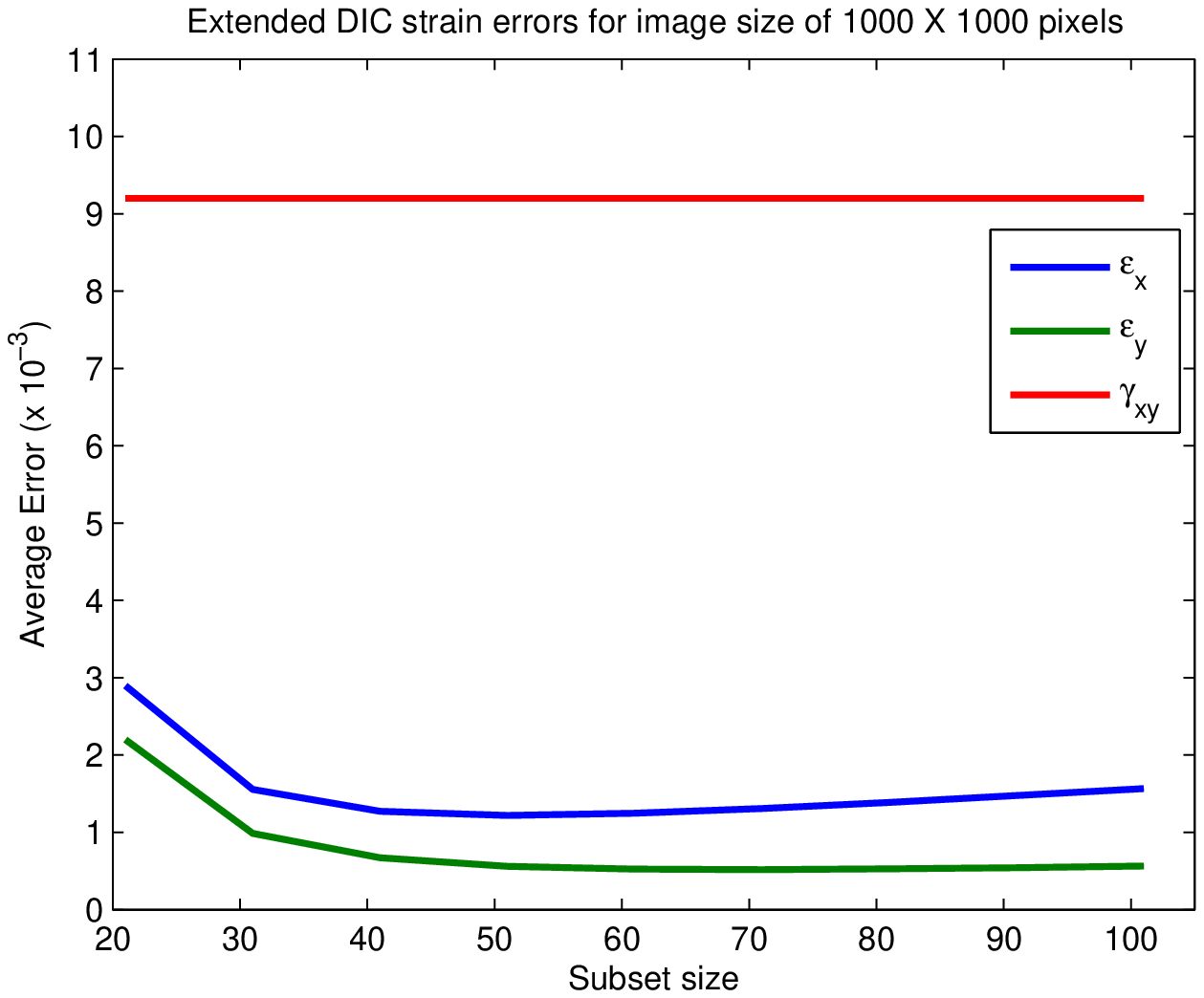}
}\\

\end{center}
\caption{Strains calculated from the displacement gradients obtained by the Extended DIC for image sizes of (a) $500 \times 500$, and (b) $1000 \times 1000$}
\label{fig:NR_strain_1}
\end{figure}

\begin{figure}[!tb]
\begin{center}

\subfigure{
\label{fig:NR_strain500_2}
\includegraphics[bb=110 270 470 570, clip, width=3in, angle=0]{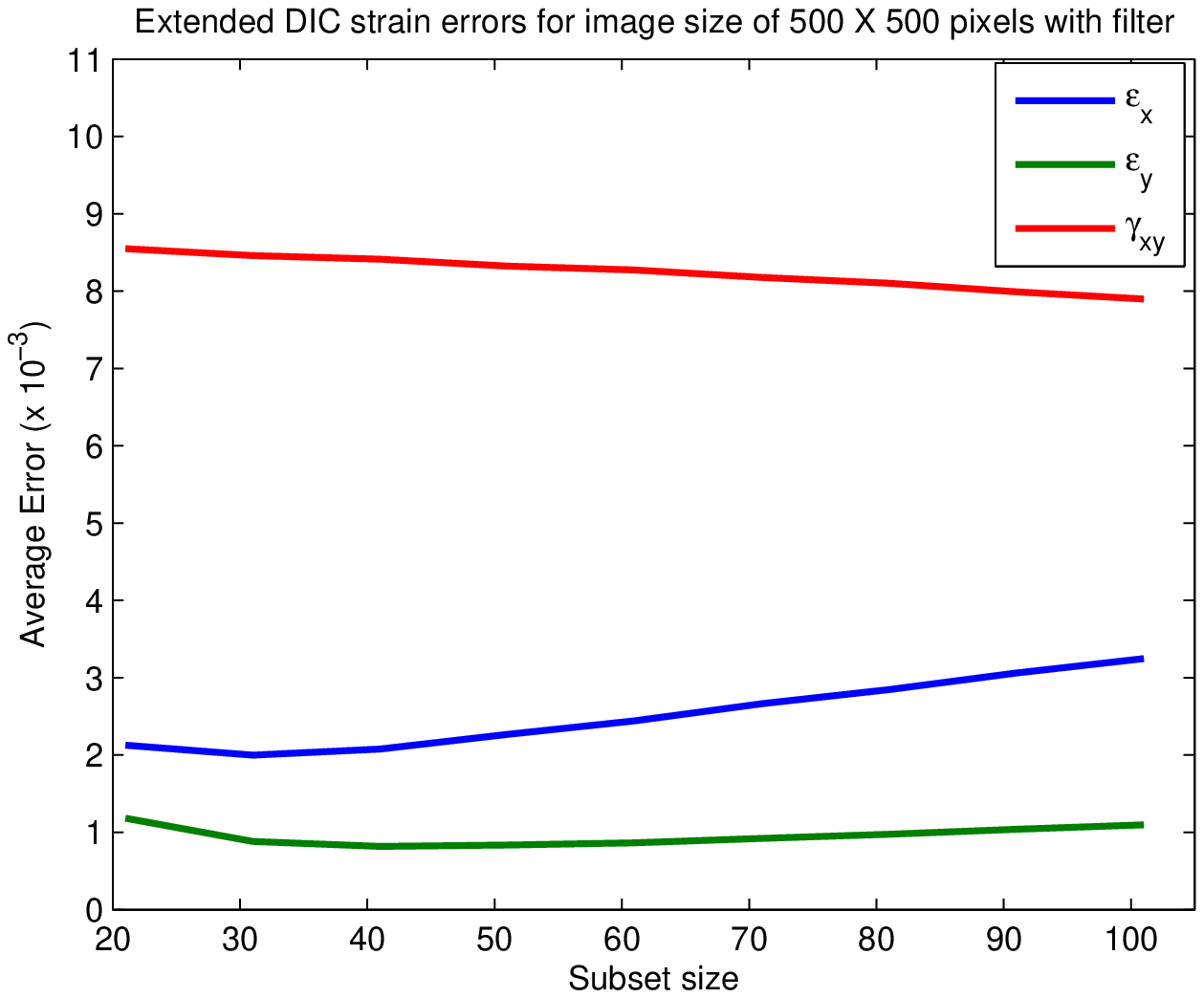}
}\\

\subfigure{
\label{fig:NR_strain1000_2}
\includegraphics[bb=110 270 475 570, clip, width=3in, angle=0]{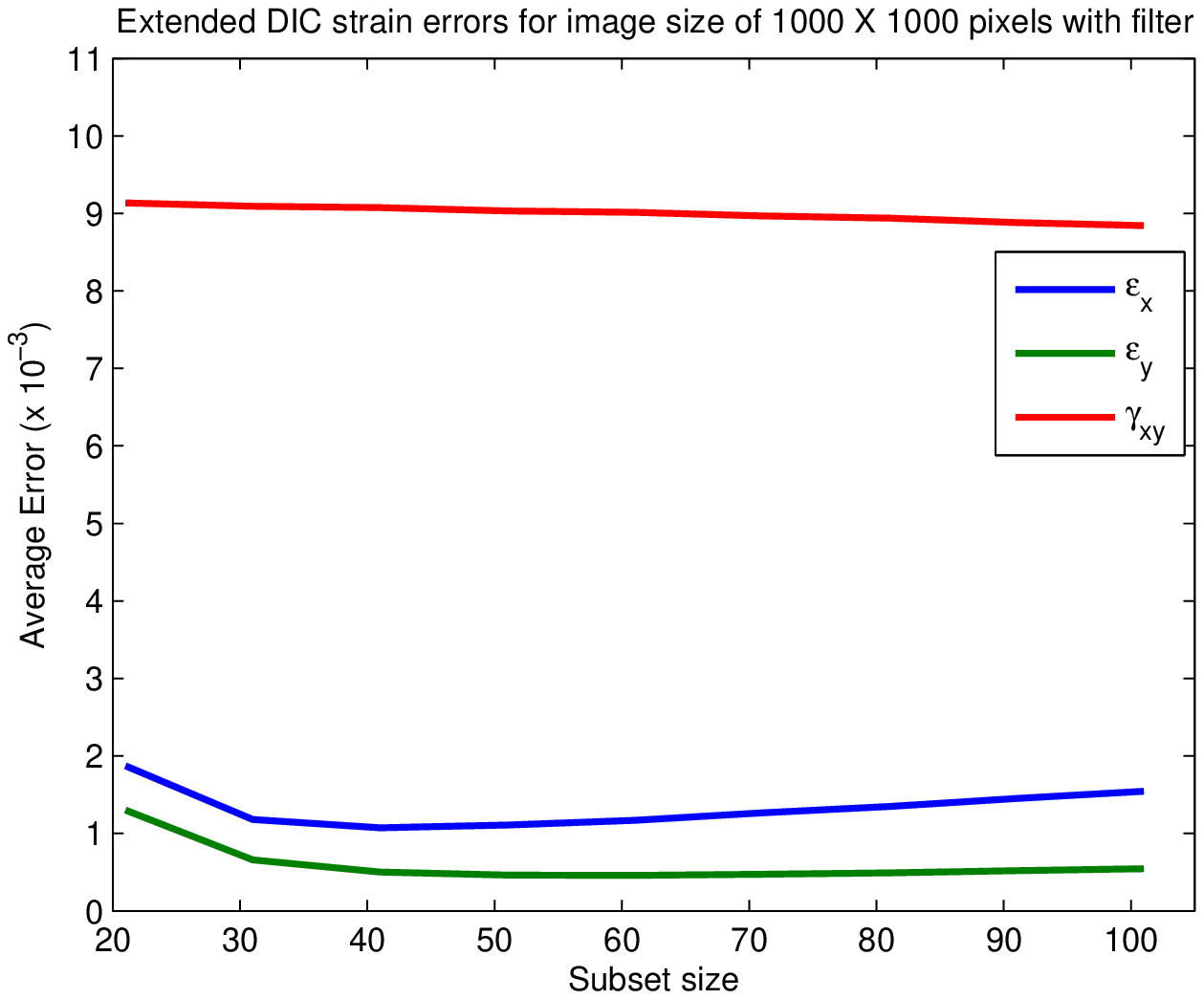}
}\\

\end{center}
\caption{Strains calculated from the displacement gradients obtained by the Extended DIC and smoothed by filter size of half of the subset size for image sizes of (a) $500 \times 500$, and (b) $1000 \times 1000$}
\label{fig:NR_strain_2}
\end{figure}

In the second method, the strain components are obtained by extracting them from the displacement gradients estimated by the Extended DIC as per in eq. (\ref{equ:var6}). The raw results for image sizes of $500 \times 500$ and $1000 \times 1000$ are presented in Figure \ref{fig:NR_strain_1}. The same strain field smoothed to reduce the noise by a filter of size equal to the half of the size of the subset is presented in Figure \ref{fig:NR_strain_2}.

\begin{figure}[!tb]
\begin{center}

\subfigure{
\label{fig:VIC_strain500_1}
\includegraphics[bb=110 270 470 570, clip, width=3in, angle=0]{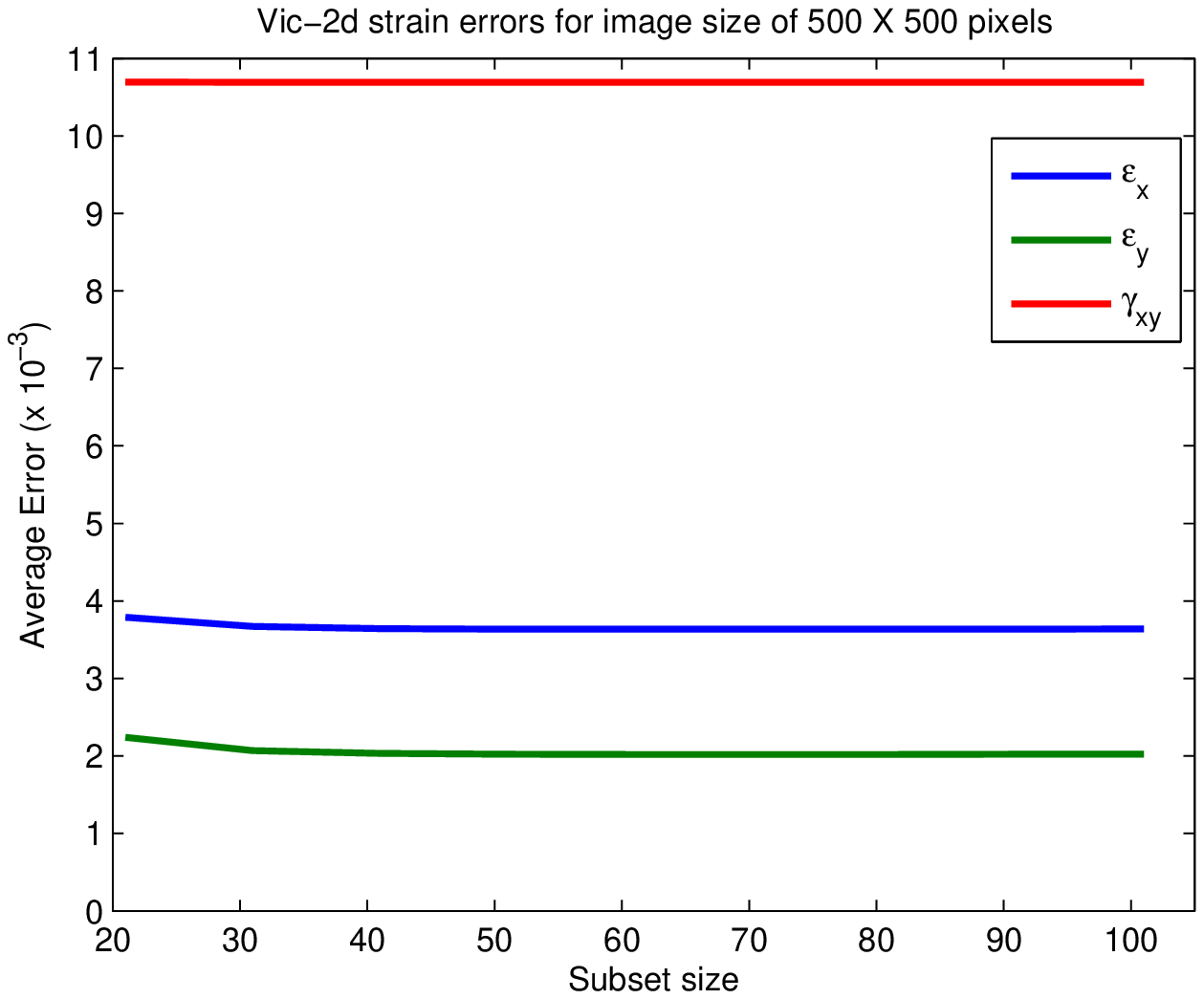}
}\\

\subfigure{
\label{fig:VIC_strain1000_1}
\includegraphics[bb=110 270 470 570, clip, width=3in, angle=0]{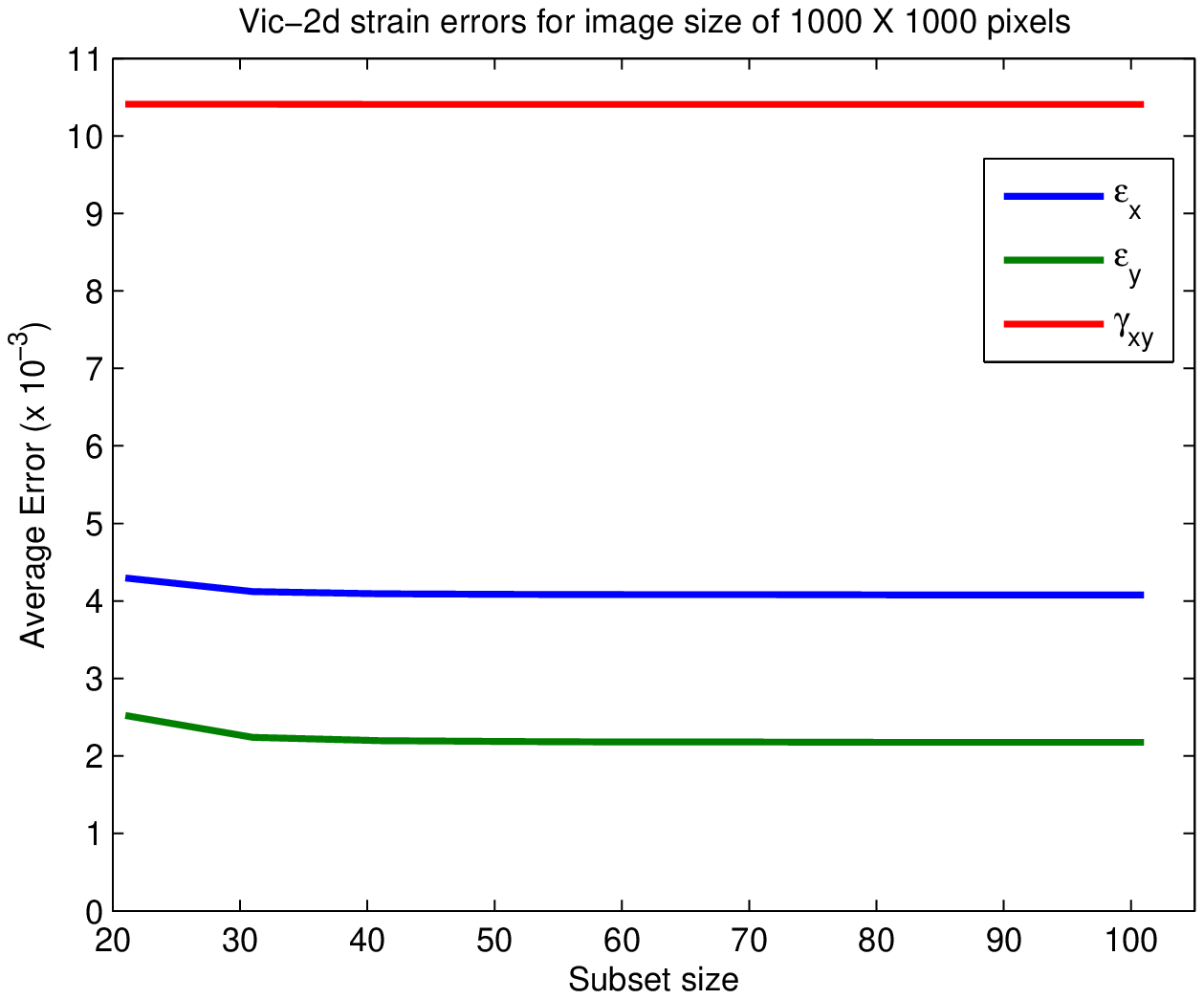}
}\\

\end{center}
\caption{ Strain results obtained by Vic-2d and smoothed by smallest filter size of $5$ pixels for image sizes of (a) $500 \times 500$, and (b) $1000 \times 1000$ }
\label{fig:VIC_strain_1}
\end{figure}

\begin{figure}[!tb]
\begin{center}

\subfigure{
\label{fig:VIC_strain500_2}
\includegraphics[bb=110 270 470 570, clip, width=3in, angle=0]{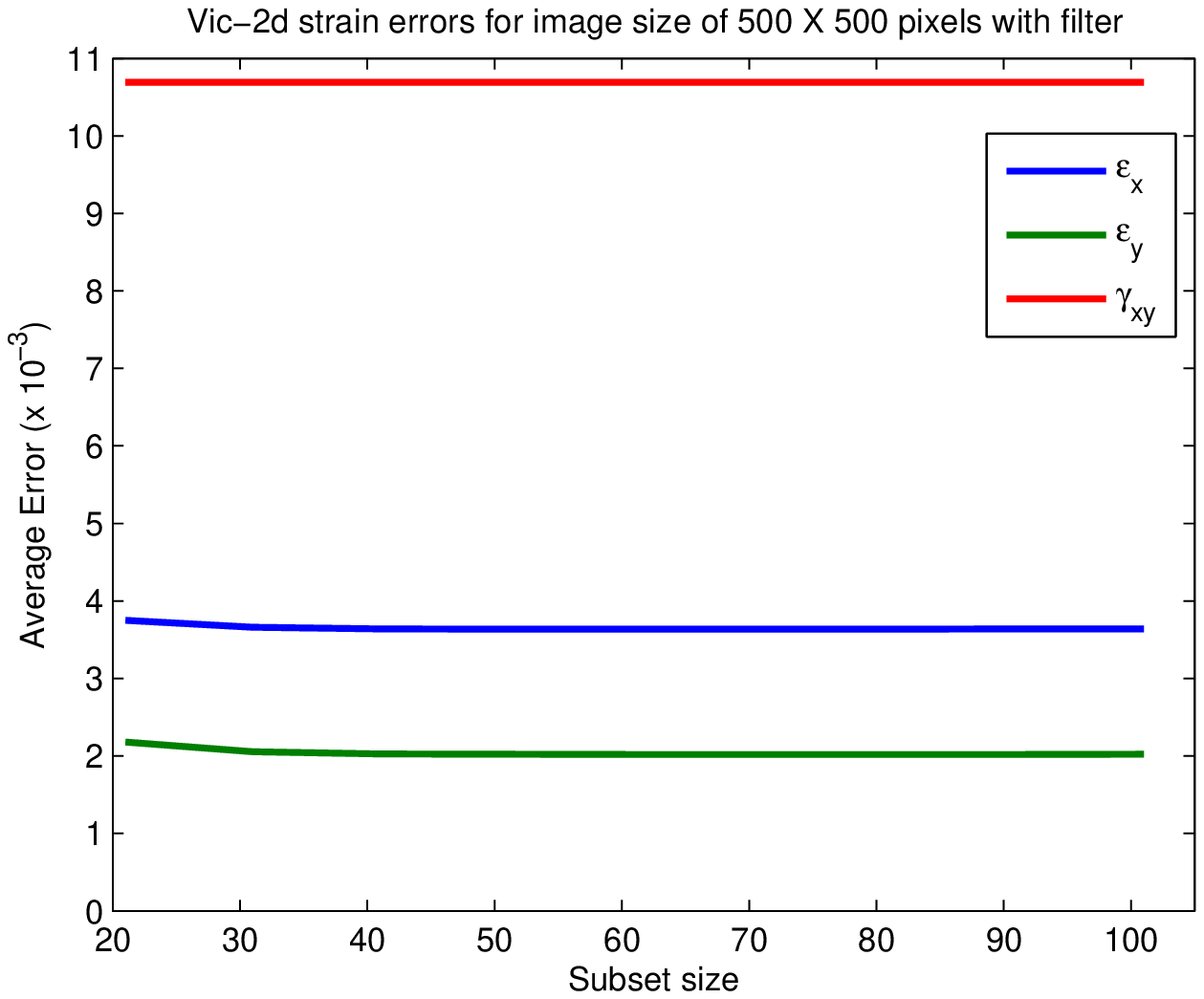}
}\\

\subfigure{
\label{fig:VIC_strain1000_2}
\includegraphics[bb=110 270 470 570, clip, width=3in, angle=0]{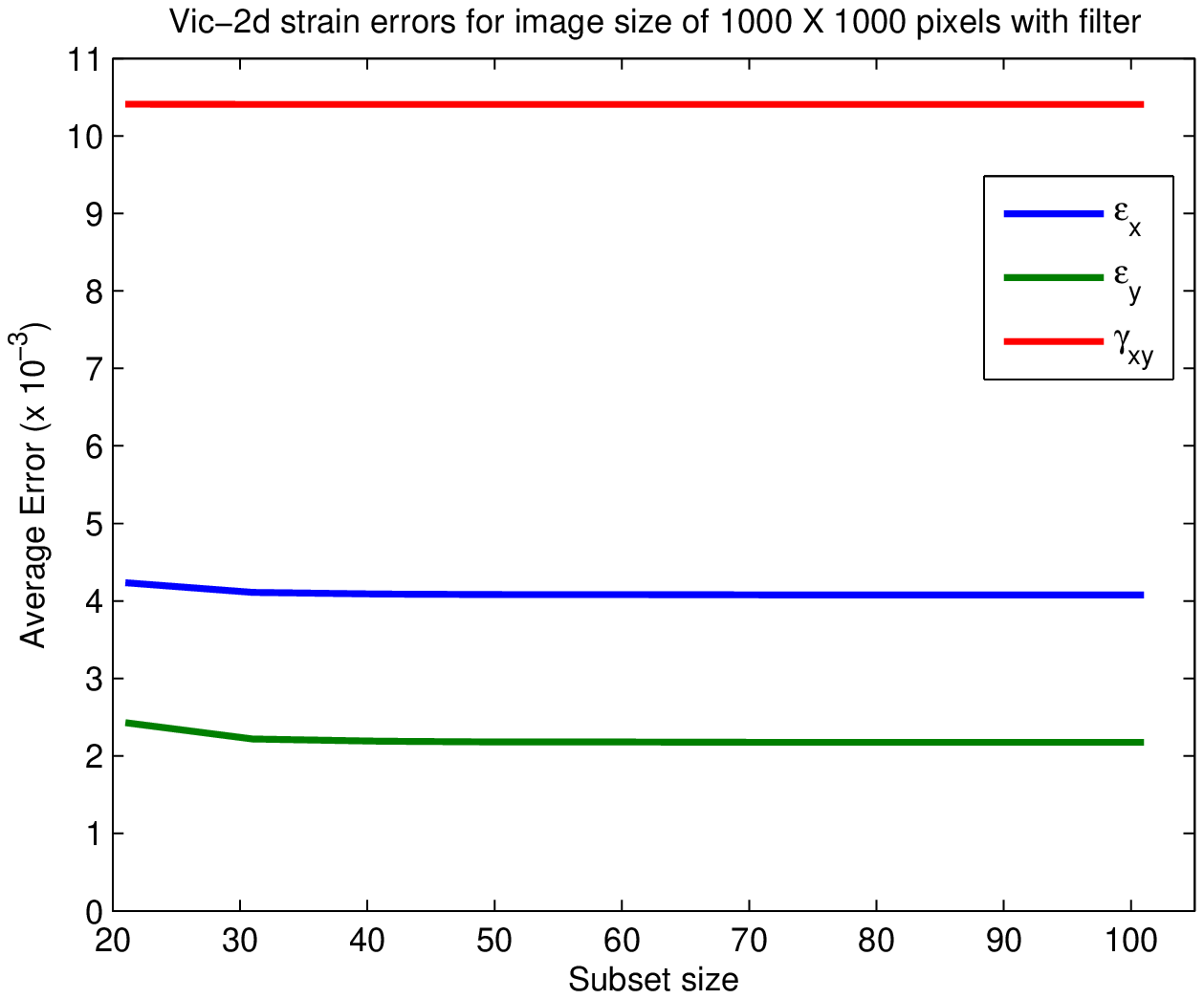}
}\\

\end{center}
\caption{ Strain results obtained by Vic-2d and smoothed by filter size of half of the subset size for image sizes of (a) $500 \times 500$, and (b) $1000 \times 1000$ }
\label{fig:VIC_strain_2}
\end{figure}

In the third method, Vic-2d is used to extract strain at each point of the specimen. It is assumed that Vic-2d uses numerical differentiation of the obtained displacement field to reconstruct the strain field. Whenever the strain is extracted, it is mandatory in Vic-2d to smoothen the reconstructed strain field by a filter which range from $5$ pixels to size of the image. For the comparative study the minimum size of filter is used and results are depicted in Figure \ref{fig:VIC_strain_1}. Later, the size of the filter is increased to half of the size of the subset and the results obtained are presented in Figure \ref{fig:VIC_strain_2}. It is seen that the strain field reconstruction is provided in Vic-2d as a the post-processing option consumes extra time after the displacement reconstruction.

From the analysis of the strain reconstruction, it is observed that the optimal results are obtained by the second method, where the strain is extracted from the displacement gradients estimated by the Extended DIC. The use of the filter to smoothen the strain field improves the results slightly but not significantly. Thus, the differentiation of displacements even after smoothing cannot match the accuracy of direct computation of the strains from the displacement gradients determined by the Extended DIC in the process of maximization of the cross correlations.

It is further seen from Figures \ref{fig:NR_strain_1} and \ref{fig:NR_strain_2} that the results are not significantly changed by the change of image size or subset size. The subset size of $40 \times 40$ provides slightly better results than other subset sizes.

\subsection*{Recommendations}

\begin{table}[!tb]
  \renewcommand{\arraystretch}{1.5}
  \caption{Summary of results of DIC variants}
    \begin{tabular*}{\textwidth}{p{0.3\textwidth} p{0.2\textwidth} p{0.2\textwidth} p{0.2\textwidth}}
    \toprule
    \textbf{Description} & \textbf{Basic DIC} & \textbf{Vic-2d} & \textbf{Extended DIC} \\
    \midrule
    Minimum accuracy in measuring displacement & 0.1 pixel & 0.01 pixel & 0.01 pixel \\
    Accuracy in displacement field reconstruction & Good  & Very Good & Excellent \\
    Accuracy in strain field reconstruction & ---   & Good  & Very good \\
    computation time & Excellent & Fair* & Fair \\
    Effect of increase in the subset size on accuracy & Improves slightly till it reaches best & Improves significantly till it reaches best & Improves significantly till it reaches best \\
    Effect of increase in the subset size on computation time & Increases linearly & increases parabolically & increases parabolically \\
    Effect of increase in the image size on accuracy & Slight improvement & Slight improvement on large subset sizes & Slight improvement on large subset sizes \\
    Effect of increase in the image size on computation time & remains same & slightly reduces* & Initially remans same but increase at higher image sizes \\
    \bottomrule
    \multicolumn{4}{r}{*Computation time cannot be compared properly as Vic-2d is not open source} \\
    \end{tabular*}%
  \label{tab:comparison}%
\end{table}%

Summary of the analysis of the DIC variants are presented in Table \ref{tab:comparison}. From this comparison the following recommendations can be drawn.
\begin{itemize}
  \item For high accuracy of displacement (one-hundredth of a pixel) and strain reconstruction, the Extended DIC is preferred. However the high accuracy is achieved at the expense of the computation time.
  \item For fast computation, the Basic DIC is preferred but it delivers comparatively low accuracy of displacement reconstruction (one-tenth of a pixel). Strain reconstructed by differentiating displacements obtained by the Basic DIC is of low accuracy.
  \item The accuracy of displacement and strain reconstruction primarily depends upon the intensity variation in the subset and its size, which play a critical role in successful cross-correlation of subsets. It is recommended to have on average of $4$ or more speckles in each subset.
  \item The accuracy in displacement reconstruction improves marginally with increase of the subset size. However, increase in the size of subset increases the time consumed for processing and after reaching the subset of $4$ speckles in a subset, the improvement in accuracy is not significant. The same conclusion can be made about the strain reconstruction.
  \item The increase of the image size (the pixel count) allows capturing more details and would be expected to provide better accuracy. However, our analysis shows that the accuracy of the displacement and strain reconstruction improves insignificantly. Furthermore, the large image size means increase of speckle size in pixels and requires large subset size for optimal performance, which significantly increases the computational time.
  \item The end-to-end point errors are a better indicator of the performance of a DIC as compared to the separate horizontal $u$ and vertical $v$ displacement errors.
  \item The open source DISTRESS Simulator developed in this study provides a reliable tool to measure performance of the DIC before using it for real world measurements.
\end{itemize}

\section{Conclusion}

In this paper, we analyzed the photogrammetric solutions for reconstruction and monitoring of displacement and strain fields in natural and engineering materials and structures. The most commonly used photogrammetric technique is the Digital Image Correlation (DIC) which is based on comparing images before and after deformation. The photographed surface needs to have unique speckle pattern.

We analyzed the accuracy of different variants of DIC, which are, the Basic DIC, Extended DIC and the commercial software VIC 2D. The Basic DIC and VIC 2D reconstruct the displacement field and strain field is then inferred in post processing using numerical differentiation. The Extended DIC reconstructs both displacement and strain fields simultaneously. For the analysis we developed the DISTRESS Simulator which is used to generate synthetic images of displacement and strain fields. We found that the Basic DIC and VIC 2D are faster in processing at the expense of accuracy. The Extended DIC provides the highest accuracy of displacement and strain reconstruction and is preferred to be used for the case of rotation in deformation. It is also observed that the speckle pattern plays a critical role in achieving high accuracy of the reconstruction. The increase in the subset size and pixel count of the image do not significantly improve the accuracy of DIC. In order to achieve high accuracy using DIC one needs to keep the subset size large enough to have average of four or more speckles. The end-to-end point errors provide a better indicator of performance measurement.

\section*{Acknowledgement}
We are thankful to the Australian Postgraduate Award, the Bruce \& Betty Scholarship and the University of Western Australia merit top-up Scholarship for their funding.

{\small
\bibliographystyle{ieeetr}
\bibliography{refrences}

\begin{thebibliography}{10}

\bibitem{Tyson11}
J.~Tyson, ``Optical 3{D} deformation and strain measurement,'' in {\em Pumps
  and Pipes} (M.~G. Davies, A.~B. Lumsden, W.~E. Kline, and I.~Kakadiaris,
  eds.), pp.~147--164, Springer US, 2011.
\newblock 10.1007/978-1-4419-6012-2-13.

\bibitem{Gruen1985}
A.~Gruen, ``Adaptive least squares correlation: a powerful image matching
  technique,'' {\em South African Journal of Photogrammetry, Remote Sensing and
  Cartography}, vol.~14, no.~3, pp.~175--187, 1985.

\bibitem{Barazzetti2011iWitness}
L.~Barazzetti and M.~Scaioni, ``Photogrammetric tools for deformation
  measurements,'' in {\em XX Congresso Associazione Italiana di Meccanica
  Teorica e Applicata (AIMETA)}, p.~10, 2011.

\bibitem{Werner2002}
T.~Werner and A.~Zisserman, ``New techniques for automated architectural
  reconstruction from photographs,'' in {\em Computer Vision—ECCV 2002},
  pp.~541--555, Springer, 2002.

\bibitem{Pan09review}
B.~Pan, K.~Qian, H.~Xie, and A.~Asundi, ``Two-dimensional digital image
  correlation for in-plane displacement and strain measurement: a review,''
  {\em Measurement Science and Technology}, vol.~20, no.~6, p.~062001, 2009.

\bibitem{Sutton08review}
M.~A. Sutton, ``{Digital Image Correlation for Shape and Deformation
  Measurements},'' in {\em Springer Handbook of Experimental Solid Mechanics}
  (W.~N. Sharpe, ed.), ch.~20, pp.~565--600, Boston, MA: Springer US, 2008.

\bibitem{Rodriquez2011}
A.~Rodriguez, J.~Rabunal, J.~Perez, and F.~Martinez-Abella, ``Study of strength
  tests with computer vision techniques,'' in {\em New Challenges on
  Bioinspired Applications} (J.~Ferrandez, J.~Alvarez~Sanchez, F.~Paz, and
  F.~Toledo, eds.), vol.~6687 of {\em Lecture Notes in Computer Science},
  pp.~257--266, Springer Berlin Heidelberg, 2011.

\bibitem{Rodriguez2012}
A.~Rodriguez, J.~R. Rabuñal, J.~L. Pérez, and F.~Martínez-Abella, ``Optical
  analysis of strength tests based on block-matching techniques,'' {\em
  Computer-Aided Civil and Infrastructure Engineering}, vol.~27, no.~8,
  pp.~573--593, 2012.

\bibitem{Roux2009}
S.~Roux, J.~Rethore, and F.~Hild, ``Digital image correlation and fracture: an
  advanced technique for estimating stress intensity factors of 2{D} and 3{D}
  cracks,'' {\em Journal of Physics D: Applied Physics}, vol.~42, no.~21,
  p.~214004, 2009.

\bibitem{Brown1992survey}
L.~G. Brown, ``A survey of image registration techniques,'' {\em ACM computing
  surveys (CSUR)}, vol.~24, no.~4, pp.~325--376, 1992.

\bibitem{Shah2010}
S.~A.~A. Shah, M.~A. Manzoor, M.~Farooq, A.~A. Shah, A.~Bais, K.~Yahya, and
  G.~Hassan, ``Efficient implementation of image registration based on feature
  tracking,'' in {\em Wireless Communications Networking and Mobile Computing
  (WiCOM), 2010 6th International Conference on}, pp.~1--3, IEEE, 2010.

\bibitem{TectonicPlatesRef2012}
B.~Azua, I.~Tello, J.~Sanchez, and M.~Escamilla, ``Three-dimensional
  visualisation of the tectonic movement over mexico by means of the global
  positioning system (gps) measurements,'' in {\em True-3D in Cartography}
  (M.~Buchroithner, ed.), Lecture Notes in Geoinformation and Cartography,
  pp.~239--255, Springer Berlin Heidelberg, 2012.

\bibitem{Tong05}
W.~Tong, ``An evaluation of digital image correlation criteria for strain
  mapping applications,'' {\em Strain}, vol.~41, no.~4, pp.~167--175, 2005.

\bibitem{Chu85}
T.~Chu, W.~Ranson, and M.~Sutton, ``Applications of digital-image-correlation
  techniques to experimental mechanics,'' {\em Experimental Mechanics},
  vol.~25, pp.~232--244, 1985.
\newblock 10.1007/BF02325092.

\bibitem{Hoult2013}
N.~A. Hoult, W.~A. Take, C.~Lee, and M.~Dutton, ``Experimental accuracy of two
  dimensional strain measurements using digital image correlation,'' {\em
  Engineering Structures}, vol.~46, no.~0, pp.~718 -- 726, 2013.

\bibitem{Sutton00}
M.~Sutton, S.~McNeill, J.~Helm, and Y.~Chao, ``Advances in two-dimensional and
  three-dimensional computer vision,'' in {\em Photomechanics} (P.~Rastogi,
  ed.), vol.~77 of {\em Topics in Applied Physics}, pp.~323--372, Springer
  Berlin / Heidelberg, 2000.
\newblock 10.1007/3-540-48800-6-10.

\bibitem{Ha09}
H.-G. Ha, I.-S. Jang, K.-W. Ko, and Y.-H. Ha, ``Robust subpixel shift
  estimation using iterative phase correlation of a local region,''
  pp.~724115--724115--9, 2009.

\bibitem{Gilo11}
M.~Debella-Gilo and A.~K{\"a}{\"a}b, ``Sub-pixel precision image matching for
  measuring surface displacements on mass movements using normalized
  cross-correlation,'' {\em Remote Sensing of Environment}, vol.~115, no.~1,
  pp.~130--142, 2011.

\bibitem{White2003}
D.~White, W.~Take, and M.~Bolton, ``Soil deformation measurement using particle
  image velocimetry (piv) and photogrammetry,'' {\em Geotechnique}, vol.~53,
  no.~7, pp.~619--631, 2003.

\bibitem{Roscoe1963}
K.~Roscoe, {\em Determination of Strains in Soils by X-ray Method}.
\newblock Defense Technical Information Center, 1963.

\bibitem{Andrawes1973}
K.~Andrawes and R.~Butterfield, ``The measurement of planar displacement of
  sand grains,'' {\em Geotechnique}, vol.~23, no.~4, pp.~571--576, 1973.

\bibitem{Butterfield1970}
R.~Butterfield, R.~M. Harkness, and K.~Z. Andrews, ``A stero-photogrammetric
  method for measuring displacement fields,'' {\em Géotechnique}, vol.~20,
  pp.~308--314(6), 1970.

\bibitem{Foreman1965}
J.~W. Foreman, E.~George, and R.~Lewis, ``Measurement of localized flow
  velocities in gases with a laser doppler flowmeter,'' {\em Applied Physics
  Letters}, vol.~7, no.~4, pp.~77--78, 1965.

\bibitem{Barker1977}
D.~B. Barker and M.~E. Fourney, ``Measuring fluid velocities with speckle
  patterns,'' {\em Opt. Lett.}, vol.~1, pp.~135--137, Oct 1977.

\bibitem{Adrian1986}
R.~J. Adrian, ``Image shifting technique to resolve directional ambiguity in
  double-pulsed velocimetry,'' {\em Appl. Opt.}, vol.~25, pp.~3855--3858, Nov
  1986.

\bibitem{Mohammad2001}
M.~Dibajnia, ``Paper no: 145 observation of wave groups in the surf and swaxh
  zones using particle image velocimetry,'' 2001.

\bibitem{White2005}
D.~White, M.~Randolph, and B.~Thompson, ``An image-based deformation
  measurement system for the geotechnical centrifuge,'' {\em International
  Journal of Physical Modelling in Geotechnics}, vol.~5, pp.~1--12(11), 2005.

\bibitem{Peters82}
W.~Peters and W.~Ranson, ``Digital imaging techniques in experimental stress
  analysis,'' {\em Optical Engineering}, vol.~21, no.~3, pp.~427 -- 431, 1982.

\bibitem{Mccormick2012}
N.~McCormick and J.~Lord, ``Digital image correlation for structural
  measurements,'' {\em Proceedings of the ICE - Civil Engineering}, vol.~165,
  pp.~185--190(5), 2012.

\bibitem{Hall2010}
S.~Hall, M.~Bornert, J.~Desrues, Y.~Pannier, N.~Lenoir, G.~Viggiani, and
  P.~Bésuelle, ``Discrete and continuum analysis of localised deformation in
  sand using x-ray $\mu$ct and volumetric digital image correlation,'' {\em
  Geotechnique}, vol.~60, pp.~315--322(7), 2010.

\bibitem{Luo93}
P.~Luo, Y.~Chao, M.~Sutton, and W.~Peters, ``Accurate measurement of
  three-dimensional deformations in deformable and rigid bodies using computer
  vision,'' {\em Experimental Mechanics}, vol.~33, pp.~123--132, 1993.
\newblock 10.1007/BF02322488.

\bibitem{Besnard2012}
G.~Besnard, H.~Leclerc, F.~Hild, S.~Roux, and N.~Swiergiel, ``Analysis of image
  series through global digital image correlation,'' {\em The Journal of Strain
  Analysis for Engineering Design}, vol.~47, no.~4, pp.~214--228, 2012.

\bibitem{Schreir00}
H.~W. Schreier, J.~R. Braasch, and M.~A. Sutton, ``{Systematic errors in
  digital image correlation caused by intensity interpolation},'' {\em Optical
  Engineering}, vol.~39, no.~11, pp.~2915--2921+, 2000.

\bibitem{Rohde09}
G.~Rohde, A.~Aldroubi, and D.~Healy, ``Interpolation artifacts in sub-pixel
  image registration,'' {\em Image Processing, IEEE Transactions on}, vol.~18,
  no.~2, pp.~333--345, 2009.

\bibitem{Liu10}
X.-y. Liu, Q.-c. Tan, and R.-l. Li, ``Study on digital image correlation using
  artificial neural networks for subpixel displacement measurement,'' in {\em
  Advances in Neural Network Research and Applications} (Z.~Zeng and J.~Wang,
  eds.), vol.~67 of {\em Lecture Notes in Electrical Engineering},
  pp.~405--412, Springer Berlin Heidelberg, 2010.
\newblock 10.1007/978-3-642-12990-2-6.

\bibitem{Pan06}
P.~Bing, X.~Hui-min, X.~Bo-qin, and D.~Fu-long, ``Performance of sub-pixel
  registration algorithms in digital image correlation,'' {\em Measurement
  Science and Technology}, vol.~17, no.~6, p.~1615, 2006.

\bibitem{Pan07c}
B.~{Pan}, H.~{Xie}, Z.~{Guo}, and T.~{Hua}, ``{Full-field strain measurement
  using a two-dimensional Savitzky-Golay digital differentiator in digital
  image correlation},'' {\em Optical Engineering}, vol.~46, no.~3, p.~033601,
  2007.

\bibitem{Pan2009a}
B.~Pan, Z.~Wang, and H.~Xie, ``Generalized spatial-gradient-based digital image
  correlation for displacement and shape measurement with subpixel accuracy,''
  {\em The Journal of Strain Analysis for Engineering Design}, vol.~44, no.~8,
  pp.~659--669, 2009.

\bibitem{Lu00}
H.~Lu and P.~Cary, ``Deformation measurements by digital image correlation:
  Implementation of a second-order displacement gradient,'' {\em Experimental
  Mechanics}, vol.~40, pp.~393--400, 2000.
\newblock 10.1007/BF02326485.

\bibitem{Ma2012}
S.~Ma, Z.~Zhao, and X.~Wang, ``Mesh-based digital image correlation method
  using higher order isoparametric elements,'' {\em The Journal of Strain
  Analysis for Engineering Design}, vol.~47, no.~3, pp.~163--175, 2012.

\bibitem{Peters83}
W.~Peters, W.~Ranson, M.~Sutton, T.~Chu, and J.~Anderson, ``Application of
  digital correlation methods to rigid body mechanics,'' {\em Optical
  Engineering}, vol.~22, no.~6, pp.~738 -- 742, 1983.

\bibitem{Zhang06}
Z.-F. Zhang, Y.-L. Kang, H.-W. Wang, Q.-H. Qin, Y.~Qiu, and X.-Q. Li, ``A novel
  coarse-fine search scheme for digital image correlation method,'' {\em
  Measurement}, vol.~39, no.~8, pp.~710 -- 718, 2006.

\bibitem{Bruck89}
H.~Bruck, S.~McNeill, M.~Sutton, and W.~Peters, ``Digital image correlation
  using newton-raphson method of partial differential correction,'' {\em
  Experimental Mechanics}, vol.~29, pp.~261--267, 1989.
\newblock 10.1007/BF02321405.

\bibitem{Vendroux98}
G.~Vendroux and W.~Knauss, ``Submicron deformation field measurements: Part 2.
  improved digital image correlation,'' {\em Experimental Mechanics}, vol.~38,
  no.~2, pp.~86--92, 1998.

\bibitem{Pan11fastDIC}
B.~Pan and K.~Li, ``A fast digital image correlation method for deformation
  measurement,'' {\em Optics and Lasers in Engineering}, vol.~49, no.~7,
  pp.~841 -- 847, 2011.

\bibitem{Schreier02}
H.~Schreier and M.~Sutton, ``Systematic errors in digital image correlation due
  to undermatched subset shape functions,'' {\em Experimental Mechanics},
  vol.~42, no.~3, pp.~303--310, 2002.

\bibitem{Wang02}
H.~Wang and Y.~Kang, ``Improved digital speckle correlation method and its
  application in fracture analysis of metallic foil,'' {\em Optical
  Engineering}, vol.~41, no.~11, pp.~2793--2798, 2002.

\bibitem{Chen93}
D.~J. Chen, F.~P. Chiang, Y.~S. Tan, and H.~S. Don, ``Digital
  speckle-displacement measurement using a complex spectrum method,'' {\em
  Appl. Opt.}, vol.~32, pp.~1839--1849, Apr 1993.

\bibitem{Pitter01}
M.~Pitter, C.~W. See, and M.~Somekh, ``Subpixel microscopic deformation
  analysis using correlation and artificial neural networks,'' {\em Optics
  Express}, vol.~8, no.~6, pp.~322--327, 2001.

\bibitem{Shaopeng2003}
M.~Shaopeng and J.~Guanchang, ``Digital speckle correlation method improved by
  genetic algorithm,'' {\em Acta Mechanica Solida Sinica}, vol.~16, no.~4,
  pp.~366--373, 2003.

\bibitem{Pilch2004}
A.~Pilch, A.~Mahajan, and T.~Chu, ``Measurement of whole-field surface
  displacements and strain using a genetic algorithm based intelligent image
  correlation method,'' {\em Journal of Dynamic Systems, Measurement, and
  Control}, vol.~126, p.~479, 2004.

\bibitem{Poissant2010}
J.~Poissant and F.~Barthelat, ``A novel "subset splitting" procedure for
  digital image correlation on discontinuous displacement fields,'' {\em
  Experimental Mechanics}, vol.~50, no.~3, pp.~353--364, 2010.

\bibitem{Lecompte2006}
D.~Lecompte, A.~Smits, S.~Bossuyt, H.~Sol, J.~Vantomme, D.~V. Hemelrijck, and
  A.~Habraken, ``Quality assessment of speckle patterns for digital image
  correlation,'' {\em Optics and Lasers in Engineering}, vol.~44, no.~11,
  pp.~1132 -- 1145, 2006.

\bibitem{Hua2011}
T.~Hua, H.~Xie, S.~Wang, Z.~Hu, P.~Chen, and Q.~Zhang, ``Evaluation of the
  quality of a speckle pattern in the digital image correlation method by mean
  subset fluctuation,'' {\em Optics and Laser Technology}, vol.~43, no.~1,
  pp.~9 -- 13, 2011.

\bibitem{Ma12}
S.~Ma, J.~Pang, and Q.~Ma, ``The systematic error in digital image correlation
  induced by self-heating of a digital camera,'' {\em Measurement Science and
  Technology}, vol.~23, no.~2, p.~025403, 2012.

\bibitem{White2001}
D.~White, W.~Take, M.~Bolton, and S.~Munachen, ``A deformation measurement
  system for geotechnical testing based on digital imaging, close-range
  photogrammetry, and piv image analysis,'' in {\em Proceedings of the
  International Conference on Soil Mechanics and Geotechnical
  EngineeringRodriquez2011}, vol.~1, pp.~539--542, AA Balkema Publishers, 2001.

\bibitem{Reu2008}
P.~L. Reu and T.~J. Miller, ``The application of high-speed digital image
  correlation,'' {\em The Journal of Strain Analysis for Engineering Design},
  vol.~43, no.~8, pp.~673--688, 2008.

\bibitem{Slama1980}
C.~C. Slama, C.~Theurer, S.~W. Henriksen, {\em et~al.}, {\em Manual of
  photogrammetry.}
\newblock No.~Ed. 4, American Society of photogrammetry, 1980.

\bibitem{Sutton83}
M.~Sutton, W.~Wolters, W.~Peters, W.~Ranson, and S.~McNeill, ``Determination of
  displacements using an improved digital correlation method,'' {\em Image and
  Vision Computing}, vol.~1, no.~3, pp.~133--139, 1983.

\bibitem{KJetter90}
Z.~Kahn-Jetter and T.~Chu, ``Three-dimensional displacement measurements using
  digital image correlation and photogrammic analysis,'' {\em Experimental
  Mechanics}, vol.~30, pp.~10--16, 1990.
\newblock 10.1007/BF02322695.

\bibitem{Haddadi08}
H.~Haddadi and S.~Belhabib, ``Use of rigid-body motion for the investigation
  and estimation of the measurement errors related to digital image correlation
  technique,'' {\em Optics and Lasers in Engineering}, vol.~46, no.~2, pp.~185
  -- 196, 2008.

\bibitem{Tomicevc2013}
Z.~Tomicevc, F.~Hild, and S.~Roux, ``Mechanics-aided digital image
  correlation,'' {\em The Journal of Strain Analysis for Engineering Design},
  vol.~48, no.~5, pp.~330--343, 2013.

\bibitem{Barranger12}
Y.~Barranger, P.~Doumalin, J.~Dupr{\'e}, and A.~Germaneau, ``Strain measurement
  by digital image correlation: Influence of two types of speckle patterns made
  from rigid or deformable marks,'' {\em Strain}, vol.~48, no.~5, pp.~357--365,
  2012.

\bibitem{BookTimoshenko}
S.~Timoshenko and J.~Goodier, {\em Theory of elasticity}.
\newblock McGraw-Hill Book Company, Inc., New York, 2~ed., 1951.

\bibitem{Andrews2012}
T.~Andrews, ``Computation time comparison between matlab and c++ using lauch
  windows.'' Research report submitted to American Institute of Aeronautics and
  Astronautics, California Polytechnic State University San Luis Obispo, SLO,
  CA, 93407, USA, June 2012.

\bibitem{Amy2011}
A.~Rechenmacher, S.~Abedi, O.~Chupin, and A.~Orlando, ``Characterization of
  mesoscale instabilities in localized granular shear using digital image
  correlation,'' {\em Acta Geotechnica}, vol.~6, no.~4, pp.~205--217, 2011.

\bibitem{Sutton2008}
M.~Sutton, J.~Yan, V.~Tiwari, H.~Schreier, and J.~Orteu, ``The effect of
  out-of-plane motion on 2d and 3d digital image correlation measurements,''
  {\em Optics and Lasers in Engineering}, vol.~46, no.~10, pp.~746 -- 757,
  2008.

\bibitem{Barone2001}
S.~Barone, M.~Berghini, and L.~Bertini, ``Grid pattern for in-plane strain
  measurements by digital image processing,'' {\em The Journal of Strain
  Analysis for Engineering Design}, vol.~36, no.~1, pp.~51--59, 2001.

\end{thebibliography}
}

\end{document}